\documentclass[a4paper,fleqn]{cas-dc}

\usepackage{amssymb}
\usepackage{graphicx}
\usepackage[numbers]{natbib}
\usepackage{xcolor}
\usepackage{amsmath}
\usepackage{verbatim}
\usepackage{subfigure}
\usepackage{pifont}
\usepackage{multirow}
\usepackage{adjustbox}
\usepackage{hyphenat}
\usepackage{siunitx}
\usepackage{hyperref}
\usepackage{float}
\usepackage{wrapfig}
\usepackage{threeparttable}
\usepackage{tabularx}
\usepackage{amsthm}
\usepackage{color}
\usepackage{multicol}
\usepackage{graphics}
\usepackage{longtable}
\usepackage{lipsum}
\usepackage{bm}

\input{macrosNew.tex}

\graphicspath{{images/}}

\usepackage{url}
\usepackage{accents}
\usepackage{booktabs}

\newtheorem{myTheo}{Theorem}
\def\tsc#1{\csdef{#1}{\textsc{\lowercase{#1}}\xspace}}
\tsc{WGM}
\tsc{QE}
\tsc{EP}
\tsc{PMS}
\tsc{BEC}
\tsc{DE}
\usepackage{adjustbox}
\usepackage[nolist]{acronym}

\newacro{lfd}[LfD]{Learning from Demonstration}
\newacro{rl}[RL]{Reinforcement Learning}
\newacro{lwr}[LWR]{Locally Weighted Regression}
\newacro{dmp}[DMP]{Dynamic Movement Primitive}
\newacro{gmr}[GMR]{Gaussian Mixture Regression}
\newacro{gpr}[GPR]{Gaussian Process Regression}
\newacro{svr}[SVR]{Support Vector Regression}
\newacro{hmm}[HMM]{Hidden Markov Model}
\newacro{nn}[NN]{Neural Network}
\newacro{dnn}[DNN]{Deep Neural Network}
\newacro{dslfd}[DS-LfD]{Dynamical System-based \ac{lfd}}
\newacro{dsc}[DSC]{Dynamical System-based Control}
\newacro{dsil}[DSIL]{Dynamical System-based Imitation Learning}
\newacro{ds}[DS]{Dynamical System}
\newacro{il}[IL]{Imitation Learning}
\newacro{adsil}[ADSIL]{Autonomous Dynamical Systems-based Imitation Learning}
\newacro{ndsil}[NDSIL]{Non-autonomous  Dynamical Systems-based Imitation Learning}
\newacro{promp}[ProMP]{Probabilistic Movement Primitives}
\newacro{mp}[MP]{Movement Primitives}
\newacro{apf}[APF]{Artificial Potential Field}
\newacro{gmm}[GMM]{Gaussian Mixture Model}
\newacro{mpc}[MPC]{Model Predictive Control}
\newacro{ekf}[EKF]{Extended Kalman Filter}
\newacro{em}[EM]{Expectation Maximization}
\newacro{hsmm}[HSMM]{Hidden Semi-Markov Model}
\newacro{adhsmm}[ADHSMM]{adaptive duration \ac{hsmm}}
\newacro{mpg}[MPG]{Motor Primitive Generalization}
\newacro{seds}[SEDS]{Stable Estimator of Dynamical Systems}
\newacro{fsmds}[FSM-DS]{Fast and Stable Modeling for Dynamical Systems}
\newacro{clfdm}[CLF-DM]{Control Lyapunov Function-based Dynamic Movements}
\newacro{wsaqf}[WSAQF]{Weighted Sum of Asymmetric Quadratic Function}
\newacro{pcgmm}[PC-GMM]{Physically Consistent-\ac{gmm}}
\newacro{cgmr}[C-GMR]{contracting-\ac{gmr}}
\newacro{esds}[ESDS]{Energy-based Stabilizer of Dynamical Systems}
\newacro{pca}[PCA]{Principal Component Analysis}
\newacro{kpca}[K-PCA]{Kernel \ac{pca}}
\newacro{bls}[BLS]{Broad Learning System}
\newacro{qlf}[QLF]{Quadratic Lyapunov Function}
\newacro{ltl}[LTL]{Linear Temporal Logic}
\newacro{sosclf}[SOS-CLF]{Sum Of Squares-control Lyapunov Functions}
\newacro{nsqlf}[NS-QLF]{Neural Shaped-\ac{qlf}}
\newacro{icnn}[ICNN]{Input Convex Neural Network}
\newacro{plyds}[PLYDS]{PoLYnomial Dynamical System}
\newacro{lsdiqp}[LSD-IQP]{Learning Stable Dynamical systems with Iterative Quadratic Programming}
\newacro{ct}[CT]{Contraction Theory}
\newacro{cvf}[CVF]{Contracting Vector Fields}
\newacro{cdsp}[CDSP]{Continuous Dynamical Systems Prior}
\newacro{ccm}[CCM]{Control Contraction Metrics}
\newacro{ncm}[NCM]{Neural Contraction Metrics}
\newacro{sdsef}[SDS-EF]{Stable Dynamical System learning using Euclideanizing Flows}
\newacro{sde}[SDE]{Stochastic Differential Equation}
\newacro{fagil}[FAGIL]{Fail-Safe Adversarial Generative Imitation Learning}
\newacro{dt}[DT]{Diffeomorphic Transform}
\newacro{rsds}[RSDS]{Riemannian Stable Dynamical Systems}
\newacro{dof}[DoF]{Degree of Freedom}
\newacro{power}[PoWER]{Policy learning by Weighting Exploration with the Returns}
\newacro{enac}[eNAC]{episodic Natural Actor Critic}
\newacro{mmt}[MMT]{Model Mediated Teleoperation}
\newacro{elm}[ELM]{Extreme Learning Machine}
\newacro{nes}[NES]{Natural Evolution Strategies}
\newacro{es}[ES]{Evolutionary Strategies}
\newacro{xnes}[xNES]{Exponential \ac{nes}}
\newacro{cmaes}[CMA-ES]{Covariance Matrix Adaptation Evolution Strategy}
\newacro{trcmaes}[TR-CMA-ES]{Trust-Region CMA-ES}
\newacro{ddpg}[DDPG]{Deep Deterministic Policy Gradient}
\newacro{nnmp}[NNMP]{Neural Network-based Movement Primitive}
\newacro{sac}[SAC]{Soft Actor-Critic}
\newacro{orb}[ORB]{Optimal Replay Buffer}
\newacro{rlfd}[rLfD]{residual \ac{lfd}}
\newacro{ppo}[PPO]{Proximal Policy Optimization}
\newacro{cpg}[CPG]{Central Pattern Generators}
\newacro{vpg}[VPG]{Vanilla Policy Gradient}
\newacro{crkr}[CrKR]{Cost-regularized Kernel Regression}
\newacro{hireps}[HiREPS]{Hierarchical Relative Entropy Policy Search}
\newacro{mlp}[MLP]{Multi-Layer Perceptron}
\newacro{cnn}[CNN]{Convolutional Neural Network}
\newacro{tpdmp}[TP-DMP]{Trajectory Parameterized-\ac{dmp}}
\newacro{imednet}[IMEDNet]{Image-to-Motion Encoder-Decoder Networks}
\newacro{cimednet}[CIMEDNet]{Convolutional \ac{imednet}}
\newacro{stimednet}[STIMEDNet]{spatial transformer \ac{imednet}}
\newacro{rnn}[RNN]{Recurrent Neural Networks}
\newacro{dsdnet}[DSDNet]{Deep Segmented \ac{dmp} Networks}
\newacro{aldmp}[AL-DMP]{Arc Length-\ac{dmp}}
\newacro{vimednet}[VIMEDNet]{Variable \ac{imednet}}
\newacro{ndp}[NDPs]{Neural Dynamic Policies}
\newacro{hndp}[H-NDP]{Hierarchical \ac{ndp}}
\newacro{lf}[LF]{Lyapunov Function}
\newacro{pqlf}[P-QLF]{Parameterize-\ac{qlf}}
\newacro{rbfnn}[RBFNN]{Radial Basis Function Neural Network}
\newacro{spd}[SPD]{Symmetric Positive Definite}
\newacro{vic}[VIC]{Variable Impedance Control}
\newacro{pi2}[PI\textsuperscript{2}]{Policy Improvement with Path Integrals}
\newacro{andp}[ANDPs]{Autonomous Neural Dynamic Policies}
\newacro{dagger}[DAgger]{Data Aggregation Approach}
\newacro{llm}[LLM]{Large Language Model}
\newacro{ode}[ODE]{Ordinary Differential Equation}
\newacro{node}[NODE]{Neural \ac{ode}}
\newacro{clf}[CLF]{Control Lyapunov Function}
\newacro{cbf}[CBF]{Control Barrier Function}
\newacro{hsds}[HSDS]{Hand-specified Stable \ac{ds}}
\newacro{bbo}[BBO]{Black-Box Optimization}
\newacro{vsds}[VSDS]{ Variable Stiffness Dynamical System}
\newacro{}[]{}
\newacro{}[]{}

\begin{document}
\let\WriteBookmarks\relax
\def\floatpagepagefraction{1}
\def\textpagefraction{.001}
\shorttitle{Fusion DS with ML in IL: A Review}
\shortauthors{Yingbai Hu et~al.}

\title [mode = title]{
Fusion Dynamical Systems with Machine Learning in Imitation Learning: A Comprehensive Overview 
}

\author[1, 2]{Yingbai Hu}[]
\fnmark[$^{\ddagger}$]
\ead{yingbai.hu@tum.de}
\address[1]{Multi-Scale Medical Robotics Centre, Ltd., The Chinese University of Hong Kong, Hong Kong, China}
\address[2]{
 School of Computation, Information and Technology, Technical University of Munich, Munich, 85748, Germany}

\author[3]{Fares J. Abu-Dakka}[]
\fnmark[$^{\ddagger}$]
\ead{fabudakka@mondragon.edu}
\address[3]{Electronic and Computer Science Department, Faculty of Engineering, Mondragon Unibertsitatea, 20500 Arrasate, Spain}

\author[4]{Fei Chen}[]
\ead{f.chen@ieee.org}
\address[4]{Department of Mechanical and Automation Engineering, The Chinese University of Hong Kong,
Hong Kong, China}

\author[5]{Xiao Luo}[]
\ead{xluo@surgery.cuhk.edu.hk}
\address[5]{Department of Surgery, The Chinese University of Hong Kong, Hong Kong, China}

\author[1, 5]{Zheng Li}[]
\cormark[1]
\ead{lizheng@cuhk.edu.hk}

\author[2]{Alois Knoll}[]
\ead{knoll@in.tum.de}

\author[6]{Weiping Ding}[]
\cormark[1]
\ead{dwp9988@163.com}
\address[6]{School of Information Science and Technology, 
Nantong University, Nantong, 226019, China}

 \nonumnote{* Corresponding author.}
 \nonumnote{$^{\ddagger}$ These authors contributed equally to this work.}
 \nonumnote{This work was supported in part by the National Natural Science Foundation of China under Grant 61976120; in part by the Natural Science Foundation of Jiangsu Province under Grant BK20231337; and in part by the Natural Science Key Foundation of Jiangsu Education Department under Grant 21KJA510004; in part by the Research Grant Council General Research fund under Grant 14202820 and Grant 1421432; in part by the CUHK Strategic Seed Funding for Collaborative Research scheme 22/21 (SSFCRS); and in part by Basque Government (ELKARTEK) projects Proflow KK-2022/00024 and HELDU KK-2023/00055.}

\begin{abstract}
Imitation Learning (IL), also referred to as Learning from Demonstration (LfD), holds significant promise for capturing expert motor skills through efficient imitation, facilitating adept navigation of complex scenarios. A persistent challenge in IL lies in extending generalization from historical demonstrations, enabling the acquisition of new skills without re-teaching.
Dynamical system-based IL (DSIL) emerges as a significant subset of IL methodologies, offering the ability to learn trajectories via movement primitives and policy learning based on experiential abstraction.
This paper emphasizes the fusion of theoretical paradigms, integrating control theory principles inherent in dynamical systems into IL. This integration notably enhances robustness, adaptability, and convergence in the face of novel scenarios.
This survey aims to present a comprehensive overview of DSIL methods, spanning from classical approaches to recent advanced approaches.
We categorize DSIL into autonomous dynamical systems and non-autonomous dynamical systems, surveying traditional IL methods with low-dimensional input and advanced deep IL methods with high-dimensional input. Additionally, we present and analyze three main stability methods for IL: Lyapunov stability, contraction theory, and diffeomorphism mapping.
Our exploration also extends to popular policy improvement methods for DSIL, encompassing reinforcement learning, deep reinforcement learning, and evolutionary strategies. 
The primary objective is to expedite readers' comprehension of dynamical systems' foundational aspects and capabilities, helping identify practical scenarios and charting potential future directions.
By offering insights into the strengths and limitations of dynamical system methods, we aim to foster a deeper understanding among readers.
Furthermore, we outline potential extensions and enhancements within the realm of dynamical systems, outlining avenues for further exploration.
\end{abstract}

\begin{keywords}
Imitation learning 
\sep Dynamical system
\sep Fusion of theoretical paradigms
\sep Stability
\sep Policy exploration
\end{keywords}
\maketitle

\section{Introduction}
With the growing demand for robotic manipulation tasks, traditional pre-programming methods often face limitations in handling complex scenarios due to the challenges in accurately comprehending and modeling tasks \cite{mukherjee2022survey}. Over the decades, the field of robotics has experienced significant advancements propelled by progress in high-level artificial intelligence, low-level control theory, decision-making algorithms, and planning techniques \cite{deitke2022retrospectives}. As a result, the robotics community has shifted its focus towards the development of robots capable of imitating expert behaviors. The robots are envisioned to learn and replicate complex natural motor skills, akin to human capabilities. This shift holds promise in simplifying complex tasks and optimizing industrial applications by harnessing expert-level skills through robot reprogramming.

In this vein, \ac{lfd} arises as a user-friendly and intuitive methodology to teach robots acquiring new tasks. \ac{lfd} can be broadly categorized into two categories: experience abstraction-based methods and \ac{mp}-based learning methods \cite{tavassoli2023learning}. Experience abstraction involves learning new task behaviors by leveraging prior knowledge through policy improvement, where the agent interacts with the environment and updates its policy. Conversely, \ac{mp}-based learning methods primarily generate continuous control signals derived from \acp{ds}.

In practice implementation, demonstrations are collected by human experts and transferred to robots. These demonstrations serve as foundational dataset for generative models that encode motion patterns, empowering robots to perform a diverse range of tasks resembling the demonstrated actions \cite{schaal1996learning}. 

Demonstrations in \ac{il} generally derive from three primary approaches: (\emph{i}) Kinesthetic teaching \cite{kormushev2011imitation}, (\emph{ii}) observations \cite{abbeel2010autonomous, liu2018imitation}, and (\emph{iii}) teleoperation \cite{ havoutis2019learning, delpreto2020helping}. The selection of an approach depends on the specific scenario at hand and the technological resources available. Kinesthetic teaching involves users manipulating the robot to move as desired and recording trajectories in both joint and Cartesian spaces \cite{shavit2018learning}. The quality of the dataset depends on the smoothness of the user experience. However, this method is limited, especially for certain types of robots, such as legged robots. The observation approach is the passive observation of the user performing a task without the robot's active participation. This method is suitable for robots with numerous degrees of freedom or non-anthropomorphic designs, as seen in scenarios like collaborative furniture assembly, autonomous driving, and knot tying \cite{zhu2020off}. Teleoperation involves controlling the robot remotely, often using a master-slave setup or a remote interface. It proves convenient in situations where human intervention is unsafe or impractical \cite{zhang2018deep}, such as underwater operations \cite{havoutis2019learning} or confined surgical spaces \cite{pervez2017novel}. Nevertheless, its applicability is limited due to the need for specific input interfaces and hardware, such as a joystick, graphical user interface, or force feedback device.

The essence of \ac{il} revolves around two key aspects: ensuring the reproduction of demonstrated behaviors and facilitating the model's ability to adapt to novel scenarios absent from the initial dataset. Selecting the appropriate machine learning algorithm or method for \ac{il} is of significant importance. For instance, while some approaches are suitable for large dataset, they might not perform optimally with sparse dataset. Additionally, specific approaches are capable of handling noisy data while still delivering meaningful results. Therefore, the choice of the most suitable approach depends on various factors, including the data nature and the learning objectives. Several machine learning methods have been proposed for \ac{il}. These include traditional methods, \eg \ac{dmp} \cite{ijspeert2013dynamical, saveriano2023dynamic}, \ac{gmm} and \ac{gmr} \cite{calinon2012statistical, su2020improved, khansari2011learning}, as well as  \ac{gpr}, \ac{svr} \cite{khansari2014learning}. Additionally,  \acp{hmm} \cite{asfour2008imitation, herzog2008motion} and \acp{dnn} \cite{ bahl2020neural, theofanidis2021learning} have been explored. These methods frequently incorporate principles from control theory or incorporate from control theory or leverage deep learning to improve generalization performance, particularly when dealing with high-dimensional input data.

Several surveys have explored the field of \ac{lfd}, covering a diverse range of approaches from \ac{mp} to \ac{rl} or inverse \ac{rl}. Schaal\etal conducted a survey on artificial intelligence and neural computation in the context of \ac{il}, with a particular emphasis on applications for humanoid robots \cite{schaal1999imitation}. Their work provided insights into the landscape of \ac{il} techniques, emphasizing their relevance in humanoid robotics. Billard\etal contributed to this field through a survey on programming by demonstration or \ac{lfd} \cite{billard2008survey}. Their comprehensive survey extensively explored human-robot interaction within \ac{lfd}, covering a wide spectrum of techniques and methodologies involved in teaching robots tasks through human demonstrations. Additionally, the authors addressed the challenges inherent in this field while discussing potential applications of robot \ac{lfd}. Argall\etal focused on reviewing the literature related to example state-to-action mappings \cite{argall2009survey}. Their survey categorized different approaches in terms of demonstration methods, policy derivation, and performance evaluation, across various scenarios. A few years later, Billard\etal addressed general challenges within \ac{lfd} \cite{billard2013robot}. They explored fundamental issues such as what to imitate and the evaluation metrics, as well as how to imitate, incorporating aspects like agent motion and force-control tasks. Zhu\etal presented a comprehensive overview of \ac{lfd} specifically within the context of industrial assembly tasks \cite{zhu2018robot}. Their work spanned various aspects, ranging from the methodology of performing demonstrations to the techniques employed in acquiring manipulation features for \ac{il} purposes. Calinon \cite{calinon2018learning} provided a succinct survey on \ac{lfd} approaches, highlighting key research findings related to data collection, methodology, and application. Xie\etal \cite{xie2020robot} concentrated on \ac{lfd} within the domain of robot path planning. Their study highlighted the differences between \ac{il} and inverse \ac{rl} in the context of robot learning. Ravichandar\etal \cite{ravichandar2020recent} presented an overview of machine-learning approaches for robot learning from experts, including the latest advancements up to 2020, while also addressing practical applications and inherent challenges. Saveriano\etal's work offered an extensive overview of \ac{dmp} and various extended versions \cite{saveriano2023dynamic}, elucidating their performance, and application conditions, and provided a tutorial for further exploration in this area. Si\etal \cite{si2021review} focused on immersive teleoperation-based \ac{il} for manipulation skill learning. The comparison between existing reviews about \ac{lfd} and our survey is shown in Table \ref{tab.comparison_survey}.

\begin{table*}
	\caption{Comparison between existing reviews about \ac{lfd} and our survey.}
	\rmfamily\centering
	\resizebox{\textwidth}{!}{%
		\renewcommand\arraystretch{1.1}
		\begin{tabular}{m{0.18\textwidth}m{0.21\textwidth}m{0.47\textwidth}}
			\toprule
			\textbf{Survey} & \textbf{Topics} & \textbf{Description} \\
			\midrule
			Schaal\etal \cite{schaal1999imitation} & 
			\begin{itemize}[leftmargin=0pt,noitemsep, topsep=0pt]
				\item Classical \ac{lfd} methods
				\item AI and neural computation in \ac{lfd}
				\item Humanoid Robot
			\end{itemize}
			& A survey on classical \ac{lfd} methods that introduces concept of \ac{lfd}, and its application in humanoid robots. This survey draws connections to mirror neurons and supramodal representation systems and categorizes three major approaches to \ac{lfd}: learning a control policy, learning from demonstrated trajectories, and model-based \ac{lfd}. \\
			\hline	
			Billard\etal \cite{billard2008survey} & 
			\begin{itemize}[leftmargin=0pt,noitemsep, topsep=0pt]
				\item Classical \ac{lfd} methods
				\item Engineering/Biologically-oriented methods
			\end{itemize}
			& A survey on classical \ac{lfd} methods that presents the two main \ac{lfd} methods: engineering and Biologically oriented methods. \\
			\hline
			Argall\etal \cite{argall2009survey} & 
			\begin{itemize}[leftmargin=0pt,noitemsep, topsep=0pt]
				\item Classical \ac{lfd} methods
				\item Demonstrator: both
				\item Policy derivation
			\end{itemize}
			& A survey focuses on classical \ac{lfd} methods, categorizing them into two fundamental phases: gathering demonstration examples from various demonstrators to record data; the second phase centers on deriving a policy by mapping states to actions based on these examples.\\
			\hline
			Billard\etal \cite{billard2013robot} \& Calinon\etal  \cite{calinon2018learning} & 
			\begin{itemize}[leftmargin=0pt,noitemsep, topsep=0pt]
				\item Classical \ac{lfd} methods
				\item Principle and concept of \ac{lfd}
			\end{itemize}
			& These surveys provide an overview of classical \ac{lfd} methods, introducing the fundamental principles and concepts that underlie \ac{lfd} approaches, encompassing key components such as demonstrators, classical methodologies, and special functions. \\
			\hline
			Zhu\etal \cite{zhu2018robot} &
			\begin{itemize}[leftmargin=0pt,noitemsep, topsep=0pt]
				\item Classical \ac{lfd} methods
				\item Assembly operations tasks
			\end{itemize}
			& A survey that introduces the classical \ac{lfd} methods for assembly tasks.\\
			\hline
			Xie\etal \cite{xie2020robot} &
			\begin{itemize}[leftmargin=0pt,noitemsep, topsep=0pt]
				\item Classical \ac{lfd} methods
				\item Path planning
				\item \ac{rl} and Inverse \ac{rl}
			\end{itemize}
			& A survey offers a specialized perspective on \ac{lfd} in the context of path planning. This survey focuses on the classical \ac{lfd} methods, \ac{rl}, and inverse \ac{rl} methods on path planning. \\
			\hline
			Ravichandar\etal \cite{ravichandar2020recent} &
			\begin{itemize}[leftmargin=0pt,noitemsep, topsep=0pt]
				\item General \ac{lfd} methods
				\item Mature and emerging application
			\end{itemize}
			& A survey ~that conducts an extensive exploration of general \ac{lfd} methods, encompassing both mature and emerging applications.\\
			\hline
			Saveriano\etal \cite{saveriano2023dynamic} &
			\begin{itemize}[leftmargin=0pt,noitemsep, topsep=0pt]
				\item \ac{dmp}
				\item \ac{rl}, Deep \ac{il}, Lifelong Learning
			\end{itemize}
			& A survey that provides a comprehensive review and tutorial of \ac{dmp}, covering various versions of \ac{dmp} and applications.\\
			\hline
			Si\etal~ \cite{si2021review} &
			\begin{itemize}[leftmargin=0pt,noitemsep, topsep=0pt]
				\item Classical \ac{lfd} methods
				\item Demonstrator: teleoperation
				\item Manipulation
			\end{itemize}
			& A survey that provides an overview of classical \ac{lfd} methods, with a specific focus on multimodal teleoperation demonstrator-based general \ac{lfd} techniques for robot manipulation, such as force control.\\
			\hline
			\hline
			This paper   & 
			\begin{itemize}[leftmargin=0pt,noitemsep, topsep=0pt]
				\item DSIL methods
				\item NDS, ADS
				\item Stability methods
				\item \ac{rl}, Deep \ac{rl}, Deep \ac{il}
				\item Generalization
				\item Online adaptation
			\end{itemize}
			& This survey centers on DSIL, categorizing and presenting the evolution of DSIL from classical methods to the latest advancements in deep \ac{il}, and \ac{rl}. Our survey also delves into the essential stability characteristics, spanning from theoretical foundations to published papers. \\ 
			\bottomrule
		\end{tabular}
	}
	\label{tab.comparison_survey}
\end{table*} 

\begin{figure*}
	\def\svgwidth{\textwidth}
	{\fontsize{10}{10}\selectfont\sf
		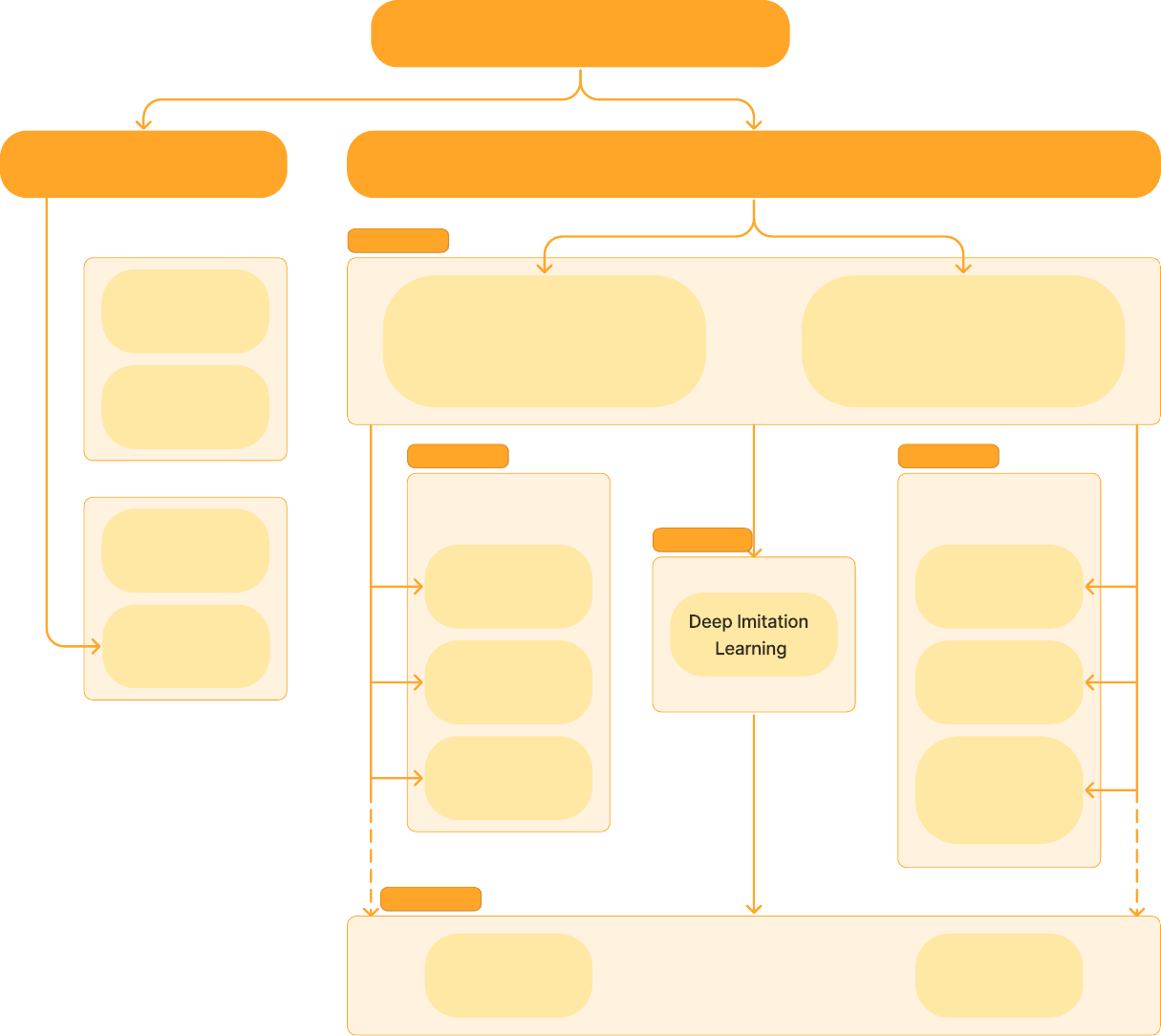}
	\caption{A taxonomy of existing directions for \ac{ds}.}
	\label{fig:survey_framework}
\end{figure*}

Despite the numerous surveys that have extensively covered the broader landscape of \ac{lfd} methods and their applications, there is a noticeable gap when it comes to \ac{dsil}. Due to the rapid growth of the field, there is a need for a  survey to summarize the latest advancements within \ac{dsil}. \ac{dsil} represents a specialized form of \ac{il}, conceptualizing the learning model as a \ac{ds}, as outlined by Zadeh \cite{zadeh2012dynamical}.

Figure \ref{fig:survey_framework} illustrates the proposed taxonomy that categorizes existing directions in \acp{ds}. \acp{ds} encompass a set of differential equations meticulously analyzing the influence of time and force on an agent's behavior, providing insights into the continuous evolution of agents over time. The left side pertains to \ac{dsc}, where such systems are categorized into two types: linear control systems and nonlinear control systems, or discrete control systems and continuous control systems \cite{katok1995introduction} \cite{brin2002introduction}. In this survey, we specifically concentrate on the \ac{dsil} depicted on the right side of Fig. \ref{fig:survey_framework}. The fusion of control theory principles with \ac{ds}-based learning models via theory fusion yields enhancements across various facets of system performance, including stability, robustness, and convergence speed. This intersection between control theory and learning systems results in a synergistic effect that magnifies the capabilities of the learning models.

Similar to general \ac{il} methods, \ac{dsil} can also be classified into two primary categories: \ac{mp}-based learning methods and experience abstraction-based methods. \ac{mp}-based methods inherently combine control system principles with machine learning capabilities. This fusion enables the fine-tuning of training parameters, thereby enhancing the model's robustness and convergence capabilities. On the other hand, experience abstraction-based methods are effectively paired with \ac{rl} or inverse \ac{rl}. This integration facilitates the refinement of policy parameters within the \ac{ds} during interactions with the environment. Consequently, the model acquires adaptability to new tasks, enabling actions such as rejecting perturbations, navigating via points, and avoiding obstacles.

The contributions are summarized as:
\begin{itemize}
    \item \textbf{Comprehensive Survey in \ac{dsil}}: This paper provides a comprehensive survey encompassing 213 papers on the landscape of \acf{dsil} methods. It covers traditional \ac{lfd} approaches suitable for low-dimensional inputs as well as the latest advancements in deep \ac{lfd} tailored for high-dimensional inputs. 
    \item \textbf{Exploration of Stability Methods in \ac{dsil}}: The paper extensively analyzes three common stability methods in the context of \ac{dsil}. it offers a deep exploration of the theoretical foundations associated with ensuring stability in \ac{dsil}.
    \item \textbf{Policy Learning Methods in \ac{dsil}}: This survey thoroughly explores a spectrum of policy learning techniques, encompassing both traditional \ac{rl} ones and methods grounded in policy learning based on \ac{es}. Furthermore, it sheds light on the most recent breakthroughs in deep \ac{rl} methodologies specifically designed to address the nuances of \ac{dsil}.
    \item \textbf{Research Directions and Challenges in \ac{dsil}}: Beyond providing an overview of the existing landscape and methodologies, this paper identifies and outlines the key research directions. It also discusses current challenges and open problems in \ac{dsil}. 
\end{itemize}
Our exploration spans both theoretical advancements and practical applications within the realm of \ac{dsil}.

The rest of this survey is organized as follows:
Section \ref{sec.mp_learning} introduces two types of \acp{ds}: autonomous and non-autonomous, contextualizing their roles within \ac{il}.
Section \ref{sec.stable} covers three common stability approaches employed in \ac{dsil}, offering insights into ensuring stability within \ac{dsil}.
Section \ref{sec.policy_learning} discusses existing policy learning methods, including \ac{rl}, \ac{es}, and deep \ac{rl}, among others.
In Section \ref{sec.deepIL}, we introduce deep \ac{il} with high-dimensional input.
In Section \ref{sec.future_directions},  we discuss challenges and future directions on \ac{dsil}.
Finally, our survey is concluded in Section \ref{sec.conclusion}.

\section{\ac{mp} learning with low dimension input}
\label{sec.mp_learning}

\begin{table*}
	\small\sf\centering
	\caption{Description of key abbreviations.}
	\resizebox{\linewidth}{!}{%
	{\renewcommand\arraystretch{1} 
		\begin{tabular}{m{0.09\linewidth}m{0.41\linewidth}||m{0.09\linewidth}m{0.41\linewidth}}
			\noalign{\hrule height 1.5pt}
			\rowcolor{Light0}
			\acs{il}	& \acl{il} & \acs{rl}		&\acl{rl}\\
			\rowcolor{Light0}
			{\acs{lwr}}	& {\acl{lwr}} & \acs{dmp}	& \acl{dmp}  \\
			\rowcolor{Light0}
			\acs{gmm}		& \acl{gmm} &
			\acs{gmr}	& \acl{gmr}\\
			\rowcolor{Light0}
			\acs{dof}	& \acl{dof}&
			\acs{gpr}	& \acl{gpr}\\
			\rowcolor{Light0}
			\acs{svr}	& \acl{svr}&
			\acs{hmm}	& \acl{hmm}\\
			\rowcolor{Light0}
			\acs{dnn}	& \acl{dnn}&
			\acs{dsil}		& \acl{dsil}\\
			\rowcolor{Light0}
			\acs{adsil}		& \acl{adsil}&
			\acs{ndsil}	& \acl{ndsil}\\
			\rowcolor{Light0}
			\acs{promp}	& \acl{promp}&
			\acs{mp}	& \acl{mp}\\
			\rowcolor{Light0}
			\acs{apf}	& \acl{apf}&
			\acs{mpc}	& \acl{mpc}\\
			\rowcolor{Light0}
			\acs{ekf}	& \acl{ekf}&
			\acs{em}	& \acl{em}\\
			\rowcolor{Light0}
			\acs{hsmm}	& \acl{hsmm}&
			\acs{adhsmm}	& \acl{adhsmm}\\
			\rowcolor{Light0}
			\acs{pi2}	& \acl{pi2}&
			\acs{cmaes}	& \acl{cmaes}\\
			\rowcolor{Light0}
			\acs{mpg}	& \acl{mpg}&
			\acs{seds}	& \acl{seds}\\
			\rowcolor{Light0}
			{\acs{fsmds}} & {\acl{fsmds}}&
			\acs{clfdm}	& \acl{clfdm}\\
			\rowcolor{Light0}
			\acs{wsaqf}	& \acl{wsaqf}&
			\acs{pcgmm}	& \acl{pcgmm}\\
			\rowcolor{Light0}
			\acs{cgmr}	& \acl{cgmr}&
			\acs{esds}	& \acl{esds}\\
			\rowcolor{Light0}
			\acs{pca}	& \acl{pca}&
			\acs{kpca}	& \acl{kpca}\\
			\rowcolor{Light0}
			\acs{bls}	& \acl{bls}&
			\acs{qlf}	& \acl{qlf}\\
			\rowcolor{Light0}
			\acs{ltl}	& \acl{ltl}&
			\acs{sosclf}	& \acl{sosclf}\\
			\rowcolor{Light0}
			\acs{nsqlf}	& \acl{nsqlf}&
			\acs{icnn}	& \acl{icnn}\\
			\rowcolor{Light0}
			\acs{plyds}	& \acl{plyds}&
			\acs{lsdiqp}	& \acl{lsdiqp}\\
			\rowcolor{Light0}
			\acs{ct}	& \acl{ct}&
			\acs{cvf}	& \acl{cvf}\\
			\rowcolor{Light0}
			\acs{cdsp}	& \acl{cdsp}&
			\acs{ccm}	& \acl{ccm}\\
			\rowcolor{Light0}
			\acs{ncm}	& \acl{ncm}&
			\acs{sdsef}	& \acl{sdsef}\\
			\rowcolor{Light0}
			\acs{sde}	& \acl{sde}&
			\acs{fagil}	& \acl{fagil}\\
			\rowcolor{Light0}
			\acs{dt}	& \acl{dt}&
			\acs{rsds}	& \acl{rsds}\\
			\rowcolor{Light0}
			\acs{ode}	& \acl{ode}&
			\acs{power}	& \acl{power}\\
			\rowcolor{Light0}
			\acs{enac}	& \acl{enac}&
			\acs{mmt}	& \acl{mmt}\\
			\rowcolor{Light0}
			\acs{elm}	& \acl{elm}&
			\acs{nes}	& \acl{nes}\\
			\rowcolor{Light0}
			\acs{es}	& \acl{es}&
			\acs{xnes}	& \acl{xnes}\\
			\rowcolor{Light0}
			\acs{trcmaes}	& \acl{trcmaes}&
			\acs{ddpg}	& \acl{ddpg}\\
			\rowcolor{Light0}
			\acs{nnmp}	& \acl{nnmp}&
			\acs{sac}	& \acl{sac}\\
			\rowcolor{Light0}
			\acs{orb}	& \acl{orb}&
			\acs{rlfd}	& \acl{rlfd}\\
			\rowcolor{Light0}
			\acs{ppo}	& \acl{ppo}&
			\acs{cpg}	& \acl{cpg}\\
			\rowcolor{Light0}
			\acs{vpg}	& \acl{vpg}&
			\acs{crkr}	& \acl{crkr}\\
			\rowcolor{Light0}
			\acs{hireps}	& \acl{hireps}&
			\acs{mlp}	& \acl{mlp}\\
			\rowcolor{Light0}
			\acs{cnn}	& \acl{cnn}&
			\acs{tpdmp}	& \acl{tpdmp}\\
			\rowcolor{Light0}
			\acs{imednet}	& \acl{imednet}&
			\acs{cimednet}	& \acl{cimednet}\\
			\rowcolor{Light0}
			\acs{stimednet}	& \acl{stimednet}&
			\acs{rnn}	& \acl{rnn}\\
			\rowcolor{Light0}
			\acs{dsdnet}	& \acl{dsdnet}&
			\acs{aldmp}	& \acl{aldmp}\\
			\rowcolor{Light0}
			\acs{vimednet}	& \acl{vimednet}&
			\acs{ndp}	& \acl{ndp}\\
			\rowcolor{Light0}
			\acs{hndp}	& \acl{hndp}&
			\acs{lf}	& \acl{lf}\\
			\rowcolor{Light0}
			\acs{pqlf}	& \acl{pqlf}&
			\acs{rbfnn}	& \acl{rbfnn}\\
            \rowcolor{Light0}
            \acs{dagger}	& \acl{dagger}&
			\acs{llm}	& \acl{llm}\\
             \rowcolor{Light0}
            \acs{clf}	& \acl{clf}&
			\acs{cbf}	& \acl{cbf}\\
             \rowcolor{Light0}
            \acs{bbo}	& \acl{bbo}&
			\acs{vsds}	& \acl{vsds}\\
			\noalign{\hrule height 1.5pt}
		\end{tabular}
}}
	\label{tab:abrr}
\end{table*} 
\begin{table*}
	\small\sf\centering
	\caption{Description of key notations.}
	\resizebox{\linewidth}{!}{%
	{\renewcommand\arraystretch{1} 
		
		\centering
		\begin{tabular}{m{0.15\linewidth}m{0.01\linewidth}m{0.34\linewidth}||m{0.155\linewidth}m{0.01\linewidth}m{0.335\linewidth}}
			\noalign{\hrule height 1.5pt}
			\rowcolor{Light0}
			$K$ & $\triangleq$ & \# of basis/Gaussian function
			&
			$k$ & $\triangleq$ & index $: k=1,2,\ldots, K$ \\
			\rowcolor{Light0}
			$\mathcal{O}$ & $\triangleq$ & \# of demonstrations
			&
			$o$ & $\triangleq$ & index $: o=1,2,\ldots, \mathcal{O}$ \\
			\rowcolor{Light0}
			$N$ & $\triangleq$ & \# of trajectory length
			&
			$n$ & $\triangleq$ & index $: n=1,2,\ldots, N$ \\
			\rowcolor{Light0}
            $\mathcal{I}$ & $\triangleq$ & \# of samples in exploration
			&
			$i$ & $\triangleq$ & index $: i=1,2,\ldots, \mathcal{I}$ \\
			\rowcolor{Light0}
			$\tau$ & $\triangleq$ & \# time modulation parameter
			&
			$c_k, \omega_k$ & $\triangleq$ & centers and widths of Gaussian \\
			\rowcolor{Light0}
			$\alpha_z, \beta_z, \alpha_\xi$ & $\triangleq$ & positive gain
			&
			$T_p$ & $\triangleq$ &  rotation matrix \\
			\rowcolor{Light0}
			$\eta_p, \mu_1, \mu_2, \mu_3, A_p$ & $\triangleq$ & positive gain
			&
			$l_1, l_2, l_3$ & $\triangleq$ & axis length of $C_p$\\
                \rowcolor{Light0}
			$\rho$ & $\triangleq$ & positive gain
			&
			$x_1, x_2, x_3$ & $\triangleq$ & centers of $C_p$\\
                \rowcolor{Light0}
               $\kappa$ & $\triangleq$ & adjusted parameter
			&
			$\eta_l$ & $\triangleq$ &  learning rate \\
			\rowcolor{Light1}
			$\xi$ & $\triangleq$ & decay phase variable
			&
			$x,\dot x, \ddot x $ & $\triangleq$ & trajectory data \\
			\rowcolor{Light1}
			$C_p$ & $\triangleq$ & obstacle function
			&
			$U(x, \dot x)$ & $\triangleq$  & potential field function \\
			\rowcolor{Light1}
			$f(x), f(x,t)$ & $\triangleq$ & dynamical system function
			&
			$G(x, \dot x)$ & $\triangleq$ & repulsive force function\\
			\rowcolor{Light1}
                $\theta$ & $\triangleq$ &  learnable parameters
			&
			$\vartheta $ & $\triangleq$ & angle between two vectors \\
			\rowcolor{Light1}
			  $x^*$ & $\triangleq$ & goal position 
			&
			${\varphi _i}$ & $\triangleq$ & basis function  \\
			\rowcolor{Light1}
			$q$,  $q_r$ and  $\dot q$, ${\dot q}_r$  & $\triangleq$ & actual and desired joint angle , its 1st time-derivative
			&
			$z, \dot z$ & $\triangleq$ & scaled velocity and acceleration  \\
			\rowcolor{Light1}
			$f_{\xi, \theta}$, $f_\xi$ & $\triangleq$ & forcing term for different spaces
			&
			$h_i$ & $\triangleq$ & weights \\
			\rowcolor{Light1}
			$p(k)$, $p(x|k)$ & $\triangleq$ & prior probability and conditional Probability 
			&
			$R$ & $\triangleq$ & parameter matrix \\
			\rowcolor{Light1}
			$\mu _k^{\dot x}$, $\mu _k^x$, $\mu _\theta$ & $\triangleq$ & mean
			&
			$Dis(x)$ & $\triangleq$ & distance between robot and obstacles \\
                \rowcolor{Light1}
			  $\Sigma _k^x$, $\Sigma _k^{\dot xx}$, ${\Sigma _\theta }$  & $\triangleq$ &  covariance
			&
			$V(x)$, ${\tilde V}(x)$ & $\triangleq$ & Lyapunov function \\ 
               \rowcolor{Light3}
			  $\varpi $  & $\triangleq$ &  weight of $\rm{PI}^2$
			&
			$u_{t_j}$, $u$, ${u_{f\omega }}$ & $\triangleq$ & control input \\
			\rowcolor{Light3}
			$\delta_x$, $\dot{\delta}_x$  & $\triangleq$ & virtual displacement, its 1st time-derivative
			&
			$\mathcal{M}$ & $\triangleq$ & a Riemannian manifold\\
			\rowcolor{Light3}
			$J$, $J_\Psi$ & $\triangleq$ & Jacobian matrix
			&
			$\psi$  & $\triangleq$ & diffeomorphism mapping \\
			\rowcolor{Light3}
			$\hbar$  & $\triangleq$ & coordinate transformation output
			&
			$S$, $\bar S$ & $\triangleq$ & cost function \\
                \rowcolor{Light3}
			$\phi_{{t_N}}$, $\gamma_{{t_i}}$  & $\triangleq$ & terminal and immediate cost
			&
			$\lambda \left( x \right)$ & $\triangleq$ & eigen values \\
                \rowcolor{Light3}
			$P$  & $\triangleq$ & positive define matrix
			&
			$Q$, $H$ & $\triangleq$ & negative define matrix \\
                \rowcolor{Light3}
			$\mathbb{J}$  & $\triangleq$ & expected cost
			&
			$\varepsilon _t$, $\varepsilon _{k, t}$ & $\triangleq$ & exploration noise \\
                \rowcolor{Light3}
			${\nu_i}$  & $\triangleq$ & utilities function
			&
			$r\left( {{\theta _i}} \right)$ & $\triangleq$ & reward \\
               \rowcolor{Light3}
			$\Phi$, ${\Phi _{k,t}}$  & $\triangleq$ & control matrix
			&
			$\nabla $ & $\triangleq$ & gradient \\
                \rowcolor{Light3}
			$L_i$  & $\triangleq$ & i-th sample trajectory of $\rm{PI}^2$
			&
			$F$ & $\triangleq$ & Fisher matrix \\
            \rowcolor{Light3}
			$K^p$  & $\triangleq$ & stiffness gains
			&
			$K^v$  & $\triangleq$ & damping gains \\
			\noalign{\hrule height 1.5pt} 
		\end{tabular}
}}
	\label{tab:notation}
\end{table*} 

\ac{mp} can be categorized into two main types: (a) \textbf{Dynamics-based approaches:} These methods are capable of generating smooth trajectories from any given initial state. A notable example of this type is \ac{dmp} \cite{ijspeert2002movement}. (b) \textbf{Probabilistic approaches:} This category focuses on capturing higher-order statistics of motion. An example within this category is \ac{promp} \cite{paraschos2018using}. These two categories offer distinct approaches for modeling and generating movements, each with its strengths and applications.

This section introduces an overview of \ac{mp}-based \ac{ds} learning focusing on low-dimensional input. In \cite{zeestraten2016variable}, the authors categorized \ac{mp} learning methods into two categories: autonomous and non-autonomous systems.
\ac{ndsil} is a field that deals with the imitation and control of systems whose behavior evolves, over time, in response to external inputs or forces. \ac{ndsil} depends on time-varying factors or control inputs to drive their trajectories.

\begin{enumerate}
    \item Time-dependent behavior: \ac{ndsil} deals with systems whose behavior is influenced by external factors or control inputs, adding complexity to modeling and imitation. This time-dependent aspect requires sophisticated approaches to accurately replicate and predict behaviors that evolve over time.
    \item Generalization: in \ac{ndsil}, \ac{lfd} involves capturing not only the nominal behavior but also the variations induced by external inputs. Robust generalization in \ac{ndsil} demands a broader understanding of how diverse external factors affect system behavior, enhancing adaptability and performance in varied scenarios.
\end{enumerate}

\change{The distinctions between autonomous and non-autonomous \ac{ds} lie primarily in how they evolve over time and the factors that influence their behavior, Table \ref{tab.ADSvsNADS}.}
\begin{table}
	\caption{\change {Autonomous \ac{ds} vs. non-autonomous \ac{ds}.}}
	\rmfamily\centering
	\resizebox{\linewidth}{!}{%
		\renewcommand\arraystretch{1} 
		\centering
		\change {\begin{tabular}{|>{\arraybackslash}m{0.47\linewidth}|>{\arraybackslash}m{0.47\linewidth}|}
				\noalign{\hrule height 1.5pt}
				\multicolumn{1}{|c|}{\textbf{Autonomous \ac{ds}}}        & \multicolumn{1}{c|}{\textbf{Non-autonomous \ac{ds}}}  \\ \hline
				The evolution of the system depends solely on its current state. There are no external inputs or time-varying parameters that directly influence the system's dynamics, \eg equation \eqref{eq.autonomous} & The evolution of the system depends not only on its current state but also on external inputs or time-varying parameters that directly influence its dynamics, \eg equation \eqref{eq.nonautonomous}\\ \hline
				They are often characterized by fixed equations of motion that describe how the system evolves over time. & They are often characterized by equations of motion that explicitly include time-varying terms or external inputs. \\ \hline
				Examples include simple mechanical systems like pendulums, as well as more complex systems like autonomous vehicles navigating without external control. & Examples include systems subjected to time-varying external forces, such as robots controlled by external commands or systems influenced by environmental factors that change over time. \\ \hline
				\noalign{\hrule height 1.5pt}
		\end{tabular}}%
	}
	\label{tab.ADSvsNADS}
\end{table}

\begin{alignat}{2}
\dot {x}&=f(x) \label{eq.autonomous}\\
\dot {x}&=f(x,t)  \label{eq.nonautonomous}
\end{alignat}

These \ac{mp} models,  described by \eqref{eq.autonomous} and \eqref{eq.nonautonomous},  facilitate the generation of motion sequences,  incorporating an initial state as part of their functionality. This allows the prediction or simulation of behaviors based on the specified starting conditions.  For a better understanding, we have summarized the used abbreviations and the key notations in Tables \ref{tab:abrr} and \ref{tab:notation}.

In the following subsections, we begin by introducing the \ac{ndsil} methods. Subsequently, we present the \ac{adsil} methods.

\subsection{\acf{ndsil}}
\label{sec.ndsil}

The evolution of non-autonomous systems relies on external variables beyond the system state.
 One of the classical methods of \ac{ndsil} is the \acp{dmp} which was initially developed by Ijspeert\etal in 2002 \cite{ijspeert2002movement} and further refined in 2013 \cite{ijspeert2013dynamical}.
 
The \ac{dmp} framework is a straightforward damped spring model coupled with a forcing function to learn trajectories. The damped spring model attracts the robot towards a defined goal position, while the forcing function guides the robot to follow a given trajectory. Consequently, it exhibits a property of globally converging towards a goal position from any initial position.
The concept behind \ac{dmp} involves envisioning complex movements as compositions of sequential or simultaneous primitive movements. Consequently, \ac{dmp} is capable of imitating demonstrations and reproducing similar motions, particularly for point-to-point trajectories or periodic trajectories.
Moreover, advancements in \ac{dmp} extend its capabilities to encode the orientation \cite{AbuDakka2015Adaptation,saveriano2019merging,seleem2019guided,seleem2020development,sidiropoulos2021human,koutras2020correct} and \ac{spd} \cite{abudakka2020Geometry} trajectories, facilitating continuous transitions between successive motion primitives.  Interested readers can refer to \cite{saveriano2023dynamic} for a more comprehensive survey with various formulations and extensions of \acp{dmp}.

\begin{figure*}
	\def\svgwidth{\textwidth}
	{\fontsize{10}{10}\selectfont\sf
		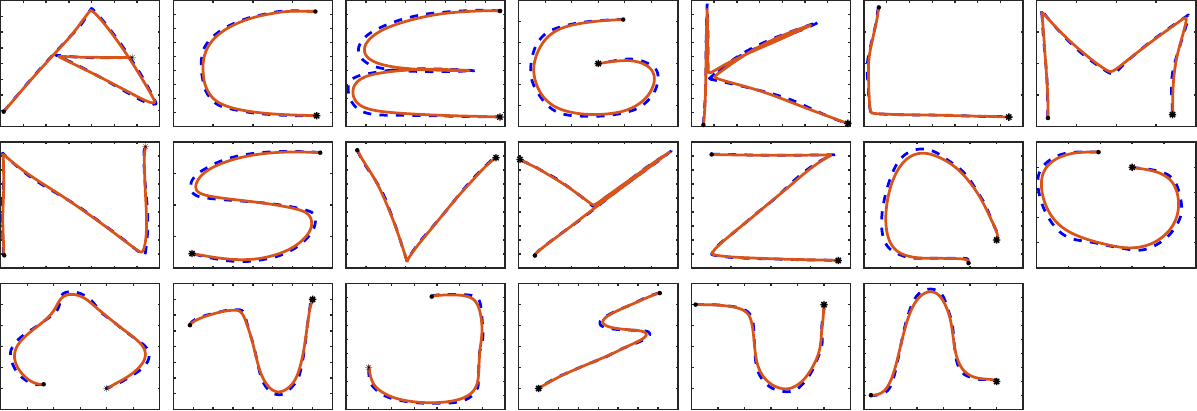}
	\caption{The imitation performance of \ac{ndsil}  utilizing \ac{dmp} is evaluated on a dataset containing 20 instances of human handwriting motions. The black `$\cdot$' and `*' symbols denote the initial and goal points, respectively. The blue dashed line represents the demonstration, while the solid brown line illustrates the reproduced imitation.}
	\label{fig.dmp_20huamnmotion}
\end{figure*}

The basic formulation of a single \ac{dmp} is defined as:
\begin{alignat}{2}
\tau \dot z&=\alpha_z\left(\beta_z(x^*-x)-z\right)+f_{\xi, \theta}(\xi, \theta) \label{eq.Transdyn1}\\
\tau \dot x&=z \label{eq.Transdyn2}\\
\tau \dot \xi&=-\alpha_\xi \xi \label{eq.Cansys}\\
f_{\xi, \theta}\left( \xi, \theta \right) &= \frac{{\sum\nolimits_{k = 1}^K {{\theta _k}{\varphi _k}(\xi)} }}{{\sum\nolimits_{k = 1}^K {{\varphi_k}(\xi)} }}\xi \label{eq.forceterm}\\
{\varphi _k}(\xi) &= \exp \left( { - {\omega_k}{{(\xi - {c_k})}^2}} \right) \label{eq.Gausskernel}
\end{alignat}
where Eqs. \eqref{eq.Transdyn1} and \eqref{eq.Transdyn2} are transformation system,  Eq. \eqref{eq.Cansys} is the canonical system, Eq. \eqref{eq.forceterm} is the forcing term, and  Eq. \eqref{eq.Gausskernel} is the Gaussian kernel. The parameters $\tau$, $\alpha_z$, $\beta_z$, $\alpha_\xi$ are positive constant and $\theta$ is the shape parameter which is used for training. $x^*$ is the goal position of the \ac{ds}. Indeed, \ac{dmp} represents a time-dependent \ac{ds}. 
The performance of \ac{ndsil} utilizing \ac{dmp} on the LASA dataset $\footnote[1]{dataset:https://bitbucket.org/khansari/lasahandwritingdataset/src/master}$ \cite{khansari2011learning} is illustrated in Fig. \ref{fig.dmp_20huamnmotion}.

\begin{table*}
	\caption{Different potential functions for obstacle avoidance of \ac{ndsil}. $Dis(x)$ is the distance between robot and obstacles; $Dis_0$ is the threshold value; $\eta_p$, $\mu_1$, $\mu_2$, $\mu_3$ and $A_p$ are constant parameters.}
	\rmfamily\centering
	\resizebox{\linewidth}{!}{%
		\renewcommand\arraystretch{1} 
		\centering
		\begin{tabular}{|>{\arraybackslash}m{0.12\textwidth}|>{\centering\arraybackslash}m{0.23\textwidth}|>{\centering\arraybackslash}m{0.45\textwidth}|}
			\noalign{\hrule height 1.5pt}
			\multirow{1}{*}{Methods} 
			& Repulsive functions        &Auxiliary function   \\
			\hline
			Point-steering \cite{hoffmann2009biologically} 
			& $G\left( {x,\dot x} \right) = \eta_p T_p\dot x\vartheta \exp \left( { - \mu_1 \vartheta } \right)$        &$\vartheta  = \arccos \left( {\frac{{\left\langle {o - x,\dot x} \right\rangle }}{{\left\| {o - x} \right\|\left\| {\dot x} \right\|}}} \right)$                                \\
			\hline
			Point-static \cite{warren1989global, ginesi2019dynamic} 
			& $G\left( {x,\dot x} \right) =  - {\nabla _x}U\left( {x,\dot x} \right)$      &$U\left( {x,\dot x} \right) = \left\{ \begin{array}{l}
				\frac{\eta_p }{2}{\left( {\frac{1}{{Dis(x)}} - \frac{1}{{Dis_0}}} \right)^2}~~Dis(x) \le {Dis_0}\\
				0{\rm{         ~~~~   ~~~  ~~~    ~~~  ~~~      ~~~~~~                }}Dis(x) > Di{s_0}
			\end{array} \right.$                               \\
			\hline     
			Point-dynamic \cite{park2008movement}
			& $G\left( {x,\dot x} \right) =  - {\nabla _x}U\left( {x,\dot x} \right)$        &$U\left( {x,\dot x} \right) = \left\{ {\begin{array}{*{20}{l}}
					{{\eta _p}{{\left( { - \cos \vartheta } \right)}^{{\mu _1}}}\frac{{\left\| {\dot x} \right\|}}{{Dis(x)}}\;\;\;\;\;{\rm{if}}\;\;\vartheta  \in \left( {\frac{\pi }{2},\pi } \right]}\\
					{0\;\;\;\;\;\;\;\;\;\;\;\;\;\;\;\;\;\;\;\;\;\;\;\;\;\;\;\;\;\;{\rm{if}}\;\;\vartheta  \in \left[ {0,\frac{\pi }{2}} \right)}
			\end{array}} \right.$                               \\
			&         &$\cos \vartheta  = \frac{{\left\langle {\dot x,x - o} \right\rangle }}{{\left\| {x - o} \right\|\left\| {\dot x} \right\|}}$                             \\
			\hline       
			Volume-static \cite{ginesi2019dynamic}
			& $G\left( {x,\dot x} \right) =  - {\nabla _x}U\left( {x,\dot x} \right)$     &$U\left( {x,\dot x} \right) = \frac{{A_p\exp \left( { - \eta_p C_p(x)} \right)}}{{C_p(x)}}$                                \\
			&         &$C_p(x) = {\left( {\frac{{{x_1} - {{\hat x}_1}}}{{{l_1}}}} \right)^{2{\mu _1}}} + {\left( {\frac{{{x_2} - {{\hat x}_2}}}{{{l_2}}}} \right)^{2{\mu _2}}} + {\left( {\frac{{{x_3} - {{\hat x}_3}}}{{{l_3}}}} \right)^{2{\mu _3}}} - 1 $                           \\
			\hline       
			Volume-dynamic \cite{ginesi2021dynamic}
			& $G\left( {x,\dot x} \right) =  - {\nabla _x}U\left( {x,\dot x} \right)$         &$U\left( {x,\dot x} \right) = \left\{ \begin{array}{l}
				\eta_p {\left( { - \cos \vartheta } \right)^{{\mu _1}}}\frac{{\left\| {\dot x} \right\|}}{{{C^{{\mu _2}}}(x)}}{\rm{~~~if }}~~\vartheta  \in \left( {\frac{\pi }{2},\pi } \right]\\
				0{\rm{                    ~~ ~~~ ~~~  \; ~~~  ~~~    ~~~  ~~~      ~~~                        if }}~~\vartheta  \in \left[ {0,\frac{\pi }{2}} \right)
			\end{array} \right.$                            \\
			&         & $C_p(x) = {\left( {{{\left( {\frac{{{x_1} - {{\hat x}_1}}}{{{l_1}}}} \right)}^{2{\mu _1}}} + {{\left( {\frac{{{x_2} - {{\hat x}_2}}}{{{l_2}}}} \right)}^{2{\mu _1}}}} \right)^{\frac{{2{\mu _2}}}{{2{\mu _1}}}}} + {\left( {\frac{{{x_3} - {{\hat x}_3}}}{{{l_3}}}} \right)^{2{\mu _2}}} - 1 $ 
			\\ \hline
			\noalign{\hrule height 1.5pt}
		\end{tabular}%
	}
	\label{tab:DMPsobstacles}
\end{table*}

In order to enhance the generalization performance of \ac{ndsil}, various methods have incorporated a steering angle term into the \ac{ds} after training, aiming to avoid obstacles in new environment \cite{hoffmann2009biologically, hu2020fuzzy, sharma2019dmp, pairet2019learning}.
However, this Point-steering solely relies on the steering angle without considering the distance between the robot and obstacles. Consequently, it might lead to oscillatory behaviors due to the absence of distance-based adjustments. Similarly, in the Point-static method, the \ac{apf} has been employed within \ac{ndsil} of \ac{dmp} \cite{sharma2019dmp}. This method utilizes global convergence as an attraction force and calculates the repulsion force between the robot and obstacles based on potential functions. However, due to the absence of velocity information, there might be a tendency for non-smooth behaviors when encountering obstacles. Park\etal \cite{park2008movement} proposed an improved potential function that incorporates both distance and velocity information known as the Point-dynamic method. However, methods like \cite{hoffmann2009biologically} and \cite{park2008movement} are point obstacle types, necessitating the calculation of information between the robot and surface point clouds of objects, resulting in a high computation burden for large volume obstacles. Addressing obstacles as entire volumes, Ginesi\etal \cite{ginesi2019dynamic} proposed a Volume-static method, modeling obstacles as convex 3D shapes., They introduced a Volume-based potential function, improving real-time performance. Nonetheless, similar to the "Point-static method", the work in \cite{ginesi2019dynamic} does not involve velocity information, potentially encountering analogous issues. In their subsequent work, Ginesi\etal \cite{ginesi2021dynamic} proposed the Volume-dynamic method, which integrates both volumes and velocity information into the potential function. Table \ref{tab:DMPsobstacles} summarizes these five methods for reference.

In \cite{krug2013representing} and \cite{krug2015model}, Krug\etal introduced a \ac{mpc} approach for \ac{ndsil} using \ac{dmp}. This approach aims to generate predictive optimal motion plans with a planning horizon of $P$-steps. This allows for real-time updates of trajectory generation at each time step, facilitating obstacle integration through constraints within the \ac{mpc}, while adhering to spatial and temporal polyhedral constraints.
Another strategy addressing obstacles involves extending the \ac{ds} formulation with a repulsive function, similar to the one represented in \eqref{eq.DMPsavoid} \cite{hu2023model}. This extension aims to incorporate obstacle avoidance directly within the \ac{ds} formulation.

In \cite{krug2013representing} and \cite{krug2015model}, Krug\etal introduced a \ac{mpc} approach for \ac{ndsil} using \ac{dmp}. This approach aims to generate predictive optimal motion plans with a planning horizon of $P$-steps. This allows for real-time updates of trajectory generation at each time step, facilitating obstacle integration through constraints within the \ac{mpc}, while adhering to spatial and temporal polyhedral constraints. Another strategy addressing obstacles involves extending the \ac{ds} formulation with a repulsive function, similar to the one represented in \eqref{eq.DMPsavoid} \cite{hu2023model}. This extension aims to incorporate obstacle avoidance directly within the \ac{ds} formulation.

\begin{figure*}
	\def\svgwidth{\textwidth}
	{\fontsize{8}{8}\selectfont\sf
		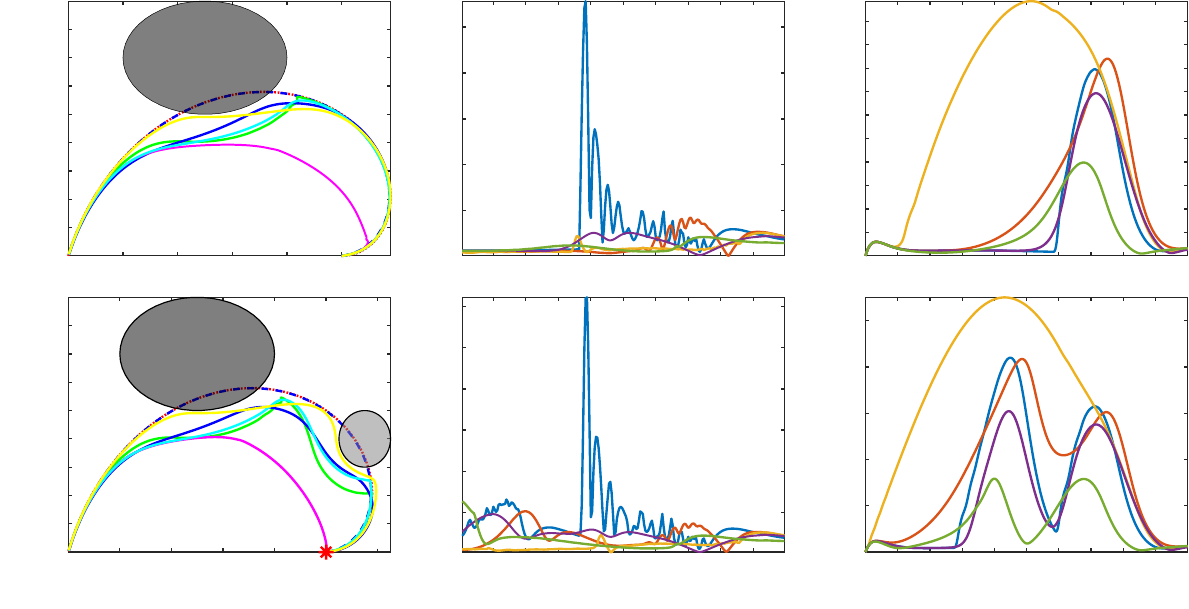}
	\caption{Different potential functions obstacle avoidance  performance of \ac{ndsil} in \ac{dmp}  \cite{ginesi2021dynamic}.}
	\label{fig.dmp_obs}
\end{figure*}

The  formulation for obstacle avoidance within \ac{ndsil} is presented as:
\begin{alignat}{2}
\ddot x{\rm{ }} = \underbrace {f\left( {x,\dot x,t} \right)}_{{\rm{attraction}}\;{\rm{force}}} + \underbrace {G\left( {x,\dot x} \right)}_{{\rm{repulsive}}\;{\rm{force}}} \label{eq.DMPsavoid}
\end{alignat}
where the attraction force $f\left( {x,\dot x,t} \right)$ embodies the trained stable \ac{ds}, akin to \eqref{eq.Transdyn1} and \eqref{eq.Transdyn2}. Meanwhile, the additional term  $G\left( {x,\dot x} \right)$ represents the repulsive force designed for obstacle avoidance. The formulation outlines five different potential functions, detailed in Table \ref{tab:DMPsobstacles}.
\change{The properties of various methods for obstacle avoidance are summarized in Table \ref{tab.obs_avoidance_characters}. The first method, Point-steering, solely calculates the steering angle for obstacle avoidance without accounting for distance, which can lead to larger errors. Additionally, the Point-steering, Point-static, and Point-dynamic methods compute the repulsion force point-by-point using a 3D point cloud of obstacles, resulting in computational burden compared to volume methods. The Point-static and Volume-static methods only support static obstacles as they do not incorporate velocity information. Acceleration and error characteristics are depicted in Fig. \ref{fig.dmp_obs}. Although the Point-static, Point-dynamic, Volume-static, and Volume-dynamic methods are not guaranteed to converge to the goal due to the possibility of local minima, a perturbation term can be easily added to push the \ac{ds} out of local minima.} The comparative performance evaluation of these methods is illustrated in Fig. \ref{fig.dmp_obs}.

\begin{table}
	\caption{\change {The properties of various methods for obstacle avoidance \cite{ginesi2021dynamic}.}}
	\rmfamily\centering
	\resizebox{\linewidth}{!}{%
		\renewcommand\arraystretch{1} 
		\centering
		\change {\begin{tabular}{|>{\arraybackslash}m{0.06\textwidth}|>{\centering\arraybackslash}m{0.06\textwidth}|>{\centering\arraybackslash}m{0.06\textwidth}|>{\centering\arraybackslash}m{0.09\textwidth}|>{\centering\arraybackslash}m{0.1\textwidth}|}
			\noalign{\hrule height 1.5pt}
			\textbf{Methods}        & \textbf{Obstacle type} & \textbf{Potential type} & \textbf{Distance  dependent} & \textbf{Guaranteed  convergence} \\ \hline
			Point-steering & point         & dynamic        & No                                                            & Yes                                                               \\ \hline
			Point-static   & point         & static         & Yes                                                           & No                                                                \\ \hline
			Point-dynamic  & point         & dynamic        & Yes                                                           & No                                                                \\ \hline
			Volume-static  & volume        & static         & Yes                                                           & No                                                                \\ \hline
			Volume-dynamic & volume        & dynamic        & Yes                                                           & No
			\\ \hline
			\noalign{\hrule height 1.5pt}
		\end{tabular}}%
	}
	\label{tab.obs_avoidance_characters}
\end{table}

Generally, non-autonomous movement representations fundamentally establish a direct relationship between a temporal signal and the dynamic attributes of the motion. The retrieval of movement from such models relies heavily on this temporal signal, which might directly signify time or employ an indirect representation through a decay term.
In addition to the \ac{dmp} methods of \ac{il}, we are exploring other classical \ac{ndsil} learning methods. These approaches offer diverse perspectives and methodologies in capturing and replicating dynamic behaviors.

In several studies \cite{calinon2007learning, calinon2010learnings, calinon2010learning, calinon2013improving, calinon2012statistical, calinon2009handling}, the problem of learning \ac{ds} has been reformulated using \ac{gmr} or \ac{hmm}. Unlike \acp{dmp}, this methodology allows for encoding multiple demonstrations.
 \begin{alignat}{2}
 \ddot x &= \sum\limits_{k = 1}^K {{h_k}(t)\left[ {K_k^p(\mu _k^x - x) - {K^v}\dot x} \right]} \label{eq.NDIL} \\
 {h_k}\left( t \right) &= \frac{{\mathcal{N}\left( {t;\mu _k^t,\Sigma _k^t} \right)}}{{\sum\nolimits_{j = 1}^K {\mathcal{N}\left( {t;\mu _j^t,\Sigma _j^t} \right)} }} \nonumber
 \end{alignat}
where $K_k^p$, $K^v$, and $\mu_k^x$ denote the full stiffness matrix, damping term, and attractor point, respectively, of the $k$-th virtual spring. Equation \eqref{eq.NDIL} shares a structural similarity with \acp{dmp}. However, it is important to note that the weight parameter $h_k$ assumes distinct interpretations in the contexts of \acp{dmp} and Eq. \eqref{eq.NDIL}. In \ac{dmp}, the determination of these weights ($h_k$)  relies on the decay term $\xi$ as defined in the system dynamics \eqref{eq.Cansys}. These weights are inherently embedded within the \ac{gmr}/\ac{hmm} representation of the motion. This alternative approach offers several advantages over \ac{dmp}: (\emph{i}) It provides enhanced flexibility in addressing spatial and temporal distortions. (\emph{ii}) Skill training and refinement are accomplished using partial demonstrations. (\emph{iii}) Learning tasks involving reaching and cyclic motions become feasible without predefined dynamics.

In \cite{calinon2011encoding}, Calinon\etal  exploited \acp{hsmm} within \ac{il} for non-autonomous systems. This focuses on integrating temporal and spatial constraints, emphasizing adaptability in the face of perturbations.
Gribovskaya\etal \cite{gribovskaya2008combining} proposed an innovative method for learning discrete bimanual coordination skills. This method integrates the automated extraction of spatio-temporal coordination constraints with a resilient motor system capable of generating coordinated movements. It operates effectively even in the presence of perturbations while adhering to learned coordination constraints.
Forte\etal \cite{forte2012line} presented a \ac{gpr} based \ac{mpg} for real-time, on-line generalization of discrete movements.
Li\etal \cite{li2023prodmp} developed a novel \ac{promp} approach. This method bridges the gap between \ac{dmp} and \ac{promp}, providing a unified framework that combines the strengths of both approaches. it enables smooth trajectory generation, goal convergence, modeling of trajectory correlations, non-linear conditioning, and online replanning with a single model. Their method demonstrates significant advantages in various robotic tasks, including reduced trajectory computation time, high-quality trajectory distribution generation, and adaptability to dynamic environments.

\subsection{Autonomous dynamical system imitation learning}
\label{sec.ASDS}

Autonomous representations model movements as \acp{ds}, capturing relationships among features like position, velocity, and acceleration independently of time. This inherent time independence grants robustness to autonomous systems, enabling them to endure disturbances that might otherwise affect the system's temporal evolution.
\ac{adsil} focuses on imitating and controlling systems that exhibit self-contained, self-evolving behavior.

Khansari\etal introduced \ac{seds} in \cite{khansari2011learning} as a method for learning stable nonlinear \acp{ds} using \ac{gmm}. \ac{seds} encapsulates classical autonomous \acp{ds}, exhibiting inherent time-invariance. It seamlessly merges machine learning principles with the Lyapunov stability theorem2, guaranteeing global asymptotic stability within \ac{il}. The incorporation of \ac{seds}  presents notable advantages in modeling a wide range of robotic motions.

The \ac{seds} model multiple demonstrations using \ac{gmm}. This encoding is represented as:
\begin{equation}
\label{eq.posterior}
\dot x = \sum\limits_{k = 1}^K {\frac{{p\left( k \right)p\left( {x|k} \right)}}{{\sum\limits_{j= 1}^K {p\left( j \right)p\left( {x|j} \right)} }}} \left( {\mu _k^{\dot x} + \Sigma _k^{\dot xx}{{\left( {\Sigma _k^x} \right)}^{ - 1}}\left( {x - \mu _k^x} \right)} \right)
\end{equation}
Equation \eqref{eq.posterior} can be reformulated as a first-order \ac{ds}:
\begin{alignat}{2}\label{eq.dymsystemGMM}
\dot x = \hat{f}\left( x \right) = \sum\limits_{k = 1}^K {{h_k}\left( x \right)\left( {{\Lambda _k}x + {d_k}} \right)} 
\end{alignat}
where:
\begin{alignat}{2}
{h_k} &= \frac{{p(k)p\left( {x|k} \right)}}{{\sum\limits_{j = 1}^K p (j)p\left( {x|j} \right)}} \label{eq.sedsweight}\\
{\Lambda_k} &= \Sigma _k^{\dot xx}{\left( {\Sigma _k^x} \right)^{ - 1}}\label{eq.sedsA}\\
{d_k} &= \mu _k^{\dot x} - {\Lambda_k}\mu _k^x \label{eq.sedsb}
\end{alignat}
Equation \eqref{eq.dymsystemGMM} presents a nonlinear combination of linear \acp{ds}. Using the Lyapunov stability theorem, a \ac{lf} can be established to derive conditions ensuring the global asymptotic stability of the system.

To tackle obstacle avoidance challenges without the need for re-teaching within the \ac{seds} framework, Khansari\etal presented a real-time obstacle avoidance method for \ac{dsil}. This approach integrates the obstacle avoidance mechanism by combining \ac{seds}/\ac{dmp} with a modulation matrix $M$ \cite{khansari2012dynamical}. The modulation matrix $M$ is adjustable, allowing for the determination of a safety margin and enhancing the robot’s responsiveness in the face of uncertainties in obstacle localization.
Although their work is tailored to scenarios involving convex obstacles, it contributes significantly to the field of autonomous systems by providing valuable insights into trajectory generation and safe navigation in complex and dynamic environments.

Apart from \ac{gmm}, other parameterized machine learning methods can be derived as \acp{ds} and integrated with control theory for \ac{il}, such as \ac{elm}.
In \cite{lemme2014neural}, Lemme\etal introduced an autonomous \ac{il} approach based on \ac{elm}. Their approach employs \ac{elm} to approximate vector fields representing \acp{ds}, incorporating stability principles derived from Lyapunov theory within predefined workspaces. The aim is to facilitate stable motion generation, particularly addressing challenges associated with sparse data and generalization.
Compared to \ac{seds}, the \ac{elm} model offers enhanced flexibility and is more trainable.
Additionally, Duan\etal \cite{duan2017fast} proposed the \ac{fsmds} approach. This method combines \ac{elm} with stability constraints, providing improved stability, accuracy, and learning efficiency in contrast to existing methods.

The \ac{seds} framework faces accuracy challenges due to an inherent conflict between accuracy and stability objectives, especially in complex and non-linear motions featuring high curvatures or deviations from attractors. This conflict arises from the constraints imposed by a \ac{qlf}, which enforces trajectories to monotonically decrease \LtwoNorm~distances \cite{neumann2015learning}.
Given the significant impact of the \ac{lf} on accuracy, one potential solution to enhance performance is to explore alternative \ac{lf}.
Khansari\etal \cite{khansari2014learning} introduced \ac{clfdm} approach ensuring global asymptotic stability in autonomous multi-dimensional \acp{ds}. 
This method learns valid \acp{lf} from demonstrations using advanced regression and optimal control techniques. \ac{clfdm} facilitates modeling complex motions and supports online learning when required. Moreover, Khansari\etal  proposed  \ac{wsaqf} parameterization to improve imitation accuracy under stable conditions.
In a related work, Jin\etal \cite{jin2023learning} presented a novel neural energy function with a unique minimum, serving as a crucial stability certificate for their demonstration learning system. This energy function is pivotal in enabling the convergence of reproduced trajectories to desired goal positions while retaining motion characteristics from the demonstrations. The study emphasizes the method's robustness against spatial disturbances, its capability to accommodate position constraints, and its effectiveness in tackling high-dimensional learning tasks. Unlike traditional methods reliant on predefined control strategies or heuristics, this approach learns an adaptable energy function, enhancing its ability to capture intricate motion patterns. Furthermore, it excels at handling position constraints, ensuring that robots operate within predefined boundaries, making it particularly valuable for safety-critical applications.

In \cite{figueroa2018physically}, Figueroa\etal presented a \ac{pcgmm} for \ac{il}. Their study extensively explores \ac{gmm} fitting, incremental learning, and the stability of merged \acp{ds}. They enhanced physical consistency by introducing a novel similarity measure based on locally-scaled cosine similarity of velocity measurements, steering trajectory clustering in alignment with linear \acp{ds}. Their approach not only outperforms Stacked End-to-End \ac{lfd} in terms of performance without relying on diffeomorphism or contraction analysis but also maintains the locality of Gaussian functions, making it suitable for recognition and incremental learning.

Jin\etal \cite{jin2019learning} introduced a novel approach that utilizes manifold submersion and immersion techniques to facilitate accurate and stable imitation of \acp{ds}.  Similar to \ac{seds}, this method relies on the Lyapunov stability theorem for \ac{ds} to establish stability conditions. By ensuring both stability and accuracy in reproducing trajectories with high-dimensional spaces, this approach presents a significant advancement in autonomous systems and \ac{il}.
In \cite{khoramshahi2018human}, Khoramshah\etal developed a parameterized \acp{ds} framework for modeling and adapting robot motions based on human interactions. The authors proposed an adaptive mechanism centered on minimizing tracking error, allowing the \ac{ds} to closely replicate human demonstrations. The study highlights the importance of hyperparameter selection and sets the foundation for future research aimed at improving the detection of human interactions, ultimately enhancing the seamlessness of human-robot collaborations.

Blocher\etal proposed the \ac{cgmr} technique \cite{blocher2017learning}. By leveraging contraction theory, \ac{cgmr} ensures stability and accuracy in generating point-to-point motions. This method employs \ac{gmm} to represent \acp{ds} and demonstrates promising results in handling complex 2D motion tasks, focusing on enhancing accuracy and optimizing training efficiency. 
Saveriano \cite{saveriano2020energy} proposed an innovative approach called \ac{esds}. This approach incorporates \acp{lf} to stabilize learned \acp{ds} at runtime, resulting in high accuracy with reduced training duration. 
\ac{esds} distinguishes itself by achieving remarkable accuracy in motion imitation while significantly reducing training times. Unlike other methods introducing substantial deformations and necessitating extensive optimization, \ac{esds} maintains fidelity in learned motions without considerable distortion. Furthermore, \ac{esds} offers a favorable balance between accuracy and training duration when compared to methods like \ac{cgmr}.

In the context of \ac{adsil}, \acp{hmm} extend their applicability beyond non-autonomous systems. Tanwani\etal \cite{tanwani2016learning} introduced task-parameterized semi-tied \acp{hsmm} for learning robot manipulation tasks. This innovative approach addresses the complexity of encoding manipulation tasks by integrating task-parameterization and semi-tied covariance matrices. Consequently, robots can autonomously adapt to various task scenarios, such as valve-turning and pick-and-place with obstacle avoidance, even in novel configurations. This work highlights the potential of \acp{hsmm} in empowering robots to acquire versatile and adaptable manipulation skills.
Zeestrate\etal \cite{zeestraten2016variable} presented an innovative approach that combines Markov chain modeling with minimal intervention control. Their approach emphasizes movement duration as a crucial aspect of skill acquisition and control. Leveraging \acp{hsmm} to represent movement variations and employing \ac{mpc}, the authors demonstrated efficacy in adapting to spatial and temporal perturbations. Moreover, their study highlights the advantages of their approach over existing methods, particularly in its versatility and capability to handle cyclic and non-cyclic behaviors.

\section{Stable \ac{dsil}} 
\label{sec.stable}
\begin{figure*}
	\def\svgwidth{\textwidth}
	{\fontsize{7}{7}\selectfont\sf
		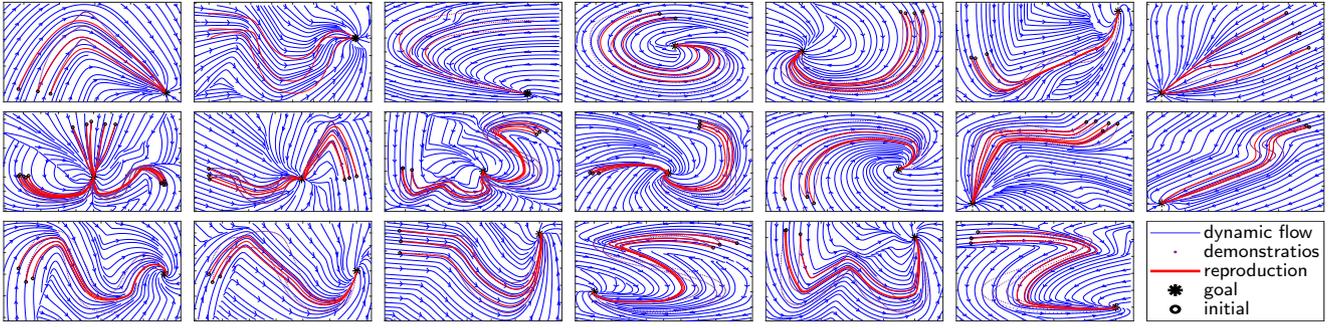}
	\caption{Imitation performance of \ac{clfdm} on 20 human handwriting motions dataset \cite{khansari2014learning}. The blue streamlines denote the dynamic flow of the energy function. The purple dashed lines represent the demonstrations, and the solid red lines are the imitation reproduction.}
	\label{fig.CLFLP_20huamnmotion}
\end{figure*}

In the context of \ac{dsil}, the stability of the system alongside its accuracy. As robots acquire motor skills through \ac{il}, ensuring robustness becomes crucial. The system must generalize effectively, maintaining convergence towards the desired behavior despite disturbances or variations.
Generally, three main methods are employed to ensure the stability of \ac{dsil}: \ac{lf} \cite{khansari2011learning}, \ac{ct} \cite{blocher2017learning}, and diffeomorphism \cite{neumann2015learning}. These methods aim to fortify the stability aspects of the learning system, mitigating deviations and disturbances that might affect the system's performance.
This section delves into these three methods, highlighting their underlying principles and their roles in enhancing the stability of \ac{dsil}.
The list of the surveyed stability methods for \ac{dsil} is tabulated in  Table \ref{tab.stability}. Additionally, the performance of one of these methods, \ac{clfdm} utilizing Lyapunov stability is illustrated in Figure \ref{fig.CLFLP_20huamnmotion}.

\subsection{Lyapunov stability}

\acp{lf} are mathematical representations that characterize the energy or potential of a \ac{ds}. In control theory, \acp{lf} are fundamental for analyzing and ensuring system stability, helping to assess a control system's convergence towards a desired behavior.
In \ac{il}, optimization techniques based on \ac{lf} involve identifying suitable functions that meet specific properties. These techniques typically use optimization methods like gradient descent, trust-region methods, or \ac{nn} training. The primary aim is to ensure stability and convergence of learned policies by demonstrating the decrease of the \ac{lf} along the system's trajectories.

In control theory, stability at an equilibrium point is desirable. 
Assuming $x^*$ is the system's attractor, stability can be represented as:
\begin{equation}
\mathop {\lim }\limits_{t \to \infty } x\left( t \right) = {x^*}:\forall x \in \Omega 
\end{equation}
This condition signifies stability, showing the system's convergence towards the desired attractor from any initial state within the system's domain $\Omega$.

For a \ac{ds} as represented in \eqref{eq.autonomous} and \eqref{eq.nonautonomous}, the stability condition can derived through the Lyapunov Stability Theorem \cite{khalil2009lyapunov}:
\begin{myTheo}
A \ac{ds} is locally asymptotically stable at
the fixed-point $x^* \in \Omega$ within the positive invariant neighborhood $\Omega \subset \mathbb{R}^d$ of $x^*$ if and only if there exists a continuous and continuously differentiable
function $V : \Omega \to  \mathbb{R}$ that satisfies the following conditions:
\end{myTheo}

\begin{figure*}
    \centering
    \subfigure{%
        \includegraphics[width=0.45\linewidth]{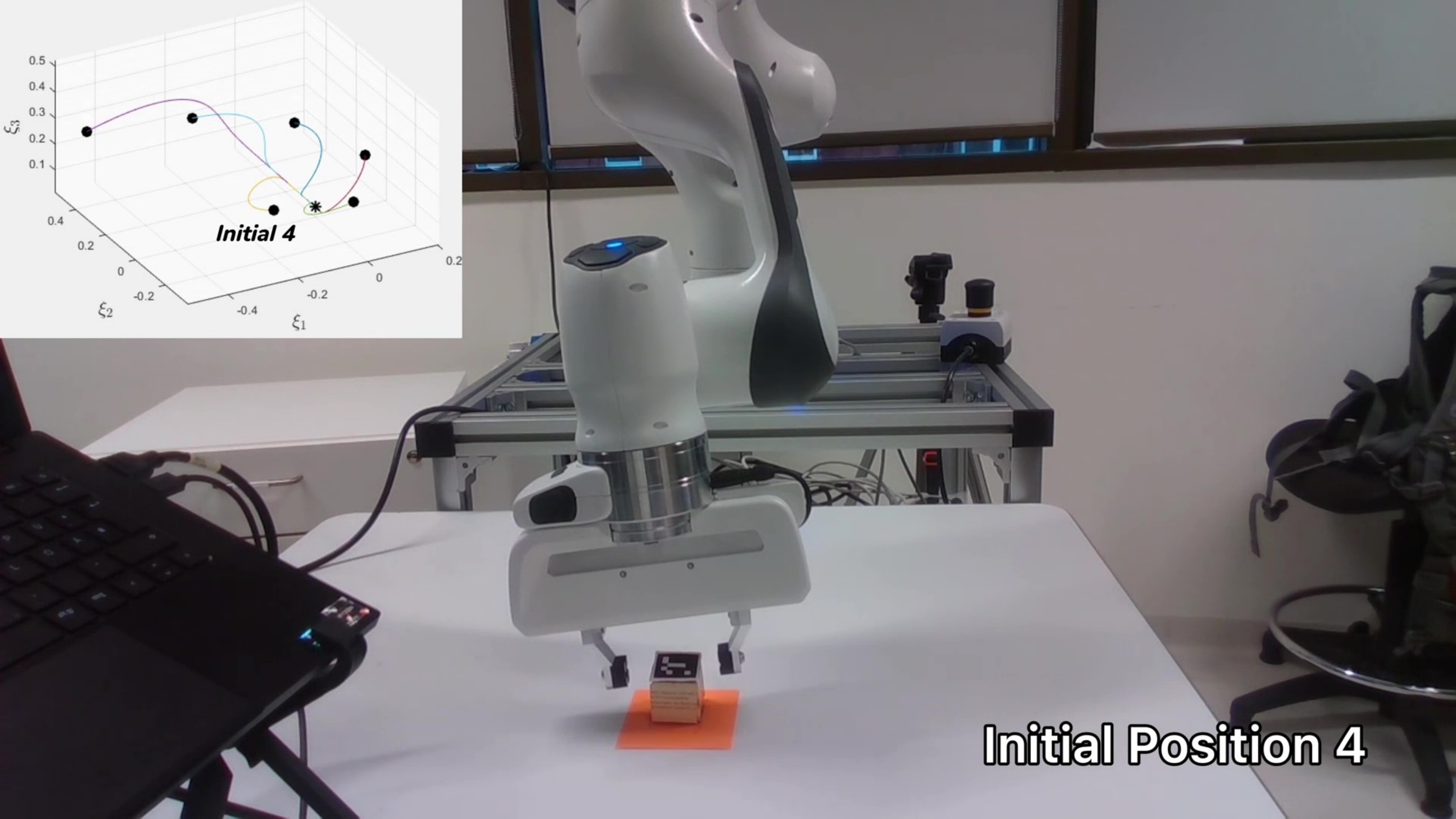}
    }
    \subfigure{%
        \includegraphics[width=0.45\linewidth]{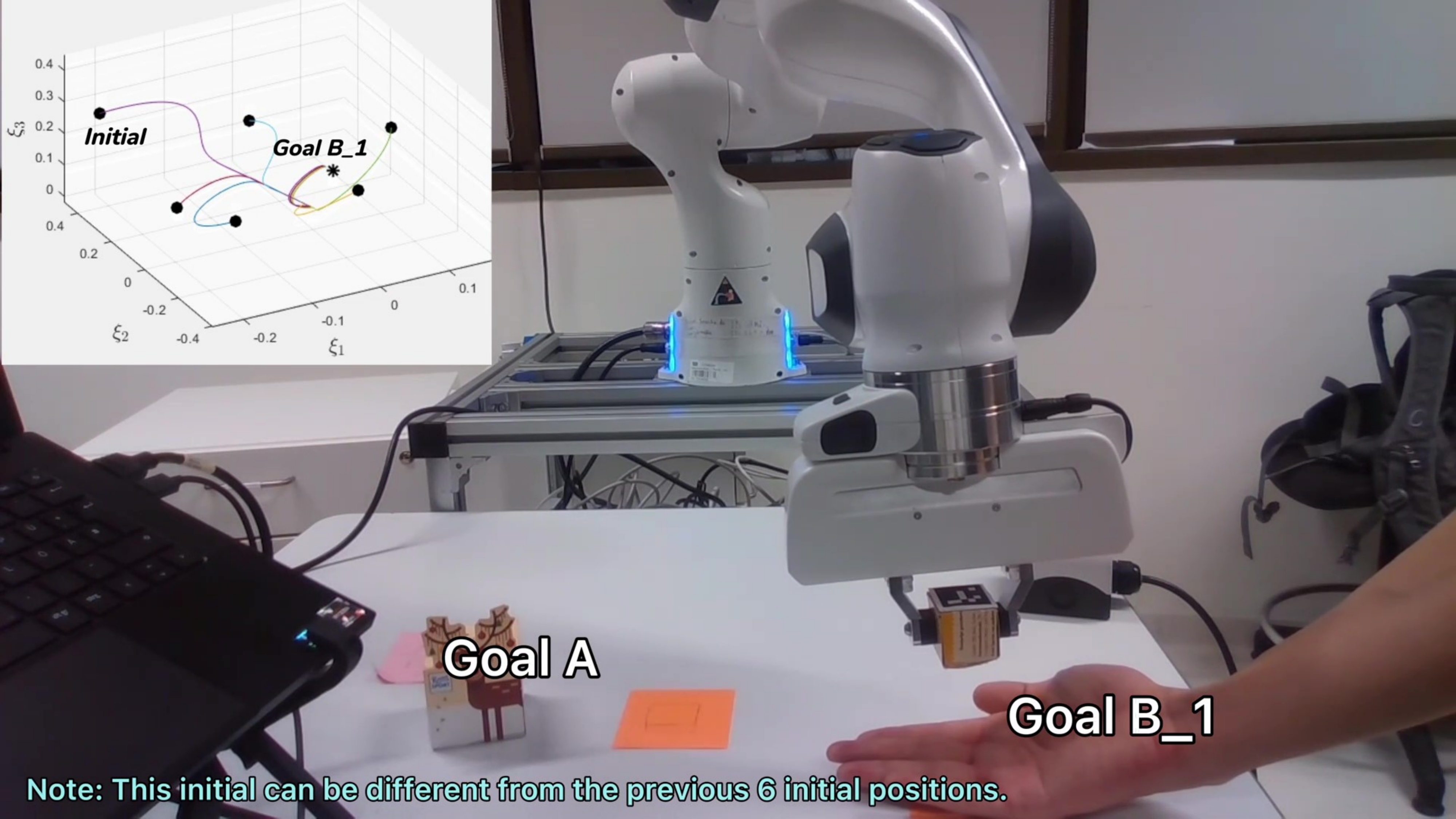}
    }
    
    \caption{The robot performs the grasping task via \ac{lfd} employing \ac{seds} \cite{hu2022robot}. \emph{Left}: the robot converges from the random initial position to the goal position.  \emph{Right}: the robot converges towards a different goal position, suggesting a switch during the task.}
    \label{fig.robotexperiment1}
\end{figure*}

\begin{equation}\label{eq.Lyapunov}
V\left(x\right)=\frac{1}{2}\left(x-x^*\right)\trsp\left(x-x^*\right)
\end{equation}
\begin{equation}\label{eq.Lyapunovconditon}
\left\{ \begin{array}{l}
V\left( x \right) > 0,\forall x \in {\mathbb{R}^d}\backslash \left\{ {{x^*}} \right\}\\ \\
\dot V\left( x \right) < 0,\forall x \in {\mathbb{R}^d}\backslash \left\{ {{x^*}} \right\}\\ \\
V\left( {{x^*}} \right)= 0~\& ~\dot V\left( {{x^*}} \right) = 0
\end{array} \right.
\end{equation}
These conditions ensure that the \ac{lf} serves a \ac{qlf}  for the autonomous \ac{ds} defined by
\begin{alignat}{2}
&\frac{{d\left( V \right)}}{{dt}} = \frac{{dV}}{{dx}}\frac{{dx}}{{dt}} \nonumber\\
& = \frac{1}{2}\frac{d}{{dx}}\left( {{{\left( {x - {x^*}} \right)}\trsp}\left( {x - {x^*}} \right)} \right)\dot x \nonumber\\
& = {\left( {x - {x^*}} \right)\trsp}\dot x \nonumber\\
& = {\left( {x - {x^*}} \right)\trsp} \cdot \hat{f}\left( x \right)
\end{alignat}
This equation demonstrates how the derivative of the \ac{lf} with respect to time is related to the dynamics of the system $\hat{f}\left( x \right)$ and the difference between the current state $x$ and the equilibrium point $x^*$.
 
The mathematical representation of the  time derivative of the \ac{qlf} in \eqref{eq.posterior}--\eqref{eq.sedsA}, and its relation to the variables and matrices involved in the \ac{ds}, is defined as in \cite{khansari2011learning, khansari2010imitation}:
\begin{alignat}{2}
\dot V& = {\left( {x - {x^*}} \right)\trsp}\sum\limits_{k = 1}^K {{h_k}\left( x \right)\left( {{\Lambda_k}x + {d_k}} \right)}  \nonumber\\
& = {\left( {x - {x^*}} \right)\trsp}\sum\limits_{k = 1}^K {{h_k}\left( x \right)\left( {{\Lambda_k}(x - {x^*}) + {\Lambda_k}{x^*} + {d_k}} \right)}  \nonumber\\
& = \sum\limits_{k = 1}^K {{h_k}\left( x \right){{\left( {x - {x^*}} \right)}\trsp}{\Lambda_k}\left( {x - {x^*}} \right)}
\end{alignat}

The unknown parameters are denoted as $\theta  = \left\{ {{A_k},{b_k}} \right\}_{k = 1}^K$.
Finally, the objective function and the sufficient condition for global asymptotic stability are presented as follows:
\begin{equation}
\begin{array}{l}
\mathop {\min }\limits_\theta \frac{1}{{2N}}\sum\limits_{ o= 1}^\mathcal{O} {\sum\limits_{n = 1}^N {\left\| {\dot x_{o, n}^{ref} - {{\dot x}_{o, n}}} \right\|^2} } \\
\rm{s.t.} \begin{array}{*{20}{c}}
{\left\{ {\begin{array}{*{20}{l}}
{{d_k} =  - {\Lambda _k}{x^*}}\\
{{\Lambda _k} + {{\left( {{\Lambda _k}} \right)}\trsp} \prec 0}\\
{{\Sigma _k} \succ 0}\\
{\sum\nolimits_{k = 1}^K {h_k = 1} ,{h_k} \in \left( {0,1} \right)}
\end{array}} \right.}
\end{array}
\end{array}
\end{equation}

The robotic grasping task employing \ac{seds} is shown in Fig. \ref{fig.robotexperiment1}.
Utilizing \ac{lf}-based optimization offers a significant advantage: it provides formal stability assurances. These guarantees ensure that the learned policy converges to the demonstrated behavior and maintains stability even in the presence of disturbances or uncertainties.
Shavit\etal \cite{shavit2018learning} proposed a method that combines joint-space \acp{ds} with task-oriented learning. This technique allows robots to adapt to various situations while maintaining stability. The approach involves leveraging dimensionality reduction methods like  \ac{pca} and \ac{kpca}  to encode activation functions and extract behavior synergies. The study demonstrates the effectiveness of this method in learning diverse behaviors, handling singular configurations, and converging to task space targets using the Lyapunov stability theorem akin to \ac{seds}.

Xu\etal \cite{xu2021learning} presented a servo control strategy utilizing \ac{bls} to achieve stable and precise trajectory imitation for micro-robotic systems. 
Their approach effectively merges \ac{bls}, which learns movement characteristics from multiple demonstrations, with Lyapunov theory to guarantee the stability of the acquired controller.
This emphasis on stability provides valuable insights into enhancing the robustness and reliability of learned control policies.

The conflict between accuracy and stability within \ac{seds}, as highlighted in Section \ref{sec.ASDS}, is mainly due to the selection of a \ac{qlf}. This function restricts trajectories to monotonically decreasing \LtwoNorm~distances from the attractor, thereby constraining \ac{seds}'s capability to manage highly nonlinear motions exhibiting high curvatures or non-monotonic behavior.
Consequently, the choice of an appropriate \ac{lf} is of paramount importance in resolving this conflict.

In  \cite{figueroa2018physically} \cite{wang2022temporal}, a method utilizing linear parameter varying-\ac{ds} learning is presented for modeling complex systems.
Through a straightforward adjustment of the \ac{gmm}'s parameters, this method is capable of outperforming \ac{seds} using standard
Lyapunov stability theory, which is without having to rely on diffeomorphism or contraction analysis.
The \ac{lf} is defined as \ac{pqlf}:
\begin{equation}\label{eq.PLF}
    V\left( x \right) = {\left( {x - {x^*}} \right)\trsp}P\left( {x - {x^*}} \right)
\end{equation}
The time derivative of $V(x)$ can be derived from \eqref{eq.posterior}--\eqref{eq.sedsA} and \eqref{eq.PLF} as:
\begin{alignat}{2}
\dot V\left( x \right) &= {\left( {x - {x^*}} \right)\trsp}\left( {\sum\limits_{k = 1}^K {{\theta _k}(x)Q} } \right)\left( {x - {x^*}} \right)
\end{alignat}
where $Q = P{\Lambda _k} + {({\Lambda _k})\trsp}P$.

The study establishes three sufficient conditions for different learning models:
\begin{enumerate}
    \item [\emph{i})] The condition for \ac{seds} method with quadratic \ac{lf}, denoted as \\ ${{\left( {{A_k}} \right)\trsp} + {A_k} \prec 0,{b_k} =  - {A_k}{x^*}}$ 
    \item [\emph{ii})] Nonconvex constraints consider an unknown matrix $P$ and the attractor at the origin, which might not satisfy all conditions due to its nonconvex nature. The conditions are \\ $Q \prec 0,{b_k} = 0,P = {P\trsp} \succ 0$
    \item [\emph{iii})] The \ac{gmm}-based linear parameter varying-\ac{ds} learning method in \cite{figueroa2018physically} and \cite{wang2022temporal}, always converging to a feasible solution under the conditions \\ $ Q\prec {H_k},{H_k} = H_k\trsp \prec 0,{b_k} =  - {A_k}{x^*}$ \\ as long as $P = P\trsp$ and possesses well-balanced eigenvalues.
\end{enumerate}

The inclusion of matrix $P$ transforms the basic \ac{qlf} into an ``elliptical'' form, allowing for trajectories that demonstrate high curvatures and non-monotonicity movement toward the target. This flexibility accommodates complex motion behaviors, allowing for more diverse and intricate paths.
Moreover, authors in \cite{wang2022temporal}, utilized \ac{ltl} specifications and sensor-based task reactivity to ensure policy stability and reliability. The utilization of invariance guarantees serves to address potential invariance failures, enhancing the approach's robustness against adversarial perturbations and execution failures. 

In \cite{khansari2014learning} \cite{gottsch2017segmentation} \cite{duan2017fast} \cite{paolillo2022learning}, a parameterized \ac{clfdm} is proposed to learn the motor skills from demonstrations to guarantee the accuracy and stability simultaneously. This approach ensures global asymptotic stability for multidimensional autonomous \acp{ds} by constructing a valid parameterized \ac{lf} through a constrained optimization process. This key aspect sets it apart from other methods that employ predefined energy functions. Furthermore, \ac{clfdm} enables the choice of the most suitable regression techniques based on task requirements, enhancing its versatility. 
Similar to the \ac{clfdm} method, authors in \cite{neumann2013neural} and  \cite{lemme2014neural}, introduced a \ac{nn}-extreme learning machine-based \ac{ds} for imitating demonstrations. This method differs from \ac{clfdm}, which acquires \ac{lf} parameters through optimization. Instead, it combines \acp{nn} with Lyapunov stability theory to guarantee stability in robot control scenarios. By directly integrating Lyapunov stability constraints into the network training process, this approach elevates the accuracy and stability of the learned dynamics, even when dealing with sparse data.
The key innovation in this work lies in the integration of stability constraints during training, which results in \acp{ds} that can generate smooth and accurate reproductions of desired motions in a three-dimensional task space. The authors emphasize the importance of finding a suitable Lyapunov candidate to ensure that the learning process aligns with stability requirements, highlighting the flexibility and robustness of the proposed approach in terms of stability and performance.
In \cite{coulombe2023generating}, Coulombe\etal
also used \acp{nn} for learning \acp{lf} and policies through \ac{il}. 
The approach discussed in the paper introduces a novel method for learning a \ac{lf} and a policy using a single \ac{nn}. The core innovation lies in satisfying the Lyapunov stability conditions, thereby ensuring that the learned policies are stable. Furthermore, the method addresses collision avoidance by incorporating a collision avoidance module, which improves the applicability of these learned policies in real-world scenarios.

In \cite{umlauft2017learning} \cite{pohler2019uncertainty}, Hirche\etal 
introduced a \ac{sosclf} for data-driven Lyapunov candidate searches by solving a convex optimization problem, significantly enhancing computational efficiency and flexibility. The approach offers a critical contribution to the stability and reliability of the Gaussian process-based \ac{ds}, which is of paramount importance in \ac{il} scenarios.

In \cite{jin2023learningss},  Jin\etal developed a \ac{nsqlf} that uses a \ac{rbfnn} to represent the \ac{qlf}.  This approach combines the power of a \ac{nsqlf} with minimal intervention control, providing an effective solution for encoding human motion skills into robotic systems. The \ac{nsqlf} offers a unique combination of a quadratic function and a \ac{rbfnn}, allowing it to fulfill the essential requirements of being a valid \ac{lf} while maintaining flexibility to capture human motion preferences in its gradient. The proposed method is implemented as a convex optimization problem, ensuring real-time applicability.
As discussed previously in Section \ref{sec.ASDS} in \cite{jin2023learning}, Jin\etal presented a concept similar to \cite{jin2023learningss}. Their work introduces a flexible neural energy function, which shares a similarity with its primary goal, aimed at ensuring globally stable and accurate demonstration learning. This approach leverages a contraction analysis framework, ensuring the stability of the \ac{ds}, and thereby enhancing its robustness in handling tasks with varying initial and goal positions.

In \cite{kolter2019learning}, Manek\etal introduced stable deep dynamics models for \ac{il}, which employ \acp{icnn} to jointly learn a convex, positive definite \ac{lf} and the associated \ac{ds}, ensuring stability throughout the state space. By incorporating stability and \acp{icnn} into deep architectures, this work paves the way for safe and reliable \ac{il} in a wide range of application domains.

In \cite{abyaneh2023learning}, Abyaneh\etal proposed the \ac{plyds} algorithm to learn a globally stable nonlinear \ac{ds} as a motion planning policy. The central concept revolves around the polynomial approximation of the policy described in Eq. \eqref{eq.autonomous} and the joint learning of a polynomial policy along with a parametric Lyapunov candidate. This joint learning approach guarantees global asymptotic stability by design, which is a crucial element in ensuring that the learned policies lead to safe and robust robotic behaviors.

In \cite{gesel2023learning}, Gesel\etal proposed the \ac{lsdiqp}, which employs iterative quadratic programming with constraint generation to optimize and ensure the stability of learned \acp{ds}.
Unlike conventional energy-based methods, \ac{lsdiqp} exhibits a unique capability: it allows the energy function to encompass not only local maximums but also saddle points. This distinctive flexibility enables LSD-IQP to acquire high reproduction accuracy and increased flexibility, including concave obstacle avoidance.

Comparisons between different Lyapunov-based methods, several variations and extensions of Lyapunov-based methods exist in the context of \ac{il}, including Lyapunov \acp{nn} and optimization methods, etc. It is essential to compare and contrast these methods to understand their strengths and limitations in different scenarios.

\subsection{Contraction theory}

\ac{ct} is a mathematical framework that plays a crucial role in ensuring the stability and robustness of learned policies in \ac{il}. It focuses on characterizing the contraction properties of \acp{ds} that determine how quickly nearby trajectories converge.

The differential relation from the \ac{ds} Eq. \eqref{eq.autonomous} or \eqref{eq.nonautonomous} is presented as:
\begin{equation}
    {{\dot \delta }_x} = \frac{{\partial f}}{{\partial x}}{\delta _x},~~J=\frac{{\partial f}}{{\partial x}}
\end{equation}
The variable of virtual displacement $\delta_x$ denotes the gap between two neighboring trajectories which are separated by an infinitesimal displacement. $J=\frac{{\partial f}}{{\partial x}}$ is the Jacobian matrix.
The squared distance of virtual displacement is $\delta _x\trsp{\delta _x}$ and the rate of change of squared distance is defined as:
\begin{equation}\label{eq.squaredis}
\frac{d}{{dt}}\left( {\delta _x\trsp{\delta _x}} \right) = 2\delta _x\trsp\frac{{\partial f}}{{\partial x}}{\dot \delta _x}=2\delta _x\trsp J_f{ \delta _x}
\end{equation}

According to contraction theory \cite{tsukamoto2021contraction}, if the symmetric part of Jacobian $J$ is uniformly negative definite,  the distance between neighboring trajectories shrinks to zero. 
Therefore, the formulation of the contraction condition is defined as:
\begin{equation}
\frac{1}{2}\left( {\frac{{\partial f}}{{\partial x}} + {{\left( {\frac{{\partial f}}{{\partial x}}} \right)}\trsp}} \right) \preceq -\lambda \left(x\right) I
\end{equation}
where $\lambda \left(x\right)>0$. Then, the Eq. \eqref{eq.squaredis} can be further rewritten as:
\begin{equation}
\frac{{\rm{d}}}{{dt}}\left( {\delta_x\trsp\delta_x} \right) \le  - 2\lambda \left( x \right)\delta_x\trsp\delta_x
\end{equation}
By path integration of both sides, it can be achieved as:
\begin{equation}
\left\| {\delta_{x_t}} \right\|_2^2 \le \left\| {\delta_{x_0}} \right\|{e^{ - \int_0\trsp {\lambda \left( x \right)dt} }}
\end{equation}
We can easily conclude that the distance between neighboring trajectories  $\delta_x$ exponentially converges to zero. The contraction theory in \ac{il} can provide robustness and stability guarantees. Contraction-based methods ensure that the learned policy remains stable even in the presence of uncertainties, disturbances, or variations in the environment.

The CT-based \ac{il} method is stated in Section \ref{sec.ASDS} of autonomous \ac{ds}.
Here, we will introduce the work \cite{blocher2017learning} in terms of stability.
This work explores the application of contraction theory to the learning of point-to-point motions through \ac{gmr}-based \acp{ds}. It introduces contracting GMR that leverages the principles of contraction analysis to achieve both stability and high-quality motion reproduction.

In \cite{sindhwani2018learning}, Sindhwani\etal presented a non-parametric framework called \ac{cvf} for learning incrementally stable \acp{ds}. Their approach combines contraction theory and kernel function to achieve stability in learned \acp{ds}. The authors utilize kernel function to efficiently model \acp{ds} by vector-valued reproducing kernel Hilbert spaces (RKHS), effectively addressing both contraction analysis and stability concerns. 
In \cite{khadir2019teleoperator}, Khadir\etal introduced a contracting vector fields approach to teleoperator imitation, which method relies on globally optimal contracting vector fields, providing continuous-time guarantees when initialized within a contraction tube around the demonstration. 
The contraction theory serves not only the purpose of stability but also extends to facilitating obstacle avoidance functions.
Huber\etal \cite{huber2019avoidance} used the contraction metrics and contraction analysis to ensure convergence and stability in the presence of convex and concave obstacles, even multiple obstacles. The contraction metric and the generalized Jacobian play a key role in evaluating and ensuring the system's stability.
Ravichandar\etal \cite{ravichandar2017learning, ravichandar2018methods, ravichandar2019learning} proposed \ac{cdsp} algorithm for learning arbitrary point-to-point motions using Gaussian mixture models. The algorithm aims to ensure stability by developing and enforcing partial contraction analysis-based constraints during the learning process. The introduction of stable \acp{ds} under diffeomorphic transformations adds to the algorithm's robustness. 
In \cite{singh2021learning}, Singh\etal introduced the concept of \ac{ccm} as a stability method, which enforces contraction conditions over continuous state spaces. The approach leverages \ac{ccm} to learn stable \acp{ds}, allowing for bounded tracking performance in various scenarios.

When describing complex contraction metrics is challenging, another approach is directly learning contraction metrics using \acp{nn}.
In \cite{tsukamoto2021imitation}, Tsukamoto\etal proposed a \ac{ncm}-based \ac{il} method, which offers real-time, safe, and optimal trajectory planning for systems dealing with disturbances.
This method combines \ac{il} and contraction theory to construct a robust feedback motion planner, particularly demonstrating its effectiveness in decentralized multi-agent settings with external disturbances.

\subsection{Diffeomorphism mapping}

Diffeomorphism is a fundamental concept in mathematics, particularly in differential geometry and topology. It represents a mathematical function that establishes a smooth and bijective mapping between two differentiable manifolds while preserving smoothness \cite{franks1971necessary} \cite{polterovich2012geometry}.

Considering an original \ac{ds} applying a diffeomorphism to the state space, the transformed system will have the same stability properties as the original system.
The significance of diffeomorphisms in stability analysis lies in their capacity to change coordinates or state variables in a manner that simplifies the analysis of system stability. 
By selecting an appropriate diffeomorphism, it is often possible to convert a complex \ac{ds} into a simpler, \ac{hsds} with well-understood stability properties. This transformation aids in simplifying the analysis of the system.

For a bijective map $\Psi$: $\mathbb{R}^n \to \mathbb{R}^n$, where $\Psi$ denotes a diffeomorphism.
According to the definition, we know that the map $\Psi$ and inverse map  $\Psi^{-1}$ are continuously differentiable. Assuming $\Psi$ is bounded, the diffeomorphism  $\Psi$: ${x} \to {\hbar}$
generate another global coordinate $\hbar \to \mathbb{R}^n$ for the manifold $\mathcal{M}$ by the map $\hbar=\Psi(x)$.
Therefore, the \ac{ds} in Eq. \eqref{eq.autonomous} or \eqref{eq.nonautonomous} can be reformulated in other coordinate $\hbar$,
\begin{alignat}{2}
\dot \hbar  &= {\left\{ {\frac{{\partial \Psi }}{{\partial x}}f\left( x \right)} \right\}_{x = {\Psi ^{ - 1}}(\hbar )}} \nonumber\\
 &= {J_\Psi }\left( {{\Psi ^{ - 1}}(\hbar )} \right)f\left( {{\Psi ^{ - 1}}(\hbar )} \right) \nonumber \\
 &: = \tilde f\left( \hbar  \right) \label{eq.diffeomorphsim}
\end{alignat}
where ${J_\Psi }\left( x \right) = \frac{{\partial \Psi }}{{\partial x}}$ denotes the Jacobian matrix.

Noted that Eq. \eqref{eq.autonomous} and  \eqref{eq.diffeomorphsim} represent the same internal \ac{ds} evolving on the manifold $\mathcal{M}$, which means they share stability properties.
Assuming there exists a \ac{lf} $V(x)$ with the equilibrium point $x^*$, which is globally asymptotically stable, a \ac{lf} $\tilde V(x)$ is obtained through the diffeomorphism $\Psi$:
\begin{alignat}{2}
\dot{\tilde{V}}\left( \hbar  \right) &= {\left\{ {\frac{{\partial V}}{{\partial x}}\frac{{\partial {\Psi ^{ - 1}}}}{{\partial \hbar }}\dot \hbar } \right\}_{x = {\Psi ^{ - 1}}(\hbar )}} \nonumber\\
 &= {\left\{ {\frac{{\partial V}}{{\partial x}}{{\left( {{J_\Psi }(x)} \right)}^{ - 1}}\tilde f\left( \hbar  \right)} \right\}_{x = {\Psi ^{ - 1}}(\hbar )}} \nonumber\\
& = {\left\{ {\frac{{\partial V}}{{\partial x}}{{\left( {{J_\Psi }(x)} \right)}^{ - 1}}{J_\Psi }(x)f\left( x \right)} \right\}_{x = {\Psi ^{ - 1}}(\hbar )}} \nonumber\\
& = {\left\{ {\frac{{\partial V}}{{\partial x}}\dot x} \right\}_{x = {\Psi ^{ - 1}}(\hbar )}} \nonumber\\
 &= \dot V\left( {{\Psi ^{ - 1}}(\hbar )} \right) \label{eq.Lyapunovdiffeomorphism}
\end{alignat}
According to Eq. \eqref{eq.Lyapunovdiffeomorphism}, the system exhibits a globally asymptotically stable equilibrium $\hbar^*=\Psi(x^*)$ after the transformation operation in Eq. \eqref{eq.diffeomorphsim}.
Meantime, according to the bijective property: If there exists a \ac{lf} $\tilde V$ under coordinate space $\hbar$, the equilibrium point $x^*$ is globally asymptotically stable \cite{rana2020euclideanizing}.

In \cite{neumann2015learning}, Neumann\etal introduced $\tau$-\ac{seds} algorithm that uses diffeomorphic transformations to expand \ac{seds}'s abilities for complex motions. This approach enables \ac{seds} to work with various Lyapunov candidates, including non-quadratic functions, allowing it to learn a wider range of robot behaviors while maintaining stability through diffeomorphic transformations.
In \cite{perrin2016fast}, Perrin\etal employed diffeomorphic matching, offering a quick and simple process for finding stable \acp{ds} that replicate observed motion patterns. The stability method involves constructing Lyapunov candidates highly compatible with the demonstrations to achieve global asymptotic stability.
Rana\etal \cite{rana2020euclideanizing} proposed the \ac{sdsef} approach, which combines a \ac{ds} with a learnable diffeomorphism to ensure global asymptotic stability. This stability method leverages a Gaussian kernel and Fourier feature approximation, requiring minimal parameter tuning.
In \cite{urain2020imitationflow}, Urain\etal introduced a deep generative model that utilizes normalizing flows to represent and learn stable \acp{sde}. The key contribution lies in the use of diffeomorphic transformations to inherit stability properties, allowing for the representation of intricate attractors like limit cycles.
Fichera\etal \cite{fichera2022linearization} introduced a graph-based spectral clustering method for stable learning of multiple attractors in a \ac{ds} using unsupervised learning. They employed a velocity-augmented kernel to capture the temporal evolution of data points in the system, allowing for the generation of a desired graph structure and the computation of the related Laplacian. Diffeomorphism learning is achieved through Laplacian eigenmaps, resulting in good accuracy and faster training times with an exponential decay in loss.
Bevanda\etal \cite{bevanda2022diffeomorphically} introduced the ``Koopmanizing Flow'' method for learning stable and accurate dynamic systems. This approach utilizes diffeomorphism to establish a connection between nonlinear systems and their linearized counterparts, ensuring the preservation of Koopman features. By applying diffeomorphism, they construct a stable Koopman operator model, which includes both linear prediction and reconstruction, resulting in both stability and accuracy.

P{\'e}rez-Dattari\etal \cite{perez2023stable, perez2023deep} introduced the convergent dynamics from demonstrations (CONDOR) method to learn stable \acp{ds}. Stability is ensured through the use of contrastive learning and regularization techniques. Specifically, the authors employ a stability loss to enforce similar output distributions for similar inputs and determine the Stability Conditions using diffeomorphism. 
Zhi\etal \cite{zhi2022diffeomorphic} proposed a \acp{dt} method for generalized \ac{il} in robotics. \acp{dt} are utilized to transform autonomous \acp{ds} while preserving asymptotic stability properties. This framework allows for flexible adaptation of robot behavior in response to environmental changes, such as adapting to obstacles for collision avoidance and incorporating user-specified biases into robot motions.

Diffeomorphism mappings can be applied not only in Euclidean space but also in manifold space to achieve stable \acp{ds} in \ac{il}.
Urain\etal \cite{urain2022learning} introduced a learnable stable vector field on Lie groups from human demonstrations for the robot system. It proposes a Motion Primitive model that employs diffeomorphism functions to ensure stability in vector field generation. The model is evaluated in various scenarios, including 2-sphere, SE(2), and SE(3) Lie groups, demonstrating its superior stability and adaptability in real-world robot tasks.
Zhang\etal \cite{zhang2022learning} presented \ac{rsds} for learning stable vector fields on Riemannian manifolds using diffeomorphisms. \ac{rsds} ensures global asymptotic stability and outperforms Euclidean flows. The paper introduces a novel methodology for computing the pull-back operator by leveraging neural \acp{ode} and diffeomorphisms. It demonstrates the effectiveness of \ac{rsds} in real-world robotic tasks with synchronized position and orientation trajectories. 
In \cite{wang2022learning} and \cite{saveriano2023learning}, Saveriano\etal learned stable \acp{ds} governing stiffness and orientation trajectories through Riemannian geometry and manifold-based approaches. Their method involves the acquisition of a diffeomorphic transformation that enables the mapping of simple stable \acp{ds} to complex robotic skills, enhancing motion stability and safety while preserving stability and adhering to geometric constraints on Riemannian manifolds.

\change{The comparison of three stability methods is shown in Table \ref{tab:stabilityfeatures}.}

\begin{table*}
	\caption{\change {Comparison of three stability methods.}}
	\rmfamily\centering
	\resizebox{\linewidth}{!}{%
		\renewcommand\arraystretch{1} 
		\centering
		\change {\begin{tabular}{|>{\arraybackslash}m{0.12\textwidth}|>{\arraybackslash}m{0.23\textwidth}|>{\arraybackslash}m{0.23\textwidth}|>{\arraybackslash}m{0.23\textwidth}|}
			\noalign{\hrule height 1.5pt}
			\multicolumn{1}{|c|}{\textbf{Category}} & \multicolumn{1}{c|}{\textbf{Advantages}}  &  \multicolumn{1}{c|}{\textbf{Disadvantages}}&  \multicolumn{1}{c|}{\textbf{Application conditions}} \\ \hline
			\textbf{Lyapunov Stability} & Rigorous mathematical framework, handles nonlinear systems.& Construction of Lyapunov functions can be challenging, limited information outside equilibrium.& Systems with known or estimable dynamics.\\ \hline
			\textbf{Contraction Theory}  &Provides insights into global stability, handles uncertainties.   & Finding contraction metrics, computational complexity for large-scale systems. & Requires background in differential geometry, limited to systems with specific geometric properties.\\ \hline
			\textbf{Diffeomorphism} & Dynamical systems are known or can be estimated.& Requires dynamical systems to exhibit contraction properties. & Dynamical systems with known geometric properties and symmetries.
			\\ \hline
			\noalign{\hrule height 1.5pt}
		\end{tabular}}%
	}
	\label{tab:stabilityfeatures}
\end{table*}
\begin{table*}
	\caption{The list of the surveyed stability methods for \ac{dsil}.}
 \rmfamily\centering
	\resizebox{\textwidth}{!}{%
		\renewcommand\arraystretch{1} 
		\centering
		\begin{tabular}{|>{\arraybackslash}m{0.28\textwidth}|>{\centering\arraybackslash}m{0.15\textwidth}|>{\centering\arraybackslash}m{0.18\textwidth}|>{\centering\arraybackslash}m{0.15\textwidth}|}
			\noalign{\hrule height 1.5pt}
			Paper&   \ac{ds}   model  & Feature&  Stability type
			\\ \noalign{\hrule height 1pt}
			Khansari\etal \cite{ khansari2011learning, khansari2010imitation}, Shavit\etal \cite{shavit2018learning}  &  \ac{gmm} & \multirow{2}{*}{\ac{qlf}} & \multirow{14}{*}{Lyapunov} 
			\\ \cline{1-2}
			Xu\etal \cite{xu2021learning} &  \ac{bls} &  & 
			\\ \cline{1-3}
			Figueroa\etal \cite{ figueroa2018physically}, Wang\etal  \cite{ wang2022temporal} &  \ac{gmm} &  Parametrized   \ac{qlf} &  
			\\ \cline{1-3}
			Khansari\etal  \cite{ khansari2014learning}, Paolillo\etal \cite{ paolillo2022learning}          & Any regression   model   & \multirow{3}{*}{WSAQF \ac{lf}}  &   
			\\ \cline{1-2}
			Duan\etal \cite{ duan2017fast} & \ac{elm} &  &   
			\\ \cline{1-2}
       		G{\"o}ttsch\etal \cite{ gottsch2017segmentation} & \ac{gmr} &  & 
       		\\ \cline{1-3}
       		Neumann\etal \cite{neumann2013neural} & \ac{gmm} & \multirow{5}{*}{Neural \ac{lf}} & 
       		\\ \cline{1-2}
       		Lemme\etal \cite{ lemme2014neural} & \ac{elm}  &  & 
       		\\ \cline{1-2}
       		Coulombe\etal \cite{ coulombe2023generating} & \ac{nn} &  & 
       		\\ \cline{1-2}
       		Jin\etal \cite{ jin2023learning}  \cite{ jin2023learningss} & \ac{gpr} &   & 
       		\\ \cline{1-2}
       		Kolter\etal \cite{ kolter2019learning} & \ac{nn} &  & 
    		\\ \cline{1-3}
    		Umlauft\etal \cite{ umlauft2017learning}, P{\"o}hler\etal \cite{ pohler2019uncertainty} & \ac{gpr} &  \multirow{2}{*}{SOS \ac{lf}}  & 
    		\\ \cline{1-2}
    		Abyaneh\etal \cite{ abyaneh2023learning} & Polynomial regression &  & 
    		\\ \cline{1-3}
    		Gesel\etal  \cite{ gesel2023learning} & Any regression & RBFNN LP & 
    		\\ \hline
    		Blocher\etal \cite{ blocher2017learning} & \ac{gmr} & \multirow{6}{*}{Constrained optimization} & \multirow{8}{*}{Contraction theory}
    		\\ \cline{1-2}
    		Sindhwani\etal \cite{ sindhwani2018learning} & Kernel function  &  & 
    		\\ \cline{1-2}
    		Khadir\etal \cite{ khadir2019teleoperator} & Polynomial regression &  &  
    		\\ \cline{1-2}
    		Ravichandar\etal \cite{ ravichandar2017learning, ravichandar2018methods, ravichandar2019learning} & \ac{gmm} &    &
    		\\ \cline{1-2}
    		Singh\etal \cite{ singh2021learning}, Khadir\etal \cite{ khadir2019teleoperator} & General \ac{ds} model &  & 
    		\\ \cline{1-3}
    		Huber\etal \cite{huber2019avoidance} & Any regression model & Dynamic modulation matrix &  
    		\\ \hline
    		Neumann\etal \cite{neumann2015learning} & \ac{seds} & \multirow{5}{*}{Euclidean}  & \multirow{8}{*}{Diffeomorphism}
    		\\ \cline{1-2}
    		Rana\etal \cite{rana2020euclideanizing}, Urain\etal  \cite{ urain2020imitationflow}, Bevanda\etal \cite{ bevanda2022diffeomorphically}, Zhi\etal \cite{ zhi2022diffeomorphic}, Perrin\etal \cite{perrin2016fast}, Fichera\etal \cite{ fichera2022linearization} & \ac{hsds} &  & 
    		\\ \cline{1-2}
    		P{\'e}rez-Dattari\etal \cite{ perez2023stable} \cite{ perez2023deep} & \ac{dnn} & &
    		\\ \cline{1-3}
    		Urain\etal \cite{ urain2022learning}, Zhang\etal \cite{ zhang2022learning},  Wang\etal \cite{ wang2022learning}, Saveriano\etal \cite{ saveriano2023learning} &  \ac{hsds} & Riemannian & 
    		\\ \hline
			\noalign{\hrule height 1.5pt}
		\end{tabular}%
	}
	\label{tab.stability}
\end{table*}

\section{Policy learning with dynamical system}
\label{sec.policy_learning}

\begin{figure*}
	\def\svgwidth{\textwidth}
	{\fontsize{8}{8}\selectfont\sf
		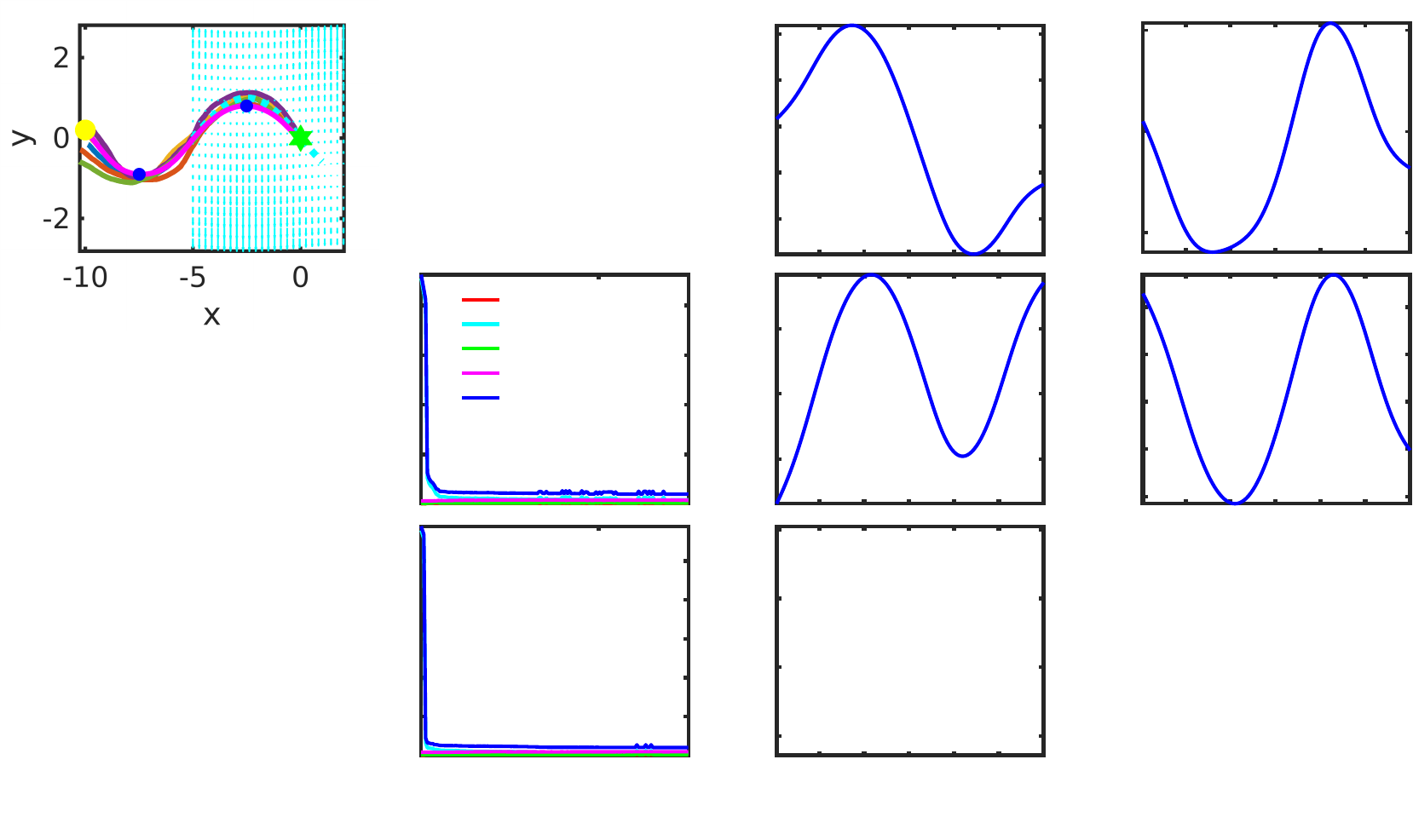}
	\caption{The \ac{rl} performance PI-ELM for sine curve \cite{hu2023pi}. The final actual trajectories (magenta), learning curves, and the learned stiffness parameters are for different tasks, which aim to go through the via points and avoid different obstacles. The cyan arrows represent the divergent force field. The cyan dashed line is the force field boundary.}
	\label{fig.RLELM1}
\end{figure*}

\begin{figure*}
	\def\svgwidth{\textwidth}
	{\fontsize{8}{8}\selectfont\sf
		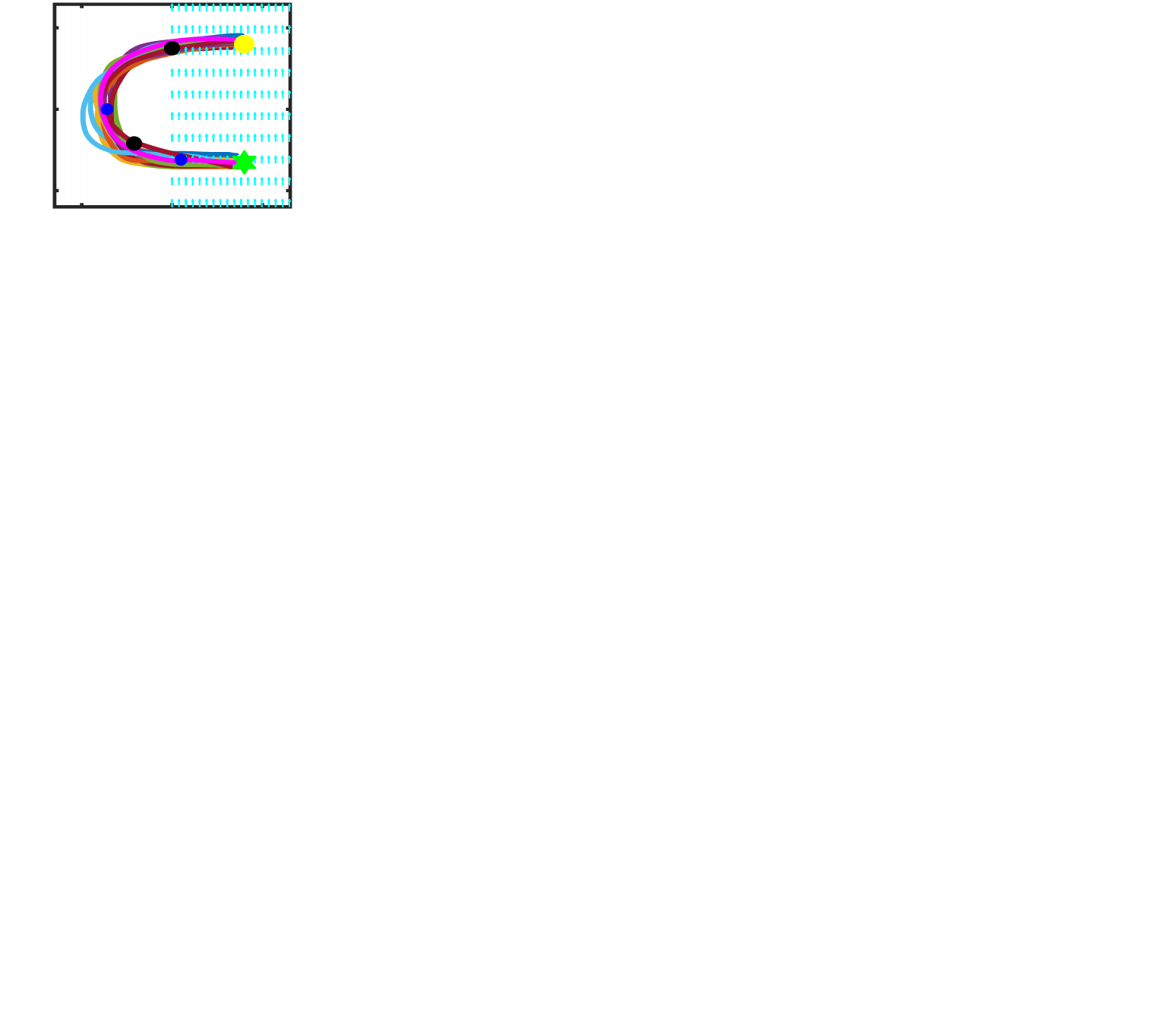}
	\caption{The \ac{rl} performance PI-ELM \cite{hu2023pi}. The final actual trajectories (magenta) and learning curves are based on the demonstrations of the human handwriting motions dataset. The cyan arrows represent the constant force field.}
	\label{fig.RLELM2}
\end{figure*}

While \ac{il} has demonstrated its capacity to generate motion \cite{khansari2011learning} and generalize to specific tasks like obstacle avoidance \cite{khansari2012dynamical} and stability \cite{huber2019avoidance} through supervised learning \cite{ijspeert2013dynamical} or unsupervised learning \cite{fichera2022linearization}, it faces challenges in generalizing when dealing with limited dataset in new environments. To enhance its adaptability and generalization capabilities, high-level \ac{rl} emerges as a promising solution, capable of addressing single or multiple tasks in novel scenarios \cite{kormushev2013reinforcement, wang2023task}. In this section, we will introduce \ac{rl} and evolution strategies for \ac{dsil}. 
The list of the surveyed policy learning methods for \ac{ds}
\ac{il} is shown in Table \ref{tab.policylearning}.
The performance of \ac{pi2} for the autonomous \ac{ds}-\ac{elm} on the human handwriting motions dataset is illustrated in Fig. \ref{fig.RLELM1} and Fig. \ref{fig.RLELM2} \cite{hu2023pi}.

\begin{table*}
	\caption{The list of the surveyed policy learning methods for \ac{dsil}}
	\resizebox{\textwidth}{!}{%
		\renewcommand\arraystretch{1} 
		\centering
		\begin{tabular}{|>{\arraybackslash}m{0.3\textwidth}|>{\centering\arraybackslash}m{0.15\textwidth}|>{\centering\arraybackslash}m{0.1\textwidth}|>{\centering\arraybackslash}m{0.1\textwidth}|>{\centering\arraybackslash}m{0.15\textwidth}|}
			\noalign{\hrule height 1.5pt}
			Paper& Policy   learning methods & Learning   type & DS   model  & Tasks/Scenario   description \\ \noalign{\hrule height 1pt}
			Theodorou\etal \cite{theodorou2010learning,theodorou2010generalized,theodorou2010reinforcement}, Stulp\etal \cite{stulp2011hierarchical,stulp2012reinforcement} & \ac{pi2} & \ac{rl} & \ac{dmp} & Trajectory \\ \hline
			Buchli\etal \cite{buchli2011variable,buchli2011learning}, Stulp\etal \cite{stulp2010reinforcement,stulp2011reinforcement,stulp2012model}, Li\etal \cite{li2017reinforcement} & \ac{pi2} & \ac{rl} & \ac{dmp} & \acs{vic} \\ \hline
			Deng\etal \cite{deng2017reinforcement}, De Andres\etal \cite{de2018reinforcement}, Beik-Mohammadi\etal  \cite{beik2020model} & \ac{pi2} & \ac{rl} & \ac{dmp} & Trajectory   of grasping \\ \hline
			Hazara\etal \cite{hazara2016reinforcement}, Colom{\'e}\etal \cite{colome2015friction}  &  \ac{pi2} & \ac{rl}  & \ac{dmp} & Physical   contact task           \\ \hline
			Yuan\etal \cite{yuan2019dmp}, Zhang\etal \cite{zhang2022motion}, Huang\etal \cite{huang2019learning} &  \ac{pi2} & \ac{rl} & \ac{dmp} & Trajectory   of exoskeleton robot \\ \hline
			Chi\etal  \cite{chi2018trajectory}, Su\etal \cite{su2020reinforcement} &  \ac{pi2} & \ac{rl} & \ac{dmp} & Trajectory of surgical robot \\ \hline
			Rey\etal \cite{rey2018learning} &  \ac{pi2} & \ac{rl} & \ac{gmm} & \acs{vic}   \\ \hline
			Hu\etal \cite{hu2023pi} &  PI-ELM & \ac{rl} & \ac{elm} & \acs{vic}   \\ \hline
			Hu\etal \cite{hu2022robot} &  \acs{xnes} & \ac{es} & \ac{gmm} & \acs{vic}   \\ \hline
			Boas\etal \cite{boas2023dmps}, Stulp\etal \cite{stulp2013robot}  &  \acs{cmaes} & \ac{es} & \ac{dmp} & Trajectory   \\ \hline
			Hu\etal \cite{hu2018evolution}  &  \acs{cmaes} & \ac{es} & \ac{dmp} & \acs{vic}   \\ \hline
			Abdolmaleki\etal \cite{abdolmaleki2017deriving}  &  \acs{trcmaes} & \ac{es} & \ac{dmp} & Trajectory   \\ \hline
			Stulp\etal \cite{stulp2012path, stulp2012adaptive, stulp2012policy}, Eteke\etal \cite{eteke2020reward}  &  \ac{pi2}-\acs{cmaes} & \ac{rl} & \ac{dmp} & Trajectory   \\ \hline
			Kim\etal \cite{kim2018learning}  &  \acs{sac} & Deep \ac{rl} & \ac{dmp} & Trajectory   \\ \hline
			Kim\etal \cite{kim2020reinforcement}  &  \acs{ddpg} & Deep \ac{rl} & \acs{nnmp} & Physical   contact task    \\ \hline
			Wang\etal \cite{wang2022adaptive}  &  \acs{sac} & Deep \ac{rl} & \ac{dmp} & Insertion   task    \\ \hline
			Chang\etal \cite{chang2022impedance}, Sun\etal \cite{sun2022integrating}  &  \acs{sac} & Deep \ac{rl} & \ac{dmp} & Physical   contact task    \\ \hline
			Davchev\etal \cite{davchev2022residual}  &  \acs{ppo}/\acs{sac} & Deep \ac{rl} & \ac{dmp} & Physical   contact task    \\ \hline
			Calinon\etal \cite{calinon2013compliant}, Kormushev\etal \cite{kormushev2010robot}, Kober\etal \cite{kober2010practical} & \acs{power} & \ac{rl} & Any/\ac{dmp} & Trajectory \\ \hline
			Andr{\'e}\etal \cite{andre2015adapting} & \acs{power}/\ac{pi2}-\acs{cmaes} & \ac{rl} & \ac{dmp} & Biped   locomotion \\ \hline
			Cho\etal \cite{cho2018relationship} & \acs{power} & \ac{rl} & \ac{dmp} & Order   task \\ \hline
			Peters\etal \cite{peters2008reinforcement} & \acs{enac} & \ac{rl} & polynomials & Trajectory \\ \hline
			Kober\etal \cite{kober2012reinforcement} & \acs{crkr} & \ac{rl} & \ac{dmp} & Trajectory \\ \hline
			Daniel\etal \cite{daniel2013autonomous} & \acs{hireps} & \ac{rl} & \ac{dmp} & Physical   contact task \\ \hline
			\noalign{\hrule height 1.5pt}
		\end{tabular}%
	}
	\label{tab.policylearning}
\end{table*}

\subsection{Dynamical system with \ac{pi2}}
\label{subsec.initialization}

As we stated in Sec \ref{sec.ndsil}, the \ac{ndsil} of \ac{dmp} provides a structured framework for encoding and reproducing complex motions, making them suitable for a wide range of \ac{il} tasks. They are defined by differential equations that represent desired trajectories and can be used to encapsulate human or expert demonstrations.
Here, we take \ac{dmp} as an example to derive the formulas of \ac{pi2}.

The \ac{dmp} equation in \eqref{eq.Transdyn1}--\eqref{eq.Gausskernel} can be reformulated as:
\begin{alignat}{2}\label{eq.DMPsrl}
\tau \ddot x& = {\alpha _z}\left( {{\beta _z}({x^*} - x) - \dot x} \right) + f_\xi ^{\rm{T}}\left( {\theta  + {\varepsilon _t}} \right) \\
{f_\xi } &= \frac{{\sum\limits_{k = 1}^K {{\varphi _k}\left( \xi  \right)\xi } }}{{\sum\limits_{k = 1}^K {{\varphi _k}\left( \xi  \right)} }} \nonumber
\end{alignat}
where the item $\varepsilon _t$ is the exploration noise; The shape vector $\theta$ is the policy parameter. Generally, the shape parameters  $\theta$ are obtained by supervised learning and then reproduce the imitation trajectory. For new tasks, the policy parameters $\theta$ should be further tuned by policy improvement of \ac{rl}.

The \ac{pi2} is a classical \ac{rl} method frequently employed in \ac{ds} learning, particularly in the context of \ac{dmp} \cite{theodorou2010learning} \cite{theodorou2010generalized}. \ac{pi2}  searches in the parameter space $\theta$ to optimize the cost function until convergence is achieved.
Generally, the cost function $J$ is designed in the following form:
\begin{alignat}{2}
\mathcal{S}\left( {{L_{n, i}}} \right) &= {\phi _{{t_N}}} + \int_{{t_n}}^{{t_N}} {\left( {{\gamma_{{t_n}}} + \frac{1}{2}u_{t_n}\trsp R{u_{t_n}}} \right)} \label{eq.costfunciton} \\
{u_{{t_n}}} &= \theta  + {M_{{t_n}}}{\varepsilon _{{t_n}}} \nonumber
\end{alignat}
where $L_{n, i}$ is the $n$-th point of the $i$-th sample trajectory; $\phi_{t_N}$ represents the terminal cost, $\gamma_{t_n}$ corresponds to the immediate cost, and $\frac{1}{2}u_{t_n}\trsp Ru_{t_n}$ is the immediate control input cost. It's important to note that the specific definitions and values of $q_{t_n}$ and $\phi_{t_N}$ are task-dependent and can be flexibly set by the users.

\begin{alignat}{2}
\varpi_{n, i} &= \frac{{\exp \left( { - \kappa \mathcal{S}\left( {{L_{n, i}}} \right)} \right)}}{{\int {\exp \left( { - \kappa \mathcal{S}\left( {{L_{n, i}}} \right)} \right)d{L _i}} }} \label{eq.prabRL}\\
\Delta {\theta _{{t_n}}} &= \int {\varpi_{n, i}{M_{{t_n}}}{\varepsilon _{{t_n}}}d{L _{n, i}}} \\
{\left[ {\Delta \theta } \right]_j} &= \frac{{\sum\nolimits_{n = 0}^{N - 1} {\left( {N - n} \right){\varphi _{j,{t_n}}}{{\left[ {\Delta {\theta _{{t_n}}}} \right]}_j}} }}{{\sum\nolimits_{n = 0}^{N - 1} {{\varphi _{j,{t_n}}}\left( {N - n} \right)} }}\\
\theta & \leftarrow \theta  + \Delta \theta  \label{eq.updatew}
\end{alignat}
where $\varpi$ is the weight of \ac{pi2} and $M_{t_n}$ is
\[{M_{{t_n}}} = \frac{{{R^{ - 1}}{f_{\xi ,}}_{{t_n}}f_{\xi ,{t_n}}\trsp}}{{f_{\xi ,{t_n}}\trsp{R^{ - 1}}{f_{\xi ,}}_{{t_n}}}}\]
The Eq. \eqref{eq.prabRL}--\eqref{eq.updatew} represents the update rule of \ac{pi2}. The searching process stops when the cost function reaches a small or sufficiently low value, indicating convergence.

In \cite{theodorou2010reinforcement}, Theodorou\etal proposed \ac{pi2} algorithm to learn the \ac{dmp} for multiple \acp{dof} via-point task, which was not recorded in demonstrations.  
The \ac{pi2} method is known for its efficiency in high-dimensional control systems, as it eliminates the need for manual learning rate adjustments typically required in gradient descent-based methods. The integration of the \ac{pi2}-\ac{dmp} framework enhances the system's ability to generalize and perform more complex tasks not covered in the demonstrations.
In \cite{stulp2011hierarchical} and \cite{stulp2012reinforcement}, Stulp\etal extend the \ac{pi2}-\ac{dmp} framework to learn both shape $\theta$ and goal parameters $x^*$ of \ac{dmp} and further proposed an extension framework for sequences of motion primitives (\ac{pi2}SEQ). 
In \cite{stulp2011hierarchical, stulp2012reinforcement}, Stulp\etal extended the  \ac{pi2}-\ac{dmp} framework to learn both the shape parameters $\theta$ and goal parameters $x^*$ of \ac{dmp}. They further introduced an extension framework for learning sequences of motion primitives known as \ac{pi2}SEQ. These extensions enhance learning capabilities and improve the robustness of tasks such as everyday pick-and-place operations.

\ac{dsil} can also encode the stiffness of \ac{vic} within a control policy for trajectory or orientation control. For instance, Michel\etal \cite{michel2023orientation} built upon the variable stiffness dynamical systems approach \cite{chen2021closed}, focusing on orientation tasks. Their study enabled a robot to follow a specified orientation motion plan governed by a first-order DS. This plan involved a customizable rotational stiffness profile, ensuring the closed-loop configuration facilitated agile and responsive robot behaviors.
In \cite{buchli2011variable, stulp2010reinforcement, stulp2011reinforcement, stulp2012model}, Stulp\etal extended the \ac{pi2}-\ac{dmp} framework to enable robots to learn \ac{vic} in complex, high-dimensional environments. This approach optimizes both reference trajectories and gain schedules, making it model-free and highly effective in learning compliant control while adhering to task constraints. Fore more details regarding learning \ac{vic} readers can refer to \cite{abu2020variable}.

Generally, the PD controller with feed-forward can be defined as:
\begin{alignat}{2}
u &=  - {K^p}\left( {q - {q_r}} \right) - {K^v}\left( {\dot q - {{\dot q}_r}} \right) + {u_{fw}}\\
{K^v} &= \rho \sqrt {{K^p}} \nonumber
\end{alignat}
where $u_{fw}$ denotes the feed-forward control; $\rho$ is a positive constant parameter.

The stiffness of \ac{vic} can be represented as follows \cite{buchli2011learning}:
\begin{alignat}{2}
{K^{p}_t} &= \Phi_{k,t}\trsp\left({\theta _k}+ \varepsilon_{k, t} \right) \label{eq.stiffKp1} \\
{\left[ {\Phi_{k,t}} \right]_j} &= \frac{{{\varphi _j}\left( \xi  \right)}}{{\sum\nolimits_{k = 1}^K {{\varphi _k}\left( \xi  \right)} }} \label{eq.stiffKp2}
\end{alignat}
where $K^p_t$ is the stiffness policy parameters of \ac{vic}ler, and $\varepsilon_{k, t}$ is the exploration noise. The equations \eqref{eq.stiffKp1} to \eqref{eq.stiffKp2} share a similar structure with the shape parameters of \ac{dmp} and follow the same learning rule of shape parameters $\theta$.

The \ac{pi2}-\ac{dmp} framework can be applied in grasping scenarios for motion generation.
In \cite{li2017reinforcement} and \cite{deng2017reinforcement}, Li\etal introduced the \ac{pi2}-\ac{dmp} framework as a planner for manipulating and grasping tasks using a humanoid-like mobile manipulator.
In \cite{de2018reinforcement}, Andres\etal proposed a hierarchical planning method to teach a 4-finger-gripper how to spin a ball. This approach utilizes high-level Q-learning for discrete actions to represent the task's abstract structure, which is then used to initialize the parameters of rhythmic \ac{dmp}. The low-level \ac{pi2}-\ac{dmp} is employed for trajectory optimization by fine-tuning policy parameters.
In \cite{beik2020model}, Beik-Mohammadi\etal proposed the \ac{rl}-\ac{dmp} framework, incorporating methods such as \ac{pi2},  \ac{power}, and \ac{enac}, for \ac{mmt} of long-distance teleoperated grasping tasks.
This research emphasizes the potential of \ac{mmt} and \ac{rl} to tackle the complexities of teleoperation, particularly under significant time delays.


The \ac{pi2}-\ac{dmp} framework can be applied for contact tasks in rigid or non-rigid environments, such as force control and compliant control.
In \cite{hazara2016reinforcement}, Hazara\etal applied  \ac{pi2}-\ac{dmp} to enhance in-contact skills of wood planing and control. The \ac{dmp} is utilized to imitate both trajectories and normal contact forces, and the imitated force policy is updated using \ac{pi2}.
In \cite{colome2015friction}, Colom{\'e}\etal used applied  \ac{pi2}-\ac{dmp} for teaching robots to perform tasks involving deformable objects in close proximity to humans, such as wrapping a scarf around the neck. The approach incorporates feedback mechanisms, including acceleration, external force estimation, and visual cues, into a comprehensive cost function.
The \ac{pi2}-\ac{dmp} framework is also versatile and finds applications in medical robotics, such as planning and control in walking exoskeletons \cite{yuan2019dmp, zhang2022motion} \cite{huang2019learning} and surgical robots \cite{chi2018trajectory, su2020reinforcement}.

Autonomous \acp{ds} can also be learned using the \ac{pi2} framework. 
In \cite{rey2018learning}, Rey\etal proposed \ac{pi2}-\ac{seds} for \ac{adsil}. This approach combines \ac{pi2} with a \ac{gmm}-based \ac{ds}, allowing robots to adapt their stiffness profiles for impedance control based on task requirements and environmental interactions. By employing a combination of cost functions, the authors demonstrate successful control for various tasks, including stable and unstable interactions. The GMM-based \ac{ds} is presented in Eq. \eqref{eq.posterior}--\eqref{eq.sedsb} and is reformulated as follows:
\begin{equation}\label{eq.dsPI2}
{{\dot x}_t} = f\left( {{x_t}} \right) + \Phi \left( {{x_t}} \right)\left( {\theta  + {\varepsilon _t}} \right)
\end{equation}
where
\[\begin{array}{l}
\theta  = {\left[ {{\theta _1},{\theta _2}, \ldots ,{\theta _K}} \right]\trsp}\\
{\Phi _k}\left( {{x_t}} \right) = {h_k}\left[ {\begin{array}{*{20}{c}}
{{x_{1t}}}&{{x_{2t}}}&0&0&1&0\\
0&0&{{x_{1t}}}&{{x_{2t}}}&0&1
\end{array}} \right]\\
{\theta _k} = \left[ {{\Lambda _{k,11}},{\Lambda _{k,12}},{\Lambda _{k,21}},{\Lambda _{k,22}},{d_{k,1}},{d_{k,2}}} \right]
\end{array}\]
where the equation \eqref{eq.dsPI2} represents a general stochastic dynamic system; $\varepsilon_t$ is the exploration noise, and $\theta$ represents the policy parameter.
Therefore, as with the \ac{pi2}-\ac{dmp} learning framework, \ac{pi2} can also be applied to the equation \eqref{eq.dsPI2}. 
 In \cite{hu2023pi}, Hu\etal replaced the \ac{gmm} with an \ac{elm} and introduced the modified \ac{pi2} algorithm called $\rm{PI}$-\ac{elm}. This approach retains the \ac{pi2} framework and incorporates the natural gradient of natural evolution strategies for learning the policy parameters of \ac{elm}. The proposed $\rm{PI}$-\ac{elm} method surpasses \ac{pi2}-\ac{seds} in terms of faster searching speed, lower cost value (accuracy), and reduced running time.

\begin{figure*}
\begin{adjustbox}{varwidth=\textwidth,fbox,center}
  \centering
  \begin{minipage}[t]{0.48\textwidth}
     \removelatexerror
     \begin{algorithm}[H]\label{alg.NES}
     {
     \KwIn{$\Sigma_\theta=C\trsp C$,~ $\theta  = \left\langle {\mu_\theta, \Sigma_\theta} \right \rangle $}
     \caption{NES}
     \While{until convergence}{
        \For{i=1 $\cdots$ $\mathcal{I}$}{
          Sampling: $\varepsilon_i \sim {\cal N}\left(0, I  \right) $;\\
          $x_i=\mu_\theta+ C\trsp \varepsilon_i$;\\
          Calculate the cost function $\mathcal{S}(x_i)$;\\
          \textcolor{green}{Calculate the log-derivatives: ${\nabla _\theta }\log p\left( {x_i|\theta } \right)$};
          }
       Calculate policy gradient: \\
       \textcolor{red}{${\nabla _\theta }\mathbb{J}\left( \theta  \right) \approx \frac{1}{\mathcal{I}}\sum\limits_{i = 1}^\mathcal{I} {\bar{\mathcal{S}}\left( {{x _i}} \right){\nabla _\theta }\log p\left( {{x _i}|\theta } \right)} $;}\\
         Update policy parameters: \\
         \textcolor{purple}{$\theta  \leftarrow \theta  + \eta_l \sum\limits_{i = 1}^\mathcal{I} {\frac{{\overline {\cal J} \left( {{x_i}} \right)}}{\mathcal{I}}{\nabla _\theta }\log p\left( {{x_i}|\theta } \right)}$ (\ac{vpg});}\\
         or\\
        \textcolor{red}{$\theta  \leftarrow \theta  + \eta_l \sum\limits_{i = 1}^\mathcal{I} {\frac{{\overline {\cal J} \left( {{x_i}} \right)}}{\mathcal{I}}{F^{ - 1}}{\nabla _\theta }\log p\left( {{x_i}|\theta } \right)}$ (natural gradient);}
         }
    }
    \end{algorithm}
  \end{minipage}%
  \hspace{0.05cm}
  \begin{minipage}[t]{0.48\textwidth}
    \vspace*{\dimexpr-\baselineskip+\topskip} 
   \removelatexerror
   \begin{algorithm}[H]\label{alg.xNES}
    {
    \KwIn{$\Sigma_\theta=C\trsp C$,~ $\theta  = \left\langle {\mu_\theta, \Sigma_\theta} \right \rangle $, $C'=C/\sigma$}
    \caption{xNES}
     \While{until convergence}{
        \For{i=1 $\cdots$ $\mathcal{I}$}{
          Sampling: $\varepsilon_i \sim {\cal N}\left(0, I  \right) $;\\
          $x_i=\mu_\theta+ \sigma {C'}\trsp \varepsilon_i$;\\
          Calculate the cost function $\mathcal{S}(x_i)$;      
          }
      \textcolor{green}{ Sort $\left\{ {\left( {{\varepsilon _i},{x_i}} \right)} \right\}$ w.r.t $\mathcal{S}$ and calculate utilities $\nu$;}\\
       Calculate gradient:\\
       \textcolor{red}{${\nabla _\delta } \mathbb{J}= \sum\limits_{i = 1}^\mathcal{I} {{\nu_i}{\varepsilon _i}}$;\\
       ${\nabla _M}\mathbb{J} = \sum\limits_{i = 1}^\mathcal{I} {{\nu_i}\left( {{\varepsilon _i}\varepsilon _i\trsp - I} \right)}$;\\   
       ${\nabla _\sigma }\mathbb{J} = {{{\rm{tr}}\left( {{\nabla _M}} \right)} \mathord{\left/
     {\vphantom {{{\rm{tr}}\left( {{\nabla _M}\mathbb{J}} \right)} d}} \right.} d}$;\\
     ${\nabla _{C'}}\mathbb{J} = {\nabla _M}\mathbb{J} - {\nabla _\sigma }\mathbb{J} \cdot I$;}\\
     Update parameters:\\
     \textcolor{red}{${\mu _\theta } \leftarrow {\mu _\theta } + {\eta _\delta } \cdot \sigma C' \cdot {\nabla _\delta }$;\\
     $\sigma  \leftarrow \sigma  \cdot \exp \left( {{{{\eta _\delta }} \mathord{\left/
     {\vphantom {{{\eta _\delta }} 2}} \right.} 2} \cdot {\nabla _\sigma }} \right)$;\\
     $C' \leftarrow C' \cdot \exp \left( {{{{\eta _{C'}}} \mathord{\left/
     {\vphantom {{{\eta _{C'}}} 2}} \right.} 2} \cdot {\nabla _{C'}}} \right)$;}
         }
    }
    \end{algorithm}
  \end{minipage}
 \\ 
  \begin{minipage}[t]{0.48\textwidth}
    \vspace*{\dimexpr-\baselineskip+\topskip} 
   \removelatexerror
   \begin{algorithm}[H]\label{alg.CMA-ES}
    {\KwIn{$\Sigma_\theta=C\trsp C$,~ $\theta  = \left\langle {\mu_\theta, \Sigma_\theta} \right \rangle $, $C'=C/\sigma$}
    \caption{CMA-ES}
     \While{until convergence}{
        \For{i=1 $\cdots$ $\mathcal{I}$}{
          Sampling: $\varepsilon_i \sim {\cal N}\left(0, I  \right) $;\\
          $x_i=\mu_\theta+ \sigma {C'}\trsp \varepsilon_i$;\\
          Calculate the cost function $\mathcal{S}(x_i)$;
          }
           \textcolor{green}{Sort $\left\{ {{{\varepsilon _i}} } \right\}$ w.r.t $\mathcal{S}$ and calculate utilities $\varpi$;\\
         \For{i=1 $\cdots$ $\mathcal{I}$}{
           ${\varpi_i} = \left\{ {\begin{array}{*{20}{l}}
    {\ln \left( {\max (\mathcal{I}/2, \mathcal{I}_e) + 0.5} \right) - \ln \left( i \right),\;{\rm{if}}\;{i} \le {\mathcal{I}_e}}\\
    {0,\;{\rm{if}}\;i > \mathcal{I}_e}
    \end{array}} \right.$
          }}
         Update mean:\\
         \textcolor{red}{$\Delta \mu_\theta  = \sum\nolimits_{i = 1}^\mathcal{I} {{\varpi_i}{\varepsilon _i}} $;\\
          ${\mu_\theta  \leftarrow \mu_\theta  + \Delta \mu_\theta }$;}\\
         Update covariance matrix:\\
         \textcolor{red}{ ${{p_\sigma } \leftarrow  ( {1 - {c_\sigma }} ){p_\sigma } + \sqrt {{c_\sigma }\left( {2 - {c_\sigma }} \right){\mu _P}{\Sigma ^{ - 1}}} \frac{{\Delta \mu_\theta }}{\sigma }}$;\\
         ${\sigma  \leftarrow \sigma  \times \exp \left( {\frac{{{c_\sigma }}}{{{d_\sigma }}}\left( {\frac{{\left\| {{p_\sigma }} \right\|}}{{E\left\| {N(0,I)} \right\|}} - 1} \right)} \right)}$;\\
         ${{p_\Sigma } \leftarrow \left( {1 - {c_\Sigma }} \right){p_\Sigma } + {\eta _\sigma }\sqrt {{c_\Sigma }\left( {2 - {c_\Sigma }} \right){\mu _P}} \frac{{\Delta \mu_\theta }}{\sigma }}$;\\
         $\Sigma_\theta  \leftarrow  \left( {1 - {c_1} - {c_\mu }} \right)\Sigma  + {c_1}\left( {{p_\Sigma }p_\Sigma\trsp + \Delta {\eta _\sigma }\Sigma } \right)$
         $+ {c_\mu }\sum\nolimits_{i = 1}^\mathcal{I} {{P_i}{\varepsilon _i}} \varepsilon _i\trsp$;}
      }}
    \end{algorithm}
  \end{minipage}
    \begin{minipage}[t]{0.48\textwidth}
    \vspace*{\dimexpr-\baselineskip+\topskip} 
   \removelatexerror
   \begin{algorithm}[H]\label{alg.PI2}
     {\KwIn{$\Sigma_\theta=C\trsp C$,~ $\theta  : = \left\langle {\mu_\theta} \right \rangle $}
\caption{\ac{pi2}}
 \While{until convergence}{
    \For{i=1 $\cdots$ $\mathcal{I}$}{
        \textcolor{blue}{
         \For{n=1 $\cdots$ N}{
          Sampling: $\varepsilon_{n,i} \sim {\cal N}\left(0, I  \right) $;\\
          $x_{n, i}=\mu_\theta+ C\trsp \varepsilon_{n, i}$;}
        \For{n=1 $\cdots$ N}{
          Calculate the cost function ${\cal{S}}(x_{n, i})$;} }
      }
      
      \For{n=1 $\cdots$ N}{
      
        \textcolor{green}{
         \For{i=1 $\cdots$ $\mathcal{I}$}{
          Compute weight: ${\varpi _{n, i}} = \frac{{\exp \left( { - \kappa {\cal S}\left( {{x_{n, i}}} \right)} \right)}}{{\sum\nolimits_{n = 1}^\mathcal{I} {\exp \left( { - \kappa {\cal S}\left( {{x_{n, i}}} \right)} \right)} }}$;
          }}
          \textcolor{red}{
          $\Delta {\mu _{{\theta _n}}} = {M_{{t_n}}}\sum\nolimits_{i = 1}^\mathcal{I} {{\varpi _{n, i}}{\varepsilon _{n, i}}} $;}
      }
       Update parameters: \\
       \textcolor{red}{
      ${\left[ {\Delta {\mu _\theta }} \right]_j} = \frac{{\sum\nolimits_{n = 0}^N {\left( {N - n} \right){\varphi _{j,n}}{{\left[ {\Delta {\mu _{{\theta _i}}}} \right]}_j}} }}{{\sum\nolimits_{n = 0}^N {\left( {N - n} \right){\varphi _{j,n}}} }}$;\\
     ${\mu _\theta } \leftarrow {\mu _\theta } + \Delta {\mu _\theta }$;\\
     (${\theta  \leftarrow \theta  + \Delta \theta }$)}
  }
}
    \end{algorithm}
  \end{minipage}
  \end{adjustbox}
  \caption{\change{Connection: policy improvement methods from evolution strategies to \ac{pi2}. \ac{vpg} ,\ac{es}, and \ac{pi2} all utilize parameter space exploration, \ac{vpg} is directly depend on policy gradient, \ac{nes} and \ac{nes} use natrual gradient,  \ac{pi2} and \ac{cmaes} update policy parameters through reward-weighted averaging. They have distinct characteristics highlighted with different colors.} }
  \label{fig.connection}
\end{figure*}

\begin{table*}
	\caption{\change {Comparison of different policy learning methods.}}
	\rmfamily\centering
	\resizebox{\linewidth}{!}{%
		\renewcommand\arraystretch{1} 
		\centering
		\change {\begin{tabular}{|>{\arraybackslash}m{0.08\linewidth}|>{\arraybackslash}m{0.2\linewidth}|>{\arraybackslash}m{0.2\linewidth}|>{\arraybackslash}m{0.2\linewidth}|>{\arraybackslash}m{0.2\linewidth}|}
				\noalign{\hrule height 1.5pt}
				\multicolumn{1}{|c|}{\textbf{Method}} & \multicolumn{1}{c|}{\textbf{Differences}} & \multicolumn{1}{c|}{\textbf{Advantages}} & \multicolumn{1}{c|}{\textbf{Limitations}} & \multicolumn{1}{c|}{\textbf{Application conditions}} \\
				\hline
				\ac{vpg} & Simplest form of policy gradient methods, update policy directly from policy gradients. & Simple to implement and computationally efficient. & Prone to high variance and slow convergence, requires careful tuning of learning rates. & Suitable for simple environments with low-dimensional action spaces.
				\\ \hline
				\ac{nes}/\acs{xnes} & Utilizes natural gradient to optimize the policy parameters. & Less sensitive to hyperparameters compared to other methods, and handles noisy gradients well. \ac{xnes} avoids computing the inverse Fisher information matrix via a special trick, making it more principled than \ac{cmaes}, as all update rules stem from a single principle. & May struggle with high-dimensional action spaces and complex environments, and can be sensitive to hyperparameters. & Suitable for optimization problems with large-scale, continuous action spaces. 
                \\
				\hline
				\ac{cmaes} & Uses reward-weighted averaging (rather than gradient estimation) to update the policy parameters. Maintains a covariance matrix to adapt the search distribution and updates it iteratively. & Efficient for optimizing non-linear, non-convex functions, handles noisy gradients, and global optimization. & Requires more computational resources due to its iterative nature and may struggle with stochastic environments. & Suitable for optimization problems with high-dimensional, non-linear, and non-convex objective functions. 
                \\
				\hline
				\ac{pi2} & Utilizes path integral formulation (a reward-weighted averaging method) to optimize the policy parameters directly. & Effective for optimizing high-dimensional, non-linear policies, and handles stochastic environments. & Sensitive to choice of kernel function and can be computationally expensive for large action spaces. & Suitable for problems with continuous, multi-modal action spaces and complex environments with uncertainty. 
                \\
				\hline
				\noalign{\hrule height 1.5pt}
		\end{tabular}}%
	}
	\label{tab:optimization_methods}
\end{table*}


\subsection{Dynamical system with evolutionary strategies}

\ac{es} methods are known for their ability to explore a wide space of policies to find effective solutions. They integrate stochastic search techniques inspired by natural evolution, allowing them to discover effective policies through the exploration of a population of policies.

We first introduce a classical member of the \ac{es} family method-\ac{nes} \cite{wierstra2014natural} for policy learning. 
Considering a general \ac{ds} in Eq. \eqref{eq.dsPI2} and cost function $\mathcal{S}$ in Eq. \eqref{eq.costfunciton}.
The policy parameter is represented as $\theta = \left\langle {\mu_\theta, \Sigma_\theta} \right \rangle$ with $C\trsp C=\Sigma_\theta$ and $\sigma = \sqrt[d]{{\left| {\det C} \right|}}$, enabling the expression of $x=\mu_\theta+C \varepsilon$ ($\varepsilon \sim {\cal N}\left( {0,I} \right)$).
To compute the policy gradient with respect to the density function $p\left( {x|\theta } \right)$, the search distribution of the expected cost function $\mathcal {\bar S} =-\mathcal{S}$ (for gradient ascent) in Eq. \eqref{eq.costfunciton} is defined as:
 \begin{alignat}{2}
& \mathbb{J}\left( \theta  \right) = \int {\bar{\mathcal{S}}\left( x \right)p\left( {x |\theta } \right)d} x \\
&p\left( {x |\theta } \right) = \frac{1}{{{{\left( {2\pi } \right)}^{m/2}}\det \left( C \right)}}\exp \left( { - \frac{1}{2}{{\left\| {{C^{ - 1}}\left( {x  - \mu_\theta } \right)} \right\|}^2}} \right) \nonumber
\end{alignat}

The policy gradient of $\mathbb{J}\left( \theta  \right)$ is:
\begin{alignat}{2} 
{\nabla _\theta }\mathbb{J}\left( \theta  \right) &= {\nabla _\theta }\int {\bar{\mathcal{S}}\left( x  \right)p\left( {x |\theta } \right)d} x \nonumber\\
 &= \int {\bar{\mathcal{S}}\left( x  \right){\nabla _\theta }p\left( {x |\theta } \right)d} x  \nonumber\\
& = \int {\bar{\mathcal{S}}\left( x  \right){\nabla _\theta }p\left( {x |\theta } \right)\frac{{p\left( {x |\theta } \right)}}{{p\left( {x |\theta } \right)}}d} x   \nonumber\\
& = \int {\bar{\mathcal{S}}\left( x  \right){\nabla _\theta }\log p\left( {x |\theta } \right)p\left( {x |\theta } \right)d} x   \nonumber\\
& = {\mathbb E}\left[ {\bar{\mathcal{S}}\left( x  \right){\nabla _\theta }\log p\left( {x |\theta } \right)} \right] \label{eq.logliketrick}
\end{alignat}

The update rule is, 
\change{
\begin{alignat}{2}
\theta  &\leftarrow \theta  + \eta_l \nabla \mathbb{J}\left( \theta  \right) ~~(\ac{vpg}) \label{eq.policygradient}\\
\theta & \leftarrow \theta  + \eta_l {F^{ - 1}}\nabla \mathbb{J}\left( \theta  \right)   ~~(\rm{natural~ grident}) \label{eq.naturalgradient}
\end{alignat}
}
where $F$ is the fisher matrix; $ {F^{ - 1}}\nabla \mathbb{J}\left( \theta  \right)$ is the natural gradient. According to the Monte Carlo estimation, the function in Eq. \eqref{eq.logliketrick} can be approximated as,
\begin{alignat}{2}\label{eq.Monte}
{\nabla _\theta }\mathbb{J}\left( \theta  \right) \approx \frac{1}{\mathcal{I}}\sum\limits_{i = 1}^\mathcal{I} {\bar{\mathcal{S}}\left( {{x _i}} \right){\nabla _\theta }\log p\left( {{x _i}|\theta } \right)} 
\end{alignat}
Substituting Eq. \eqref{eq.Monte}  into Eq. \ref{eq.policygradient}--\eqref{eq.naturalgradient}:
\begin{alignat}{2}
&\theta  \leftarrow \theta  + \eta_l \sum\limits_{i = 1}^\mathcal{I} {\frac{{\bar {\cal S} \left( {{x_i}} \right)}}{\mathcal{I}}{\nabla _\theta }\log p\left( {{x_i}|\theta } \right)} \label{eq.policygradientfinal}\\
&\theta  \leftarrow \theta  + \eta_l \sum\limits_{i = 1}^\mathcal{I} {\frac{{\bar {\cal S} \left( {{x_i}} \right)}}{\mathcal{I}}{F^{ - 1}}{\nabla _\theta }\log p\left( {{x_i}|\theta } \right)}\label{eq.naturalgradientfinal}
\end{alignat}

\change{The difference between \ac{vpg}} and natural policy gradient method are shown in Eq. \eqref{eq.policygradient}--\eqref{eq.naturalgradient}
and Eq. \eqref{eq.policygradientfinal}--\eqref{eq.naturalgradientfinal}.

\begin{figure*}
    \centering
     \includegraphics[width=0.99\linewidth]{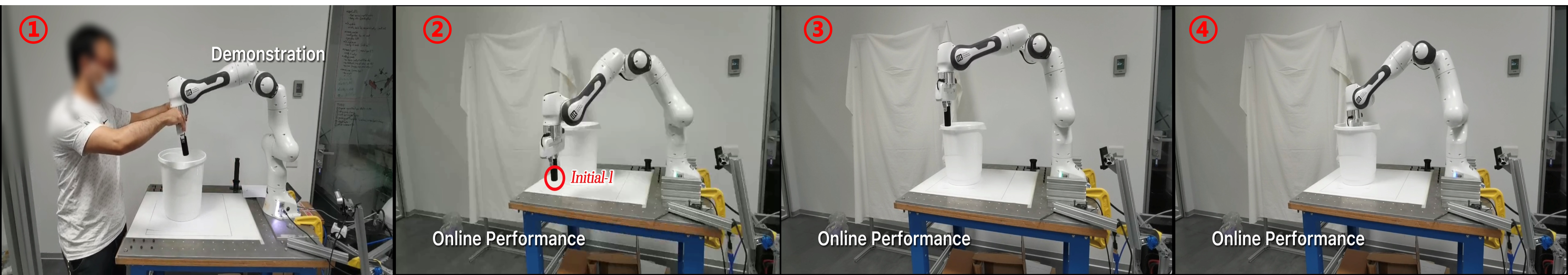}
    \caption{\ac{nes} learning for the task of reaching goal locations with obstacles \cite{hu2022robot}.}
    \label{fig:NESRL}
\end{figure*}

In \cite{wierstra2014natural}, Wierstra\etal introduced the \ac{xnes} method, which is a modified version of \ac{nes}.
Compared to \ac{nes}, the key difference lies in the replacement of the Fisher matrix and cost function in Eq. \eqref{eq.naturalgradientfinal} with utility function:
\begin{equation}
{\nabla _\theta } \tilde{\mathbb{J}}= \sum\limits_{i = 1}^\mathcal{I} {{\nu_i}{\nabla _\theta }\log p\left( {{x_i}|\theta } \right)} 
\end{equation}
where $\nu_i$ is the utility function. The \ac{xnes} can reduce the burden of computation of the Fisher matrix because it replaces the computation in another coordinate system with a linear transformation.
In \cite{hu2022robot}, Hu\etal applied \ac{xnes} for \ac{gmm}-based \ac{ds} learning at the high level. 
The paper presents experiments showcasing the effectiveness of their approach in enabling robots to avoid obstacles, counter external disturbances, and adapt to different initial and goal positions (Fig. \ref{fig:NESRL}).

The \ac{cmaes} serves a role similar to \ac{pi2}, \ac{nes} and \ac{xnes} for learning \acp{ds}, as seen in \ac{cmaes}-\ac{dmp} \cite{boas2023dmps} \cite{stulp2013robot}. 
In \cite{hu2018evolution}, Hu\etal introduced \ac{cmaes}-\ac{dmp} for learning adaptive \ac{vic} in robotics, particularly focusing on grasping tasks under uncertain conditions. 
This approach combines \ac{es} with \ac{dmp} to optimize policy parameters, including not only the mean but also the covariance parameters, which are used to modify the exploration noise in the parameter space.

In \cite{abdolmaleki2017deriving}, Abdolmaleki\etal proposed an enhanced \ac{trcmaes} algorithm, which optimizes mean and covariance updates using an expectation-maximization framework with information-geometric trust regions. Unlike the standard \ac{cmaes}, \ac{trcmaes} utilizes trust regions to adaptively adjust Lagrangian multipliers and step sizes during optimization. The \ac{trcmaes} method exhibits stronger robustness in finding solutions and faster convergence speed compared to standard \ac{cmaes}.

In \cite{stulp2012path} \cite{stulp2012adaptive},  Stulp\etal established a connection between the \ac{pi2} and \ac{cmaes} algorithms and proposed the \ac{pi2}-CMA algorithm to optimize policy parameters for \ac{dmp}. \ac{pi2}-CMA combines the structural benefits of both \ac{pi2} and \ac{cmaes}, reducing the need for manual parameter tuning in \ac{rl} tasks. In \cite{stulp2012policy}, Stulp\etal summarized the commonalities and differences between \ac{pi2}, \ac{cmaes}, and \ac{pi2}-CMA, along with various variants of \ac{pi2}-CMA, and reported their performance in different tasks.
In \cite{eteke2020reward}, Eteke\etal enhanced the \ac{pi2}-CMA method and proposed the \ac{pi2}-ES-Cov approach. This method combines a black-box strategy with an adaptive exploration strategy for policy search in \ac{dmp}. In comparison to \ac{pi2}-CMA, the $\sigma$ parameter does not require decay, eliminating the need for adjusting additional parameters. Additionally, the exploration adapts to the robot's performance, facilitating faster convergence in learning.

The connection between policy gradient, \ac{nes}, \ac{xnes},  \ac{cmaes} and \ac{pi2} is depicted in Fig. \ref{fig.connection}, highlighting the similarities and differences among these methods. 
\change{\ac{nes}, \ac{xnes}, and \ac{cmaes} belong to a family of \ac{es} algorithms, which iteratively update a search distribution using an estimated gradient on its distribution parameter. They have distinct characteristics highlighted with different colors. Regarding natural gradient vs \ac{vpg}: \ac{vpg} is sensitive to parameter scale differences leading slower convergence. In contrast, the natural gradient converges faster and is more robust, but requires additional computations for Fisher matrix.}

\change{\ac{es} vs \ac{rl}:  \ac{es} treats policy improvement as a \ac{bbo} problem, neglecting the leverage of the problem structure, while \ac{rl} algorithms take a different approach. In summary, each method depicted in Table \ref{tab:optimization_methods} has unique strengths and limitations.
}

\subsection{Dynamical system with deep RL}
Deep \ac{rl} methods are increasingly utilized for \ac{dsil}. 
Kim\etal \cite{kim2018learning} developed a hierarchical deterministic actor-critic (AC) algorithm tailored for \ac{dmp}, which effectively enables robots to learn and generalize skills from human demonstrations. This algorithm excels in guiding robots through complex tasks, demonstrated by its application in a robotic arm's pick-and-place operations.

In \cite{kim2020reinforcement}, Song\etal introduced the \ac{ddpg} algorithm, a standard \ac{rl} method for the precise execution of peg-in-hole assembly tasks. This approach leverages \ac{ddpg} to train a \ac{nnmp}, which generates continuous trajectories for force controllers, improving efficiency and stability over traditional methods.

Wang\etal \cite{wang2022adaptive} introduced an \ac{il} framework for complex contact-rich insertion tasks in robotics. This framework seamlessly integrates \ac{sac} and \ac{dmp} to efficiently generate high-quality trajectories and force control policies. Notably, this approach demonstrates sample efficiency, enhanced safety, and superior generalization compared to conventional methods. It addresses the challenge of reducing the need for repetitive human teaching efforts by adapting learned trajectories to new tasks with similar topological characteristics, thereby alleviating the burden on human demonstrators.
Chang\etal \cite{chang2022impedance} introduced a framework for \ac{il} to acquire both trajectory and force information from demonstrations of peg-in-hole tasks. The framework combines the use of \ac{sac} with admittance control and \ac{dmp} algorithms to learn a high-level control policy that adapts impedance parameters. This approach demonstrates significant robustness even in situations with minor offsets, thereby expanding its applicability in safe and robust contact scenarios.

Sun\etal \cite{sun2022integrating} introduced an actor-critic method integrated with \ac{dmp} to learn controllers for nonprehensile manipulation in a hockey task. The method leverages an \ac{orb} with ranking sampling, leading to a substantial acceleration in the convergence of the learning policy network for efficient \ac{rl}. The integrated approach demonstrates the model's effectiveness, fast learning, and generalization across different conditions.

In \cite{davchev2022residual}, Davchev\etal explored residual learning in \ac{il} for orientation corrections in contact-rich insertion tasks.
The method integrates \ac{rlfd} with \ac{rl} (\ac{ppo}, \ac{sac}) to learn the residual correction policy and thus improve \ac{dmp}'s generalization abilities. 
The approach addresses challenges posed by external factors in the environment, demonstrating increased accuracy on out-of-distribution start configurations. The study evaluates the speed of insertion and demonstrates successful policy transfer across tasks with minimal training. The residual learning technique proves effective in capturing corrections for contact-rich manipulation.

\subsection{Dynamical system with other RL methods}
Beyond the \ac{pi2} and \ac{es} methods, there are additional approaches to \ac{rl} for \acp{ds}. The \ac{em}-based \ac{rl} method offers a probabilistic approach that facilitates flexible exploration and exploitation in robot learning. Unlike policy gradient methods, \ac{em}-\ac{rl} does not require a learning rate and incorporates importance sampling to enhance the utilization of an agent's prior experiences when estimating new exploratory parameters.
The update rule is defined as:
\begin{equation}\label{eq.POWER}
{\theta ^m} \leftarrow {\theta ^{m - 1}}{\rm{ + }}\frac{{\sum\nolimits_i^\mathcal{I} {r\left( {{\theta _i}} \right)\left( {{\theta _i} - {\theta ^{m - 1}}} \right)} }}{{\sum\nolimits_i^\mathcal{I} {r\left( {{\theta _i}} \right)} }}
\end{equation}
where ${\sum\nolimits_i^\mathcal{I} {r\left( {{\theta _i}} \right)} }$ is the cumulative reward; $\theta ^m$ is the policy parameter at the current iteration step $m$.

In \cite{calinon2013compliant, kormushev2010robot}, Calinon\etal introduced a classical \ac{em}-based \ac{rl} approach called \ac{power}. This method is designed to impart complex motor skills to robots, making them adept for use in human-centric environments such as homes and offices. \ac{power} is distinguished by its capacity to enable robots to learn \ac{ds} policies, facilitating task adaptation that demands both compliance and variability, which are vital for safe and efficient human-robot interactions.
Andr{\'e}\etal  \cite{andre2015adapting} applied \ac{power} to enhance the walking capabilities of bipedal robots on slopes. By incorporating \ac{cpg} with \ac{dmp}, the humanoid robot simulation was able to modify its gait for different inclines, demonstrating considerable versatility and practical application potential.
Cho\etal \cite{cho2018relationship} expanded on \ac{power} by sequencing the learning and transfer of motor skills in robots according to the complexity of motion. They evaluated this complexity using temporal and spatial entropy measures across different motion trajectories.
Their method streamlines the \ac{rl} process, allowing robots to efficiently master and adjust motor skills from one task to another.
In \cite{peters2008reinforcement}, Peters\etal developed the \ac{enac} algorithm, a \acp{rl} approach tailored for humanoid robots' complex motor skill acquisition. This approach integrates motor primitives with stochastic policy gradient learning, enabling the algorithm to handle the high-dimensional continuous state and action spaces of human-like limbs.
In subsequent work \cite{kober2010practical}, Peters\etal extended their exploration to various \acp{rl} strategies for optimizing motor primitive policy parameters. This enabled humanoid robots to master intricate tasks, exemplified by training a robotic arm to hit a baseball. Comparative analysis across different \ac{rl} methods, including \acp{vpg}, \ac{power}, and \ac{enac}, revealed that \ac{power} outperforms its counterparts in \ac{dsil}.

In \cite{kober2012reinforcement},  Kober\etal introduced the \acp{crkr} method to learn the mapping from situations to meta-parameters of \ac{dmp}, such as goal location, goal velocity, amplitude, duration, etc.
This work innovatively applied \acp{crkr} in an on-policy meta-parameter \acp{rl} algorithm, focusing on adapting and optimizing the meta-parameters of \acp{dmp} in tasks such as table tennis and dart throwing.

Daniel\etal  \cite{daniel2013autonomous} introduced a \ac{hireps} framework to learn a hierarchical \acp{dmp} policy acquiring versatile motor skills in robot tetherball tasks. They reformulated the task of learning a hierarchical \acp{dmp} policy as a latent variable estimation problem and employed the \ac{hireps} framework with episodic \acp{rl} to facilitate the learning process. The experimental results indicate that \ac{hireps} exhibits faster convergence and the ability to maintain multiple solutions, enabling the robot to wind the ball around the pole from different sides. 

\section{Deep imitation learning}
\label{sec.deepIL}

Deep learning stands out as a promising method with the potential to significantly enhance \ac{il} capabilities, as demonstrated by previous studies \cite{chen2015efficient, chen2016dynamic}. The \ac{dsil} framework is versatile, accommodating not only low-dimensional trajectory input but also extending its application to high-dimensional image/trajectory input \cite{lonvcarevic2021robot}. 
However, when dealing with high-dimensional inputs such as images, the challenge lies in designing an effective deep network structure and loss function that accurately represents the underlying \ac{ds}.

\begin{figure*}
    \centering
     \includegraphics[width=0.9\linewidth]{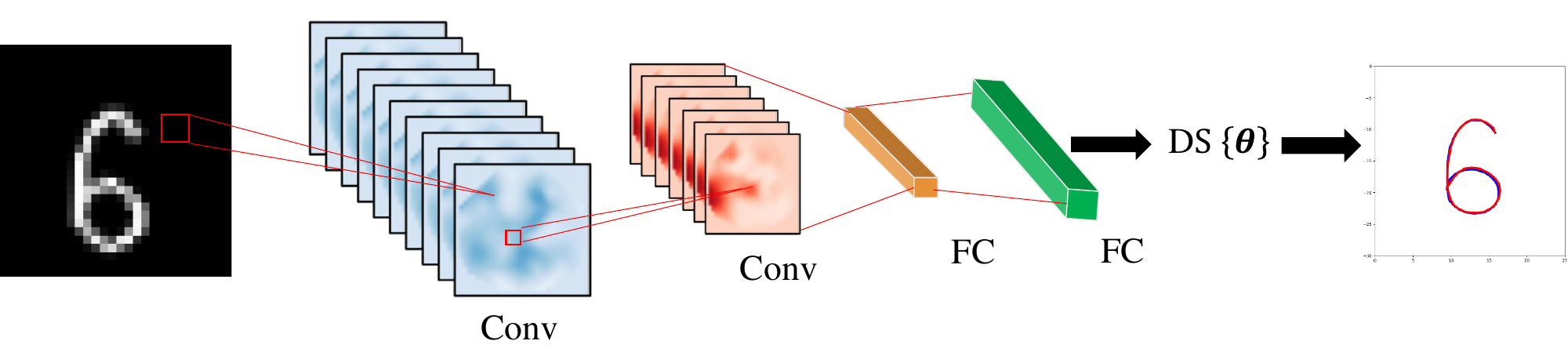}
    \caption{An example deep \ac{il} structures with high dimension image input.}
    \label{fig:deepIL}
\end{figure*}

End-to-end learning from images represents a more intuitive method for \ac{il}, particularly applicable in environments where human participation is impractical or unsafe, such as extreme or hazardous conditions.
Typically, the structure of deep \ac{il} involves convolution layers for feature extraction and \ac{mlp} to acquire the parameters of the \ac{ds}. Fig. \ref{fig:deepIL} illustrates an example deep \ac{il} structure.

In \cite{pervez2017learning}, Affan\etal proposed \ac{cnn}-based deep-\ac{dmp} to learn the mapping from the clock signal and task parameters to the forcing term of a \ac{dmp}. The method employs a \ac{cnn} to model the forcing terms of a \ac{tpdmp} without requiring specialized vision systems or markers. Tested in a sweeping task with a KUKA robot, the proposed D-DMP demonstrates adaptability to new objects and robust performance against disturbances.

However, these methods do not explicitly address the challenge of defining a proper metric for evaluating networks that directly compute \ac{dmp}. 
In \cite{gams2018deep, ridge2020convolutional}, Ude\etal introduced deep \ac{imednet} and \ac{cimednet} designed to convert grayscale images into \ac{dmp}. The dataset comprises MNIST and synthetic MNIST $\footnote[2]{dataset: https://github.com/abr-ijs/digit\_generator}$
, generating image-motion pairs for training. This method effectively reconstructs handwriting movements from these image-motion pairs, showcasing the network's ability to generalize across diverse digit variations and accurately reproduce dynamic trajectories. This suggests potential applications in human-robot interaction systems.
In \cite{ridge2019learning}, Ude\etal further developed a \ac{stimednet} for image-to-motion prediction. The spatial transformer module in \ac{stimednet} aids in finding canonical poses for digits, significantly improving trajectory prediction accuracy under affine transformations. Consequently, \ac{stimednet} outperforms its predecessors, particularly excelling in predicting digit trajectories compared to the existing \ac{imednet} and \ac{cimednet}. The authors extended their approach to a humanoid robot, demonstrating its capability to recognize and reproduce handwritten digits in arbitrary poses.
In \cite{ridge2020training}, Ude\etal proposed a modified loss function that directly calculates the physical distance between movement trajectories, departing from the approach in \cite{gams2018deep, ridge2020convolutional, ridge2019learning} that computed the distance between \ac{dmp} parameters. This new loss function can be extended to \ac{imednet}, \ac{cimednet}, \ac{stimednet}, \ac{rnn}, etc., demonstrating superior performance compared to conventional loss functions in \cite{gams2018deep, ridge2020convolutional, ridge2019learning}.

In \cite{anarossi2023deep},  Anarossi\etal introduced a  \ac{dsdnet} for learning discontinuous motions through deep \ac{dmp}. It presents evaluations across various tasks, including object cutting and pick-and-place, comparing \ac{dsdnet} with \ac{cimednet}. The results demonstrate \ac{dsdnet}'s superior performance in terms of generalization, task achievement, and data efficiency. \ac{dsdnet} overcomes challenges faced by \ac{dsdnet}, particularly in accurately predicting \ac{dmp} parameters for discontinuous motions.
In \cite{gams2020reconstructing}, Pahi{\v{c}}\etal extend a previous methodology introduced in \cite{gams2018deep, ridge2020convolutional, ridge2019learning} by incorporating  \ac{aldmp}. Notably advantageous over standard DMP, \ac{aldmp} exhibit proficiency in managing paths with sharp corners and accommodating trajectories with varying speeds, making them particularly suitable for tasks such as handwriting. The paper also introduces the \ac{vimednet} architecture, an extension of \ac{cimednet}, designed to process images of variable sizes. The efficacy of \ac{vimednet} in reproducing observed digits underscores its significance in handling diverse input scenarios

In \cite{mavsar2021intention} \cite{mavsar2022simulation}, Mavsar\etal proposed a image-to-motion recurrent \acp{nn} for the recognition of human intentions in dynamic human-robot collaboration. This work utilizes dataset consisting of video-trajectory pairs derived from assembly tasks within human-robot interaction contexts. The recurrent image-to-motion networks, trained on these datasets, employ RGB-D images to predict motion trajectories in the form of \ac{dmp} parameters.
Remarkably, the system not only achieves high accuracy in trajectory predictions but also enables real-time adjustments of robot trajectories based on identified human intentions. This dynamic adaptation enhances overall efficiency in shared workspaces, emphasizing the potential of \ac{rnn} in optimizing human-robot collaboration scenarios.
In \cite{bahl2020neural}, Bahl\etal developed \ac{ndp} for end-to-end sensorimotor \ac{il}. The framework incorporates the \ac{dmp} structure as a differentiable layer into \ac{dnn}-based policies, enabling the robot to learn directly from images and generate globally stable motions due to the convergence property inherent in \ac{dmp}. 
The study highlights \ac{ndp}' superior performance in \ac{il} tasks, outperforming traditional \ac{nn} policies such as \ac{cimednet} in dynamic scenarios like digit writing. Additionally, \ac{ndp} shows enhanced efficiency and performance in \ac{rl} setups compared to baseline methods such as \ac{ppo} \cite{martin2019variable}, demonstrating versatility across tasks with varying levels of dynamism and maintaining data efficiency.
However, \acp{dnn} are inherently limited to pick-and-place-style quasi-static tasks compared to the dynamic tasks achieved by \ac{dmp}-based methods. Although \ac{ndp} can improve this performance, they may face challenges in generalizing to unseen state configurations.
In \cite{bahl2021hierarchical}, Bahl\etal introduced \ac{hndp} for dynamic tasks. It involves both local and global policies, with local \ac{ndp} prioritizing task success in specific regions, while the global \ac{ndp} emphasizes learning from images in a generalizable manner. This hierarchical framework, leveraging \acp{ds}, outperforms in dynamic tasks, demonstrating efficacy in capturing high-level variations and achieving better sample efficiency compared to existing methods.
Additionally, \acp{hndp} seamlessly integrate with imitation and \ac{rl} paradigms.
In \cite{shaw2023videodex}, Shaw\etal introduced VideoDex, incorporating \ac{ndp} as physical priors for robot imitation from human video dataset. Utilizing 500-3000 video clips, the study highlights the importance of action priors pretrained on human internet videos for robust generalization of robot behavior. Training involves pretraining with human action priors and final policy training on teleoperated demonstrations. VideoDex consistently outperforms baselines, emphasizing the effectiveness of action priors, and the study explores the impact of physical priors and architectural choices.


The aforementioned research primarily addresses deep \ac{il} for non-autonomous \acp{ds}. Nevertheless, the same concept can be extended to autonomous neural dynamic policies, enabling end-to-end \ac{il} with high-dimensional inputs.
In \cite{totsila2023end}, Totsila\etal introduced \ac{andp} for achieving stable \ac{il} in robotics. This method, building on the strengths of \ac{ndp} and the \ac{seds}, combines \acp{nn} and autonomous \acp{ds}. \ac{andp} is designed to ensure asymptotically stable behaviors, making them well-suited for various \ac{il} scenarios, including those involving diverse sensor inputs like high-dimensional image observations.
In \cite{auddy2023scalable}, Auddy\etal  presented a stable deep \ac{il} framework by integrating two key \ac{nn} components with high-dimension trajectory input. The first \ac{nn} captures the nominal dynamics model, while the second, a \ac{lf} network, is built using an \ac{icnn}. This dual-network structure establishes a stable \ac{ds}, facilitating effective adaptation to new tasks and the retention of knowledge from previous experiences. The incorporation of continual learning, \acp{node}, and hypernetworks further strengthens the model, enhancing its resilience and stability in dynamic learning environments.

\section{\change{Applications}}

\change{
\ac{dsil} offers a promising approach to imitate experts' skills, simplifying the programming of robots and making it widely applicable across various scenarios. In studies such as \cite{zhang2022motion}, \cite{huang2019learning}, and \cite{deng2018learning}, rehabilitative movements were gathered from healthy individuals and trained using \ac{dsil}. These trained movements were then utilized for motion or gait generation in the rehabilitation training of exoskeleton robots.
Furthermore, \ac{dsil} holds potential for applications in surgical robots, enabling them to imitate the skills of surgeons. In \cite{su2020reinforcement}, \ac{rl} combined with \ac{dsil} is utilized to learn puncturing skills under remote center motion (RCM) constraints from surgeon demonstrations. This approach allows for adaptability to different constraints without the need for re-teaching. Similarly, in \cite{zhang2022human}, \ac{dsil} is integrated into a shared control framework, facilitating the transfer of skills from simulation to a da Vinci Research Kit robot for robotic surgery, such as the pick-peg task. In \cite{malekzadeh2014learning}, an \ac{il} method was proposed to imitate the versatile representations of motion to continuum robots. \ac{dsil} was used to transfer motion patterns from an octopus arm to a flexible surgical robot.}

\change{
Beyond medical scenarios, \ac{dsil} finds application in agricultural robots \cite{lauretti2023robot}. In this study, \ac{dsil} was used to encode complex activities and transfer motion skills to robotics, achieving good performance in four agricultural activities, including digging, seeding, irrigation, and harvesting. Similarly, in \cite{lauretti2024new}, a modified \ac{dmp}-based \ac{dsil} method was proposed for ever-changing and mutable agricultural environments, enabling imitation of experts' motion for similar agricultural activities.}

\change{
For interaction or cooperative manipulation tasks where the robot involves physical contact with the environment, control of forces (torque) and positions becomes crucial. In \cite{khoramshahi2019dynamical}, Khoramshahi\etal proposed a \ac{dsil}-based impedance control method to enable a robot to comply with human intention. This work successfully addressed transportation tasks where human and robot collaborated to carry and place a heavy object across the aisles with shelves on each side. In \cite{AbuDakka2015Adaptation}, Abu-Dakka\etal applied a \ac{dsil} method to imitate orientation for peg-in-hole assembly tasks, considering position, orientation, and force in complex physical interaction environments.}

 \begin{figure}
	\begin{center}
		\includegraphics[width=\linewidth]{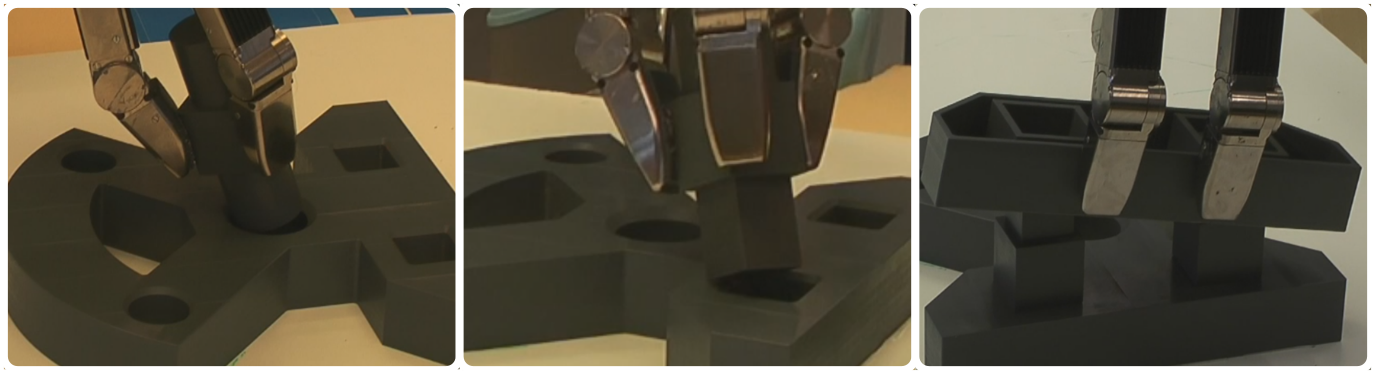}
		\caption{An example of using \ac{dsil} in Peg-in-Hole \cite{AbuDakka2015Adaptation}.
		}\label{fig.dmpPIH}
	\end{center}
\end{figure}

\change{
In underwater intervention tasks, \ac{dsil} plays a pivotal role in control and motion generation. In \cite{carrera2014learning}, Carrera\etal proposed a \ac{dsil} method to teach autonomous underwater vehicles intervention navigation towards goal positions while navigating through underwater obstacles.
Similarly, in \cite{zhang2022leveraging}, Zhang\etal designed and controlled a robotic fish to achieve pose regulation in a swimming tank while considering external disturbances. To simultaneously attain the goal position and orientation, they proposed a learning framework that combined \ac{dsil} and deep reinforcement learning to train a policy from demonstrations.
}

 \begin{figure}
 	\begin{center}
 		\includegraphics[width=0.6\linewidth]{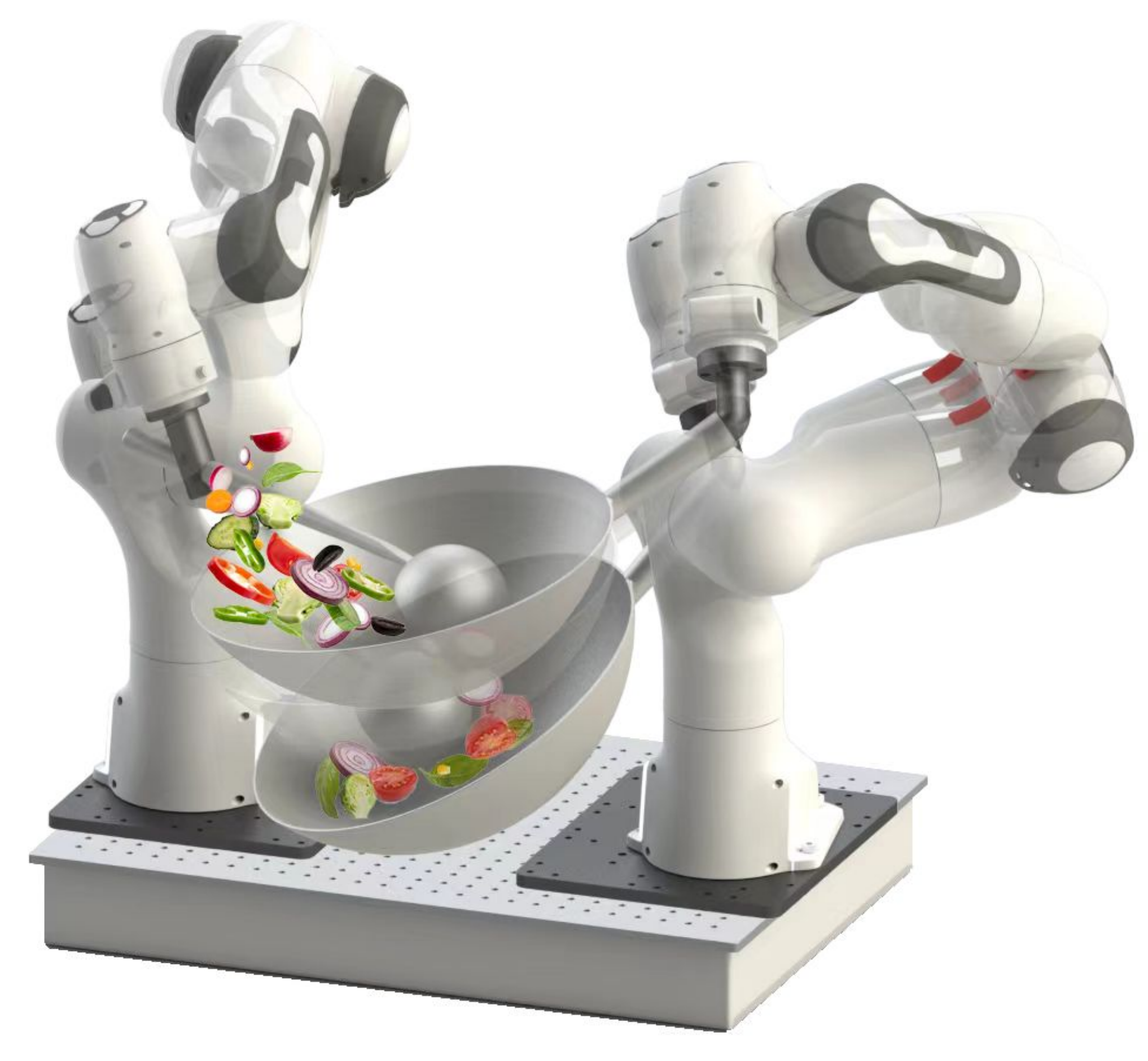}
 		\caption{An example of using \ac{dsil} in cooking \cite{liu2022robot}.
 	}\label{fig.application_cook}
 	\end{center}
 \end{figure}

\change{
\ac{dsil} can also be extended to daily life tasks, such as cooking. In \cite{liu2022robot}, Liu\etal modeled the skills of cooking art stir-fry and proposed a \ac{dsil} method to learn deformable object manipulation from a cook using a bimanual robot system, shown in  Fig. \ref{fig.application_cook}.
}

\begin{figure}
	\begin{center}
		\includegraphics[width=\linewidth]{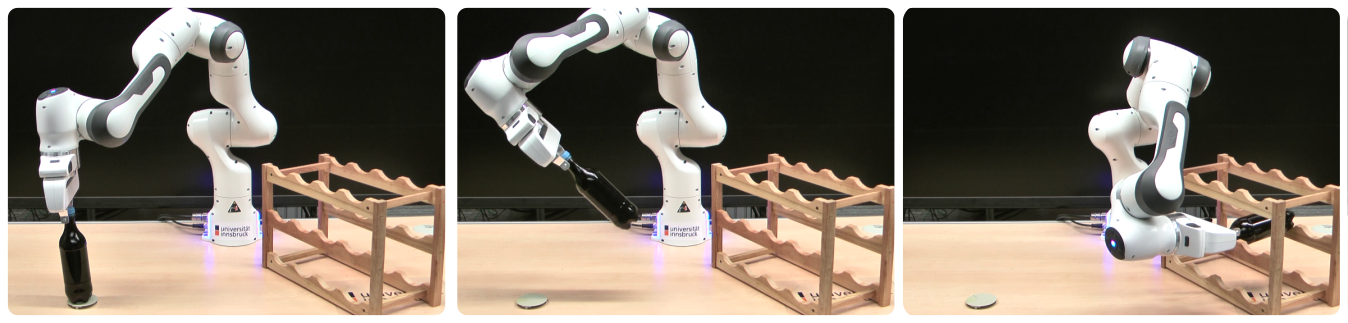}
		\caption{An example of using \ac{dsil} in Bottle Stacking \cite{saveriano2023learning}.
		}\label{fig.dsBottleStack}
	\end{center}
\end{figure}

\change{In addition to Euclidean space, \ac{dsil} in non-Euclidean spaces such as Riemannian manifolds is also explored \cite{saveriano2023learning} to solve some real-world applications, \eg bottle stacking shown in Fig. \ref{fig.dsBottleStack}, that involves non-Euclidean data, \eg orientation and impedance. Their approach discusses the utilization of mathematical techniques derived from differential geometry to ensure that learned skills adhere to the geometric constraints imposed by these complex spaces. The bottle stacking application highlights the versatility of \ac{dsil}-based trajectory generators, allowing for convergence to different goals even within complex non-Euclidean environments.
}

\begin{figure}
	\begin{center}		\includegraphics[width=0.7\linewidth]{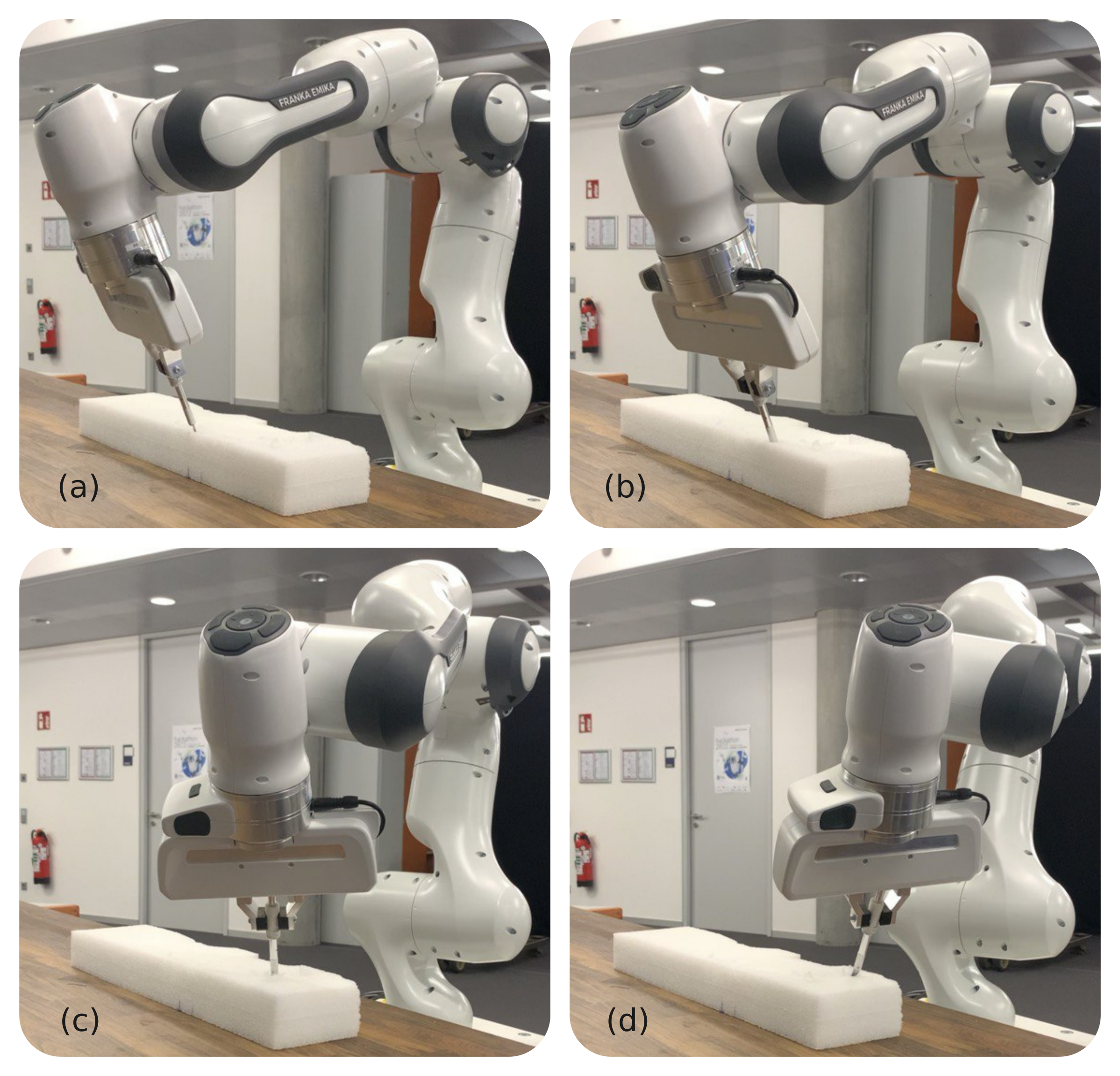}
		\caption{\change{An illustration showcasing the application of \ac{dsil} in assisted teleoperation for cutting skills \cite{Shen2024Safe}.}}
        \label{fig.cutting}
	\end{center}
\end{figure}

\change{Another example of innovative use of \ac{dsil} in non-Euclidean spaces, ensuring the safe execution of learning orientation skills, represented in Lie group SO(3), within constrained regions surrounding a reference trajectory, shown in \cite{Shen2024Safe}. The approach involves learning a stable \ac{ds} on SO(3), extracting time-varying conic constraints from expert demonstrations' variability, and bounding the \ac{ds} evolution with conic \ac{cbf} to fulfill these constraints. This approach holds promise for applications such as tissue resection in robotic surgery, where the precise orientation of surgical instruments is crucial for performing delicate procedures accurately and safely. Figure \ref{fig.cutting} is an illustrative example of such cutting skills.}

\section{Discussion}
\label{sec.future_directions}
The fusion of machine learning and \acp{ds} in the context of robot \ac{il} is a burgeoning topic that has captivated significant attention within the research community, setting the stage for further investigation. Despite notable achievements across diverse domains, particularly in robotics, several challenges persist, especially in complex environments. In this section, we discuss the limitations of \ac{dsil} and outline potential future research directions.

\begin{table*}[pos=ht]
\caption{A comparison of the main approaches for \ac{dsil}.}
\rmfamily\centering
\resizebox{\textwidth}{!}{%
\renewcommand\arraystretch{1.1}{
\begin{tabular}{|m{0.01\textwidth}|m{0.09\textwidth}|>{\centering\arraybackslash}m{0.47\textwidth}|>{\centering\arraybackslash}m{0.47\textwidth}|}
\noalign{\hrule height 1.5pt}
\multicolumn{2}{|c|}{\textbf{Approach}}        & \textbf{Advantages}    & \textbf{Disadvantages}                             \\ \hline
\multirow{27}{*}{\rotatebox[origin=c]{90}{\textbf{Existing}}} & Stability methods            & 
\begin{itemize}[leftmargin=8pt,noitemsep, topsep=2pt]
	{\item Model-based solutions, leading to more efficient learning, faster convergence, and reduced computational requirements.
	\item Robustness which ensures effectiveness in the presence of uncertainties or variations in the environment.
	\item The stability contributes to overall system safety, minimizing the likelihood of unsafe actions in real-world applications.
	\item Enhance generalization making them applicable across a diverse range of scenarios.
	\item Facilitate the convergence and ensure that the training process is more reliable and converges to effective policies.
	\item Often require fewer training samples, leading to more efficient and economical training processes.
	\item Align with fundamental principles of control theory, providing a solid theoretical foundation for the design and analysis of \ac{dsil}. This integration can lead to more principled and well-understood control strategies.
	}\vspace*{-\baselineskip}
\end{itemize}
  & 
  \begin{itemize}[leftmargin=8pt,noitemsep, topsep=2pt]
  	{\item Over-reliance on simplified models which could lead to limitations in handling intricate environments.
  	\item Complex systems with nonlinear dynamics may pose challenges for stability methods, as these methods might struggle to accurately model and account for nonlinear behaviors.
  	\item Despite their robustness, stability methods may face difficulties in generalizing well to entirely unforeseen or extreme conditions that were not adequately represented in the training data.
  	\item Stability methods might encounter difficulties in effectively handling high-dimensional state or action spaces, limiting their applicability in certain complex robotic tasks.
  	\item The success of stability methods relies heavily on having an accurate understanding of the underlying system dynamics. In cases where these dynamics are not well-known, the effectiveness of stability methods may be compromised.
  	} \vspace*{-\baselineskip}
  \end{itemize}
  \\ \cline{2-4} 
& Exploration learning & 
\begin{itemize}[leftmargin=8pt,noitemsep, topsep=2pt]
	{\item Actively exploring the state-action space, making the learned policies more resilient to uncertainties and variations in the environment.
	\item Promote better generalization, allowing the learned policies to perform well across a wider range of scenarios, including those not encountered during training.
	\item Mitigate the risk of getting stuck in local optima by encouraging the exploration of different regions of the state-action space.
	\item Strike a balance between exploration (trying new actions) and exploitation (leveraging known actions), optimizing the trade-off to efficiently discover and exploit rewarding policies.
	\item Facilitate the adaptation of learned policies to changing environments by continuously exploring and updating the model based on new information.
	}\vspace*{-\baselineskip}
\end{itemize}
& 
\begin{itemize}[leftmargin=8pt,noitemsep, topsep=2pt]
	{\item Active exploration methods may incur higher computational or time costs.
	\item Improper balance between exploration and exploitation may lead to suboptimal policies or inefficient learning.
	\item Excessive exploration can introduce noise in the learning process, leading to a less stable policy.
	\item The effectiveness of exploration learning is closely tied to the chosen exploration strategy.
	\item Sensitivity to hyperparameters.
	\item Limited exploration in limited time.
	\item They may face challenges in high-dimensional state or action spaces, where the search for informative experiences becomes more complex.
	}\vspace*{-\baselineskip}
\end{itemize}     
\\ \cline{2-4} 
& Deep \ac{il}                      & 
\begin{itemize}[leftmargin=8pt,noitemsep, topsep=2pt]
	{\item Excel at automatically learning intricate representations of the input data, enabling the model to capture complex relationships in high-dimensional state and action spaces.
	\item They can automatically extract hierarchical features, allowing them to capture both low-level details and high-level abstractions in the dynamical system, facilitating more effective policy learning.
	\item Enable end-to-end learning, which simplifies the modeling process and allows the model to learn directly from raw input data.
	\item Well-suited for capturing nonlinear relationships in dynamical systems, making them effective in scenarios where the underlying dynamics are complex and nonlinear.
	\item Continuous improvement with more data.
	}\vspace*{-\baselineskip}
\end{itemize}
& 
\begin{itemize}[leftmargin=8pt,noitemsep, topsep=2pt]
	{\item Often require large amounts of labeled training data to effectively learn complex representations, and obtaining such datasets may be challenging or expensive.
	\item Training can be computationally intensive, requiring significant computational resources, which may limit their applicability in resource-constrained environments.
	\item Overfitting risk.
	\item Hyperparameter sensitivity.
	\item Adapting deep \ac{il} models to non-stationary environments, where the underlying dynamics change over time, can be challenging, potentially leading to degradation in performance.
	}\vspace*{-\baselineskip}
\end{itemize}
\\ \hline \end{tabular}
}}
\resizebox{\textwidth}{!}{%
	\renewcommand\arraystretch{1.1}{
		\begin{tabular}{|m{0.01\textwidth}|m{1.08\textwidth}|}\hline
\rotatebox[origin=c]{90}{\textbf{Envisioned}}    &
	\begin{itemize}[leftmargin=8pt,noitemsep, topsep=3pt]
        {\item \textbf{Accuracy, stability, and adaptability:} \ac{dsil} is expected to demonstrate high accuracy and stability in imitation, along with the ability to adapt to new tasks without requiring re-teaching from experts.
        \item \textbf{Data-efficiency in learning:} The system should exhibit efficiency in learning by effectively adapting to a wide range of scenarios with minimal demonstrations.
        \item \textbf{Safety exploration:} Particularly in physical systems like robots, safety during exploration is crucial. \ac{dsil} should prioritize safety through \ac{rl} or evolution strategies in real-world environments.
        \item \textbf{High-dimensional inputs:} The capability to imitate from high-dimensional inputs, such as images or videos, is essential for addressing the complexity of real-world tasks in \ac{dsil}.
        }\vspace*{-\baselineskip}
    \end{itemize} \\ \noalign{\hrule height 1.5pt}
\end{tabular}
}}
\label{tab:main_approach}
\end{table*}

\subsection{Guidelines}

The preceding sections have provided an overview of two categories of \ac{dsil} and the varied approaches within this domain. Recognizing that each \ac{dsil} system or approach is designed to address specific conditions in applications, this section will delve into: 
(\emph{i}) Advantages and Disadvantages:
	A comprehensive discussion on the merits and drawbacks of different \ac{dsil} systems and mentioned approaches provides a nuanced understanding of their strengths and limitations.
(\emph{ii}) Potential Challenges:
	Identification and exploration of potential challenges in the context of \ac{dsil}, shedding light on areas that require further attention and innovative solutions.

\subsubsection{\ac{ndsil} vs \ac{dsil}}
In the context of \ac{il}, a spectrum of machine learning models can represent either \ac{ds} or non-autonomous \ac{ds}. The distinction between \ac{ndsil} and \ac{adsil} reveals a mutually exclusive relationship. Non-autonomous models, such as \ac{dmp} \cite{ijspeert2013dynamical,saveriano2023dynamic} and \acp{hsmm} \cite{calinon2011encoding}, offer flexibility in mimicking time-varying behavior, proving beneficial for adaptation to changing conditions and providing a realistic representation of system dynamics. These models are particularly suited for systems influenced by external inputs or time-dependent factors.

Nevertheless, non-autonomous models come with drawbacks, including (\emph{i}) increased complexity in learning algorithms, (\emph{ii}) challenges in interpreting and analyzing models, (\emph{iii}) dependency on high-quality training pairs, and (\emph{iv}) heightened computational demands compared to autonomous systems.

Autonomous systems, such as  \ac{seds} \cite{khansari2011learning}, characterized by stable behavior, excel in learning and generalization. They can cease at any time step without altering the agent's agent state, and their lack of external influences simplifies the prediction and understanding of the model's behavior over time. Although autonomous systems may lack flexibility in capturing time-varying dynamics, they stand out in stability, predictability, and simplicity, making them suitable for time-independent applications. 

The choice between \ac{ndsil} and \ac{adsil} hinges on the specific requirements and characteristics of the application under consideration.

\subsubsection{Choose the appropriate stability methods}
Choosing the most suitable stability method depends on the specific characteristics and requirements of the \ac{ds} under consideration. Each method has its unique strengths and limitations, making them more or less suitable for different application scenarios.

Lyapunov stability method is versatile and applicable to a broad spectrum of \acp{ds}, serving as a fundamental stability analysis tool. It provides a robust theoretical foundation, facilitating the development of rigorous stability proofs. Moreover, it is well-suited for global stability analysis, offering insights into the system's behavior across the entire state space. Nevertheless, Lyapunov stability conditions can be overly conservative, imposing strict constraints that may lead to rejecting potentially stable systems, and compromising accuracy. Additionally, identifying suitable Lyapunov candidate functions can be challenging, especially for complex or nonlinear systems, thereby limiting its applicability.

\ac{ct} stability method offers an effective alternative for nonlinear and time-varying systems, quantifying contraction properties and providing a metric for trajectory convergence. Moreover, it excels in robust stability analysis, demonstrating resilience to disturbances and uncertainties. Nevertheless, the computational complexity of implementing contraction metrics, particularly for high-dimensional systems, can impact real-time applications. Although it is effective for nonlinear systems, \ac{ct} may not be the preferred choice for linear systems, where simpler techniques like Lyapunov stability are more straightforward.

The diffeomorphism stability method preserves topological properties, ensuring stability without sacrificing critical structural features. Moreover, it applies to chaotic systems \cite{devaney2018introduction}, providing a method to stabilize and control chaotic behavior. Diffeomorphisms are inherently smooth transformations, ensuring smooth transitions in the state space.
However, implementing diffeomorphism-based stability methods can be complex, especially for high-dimensional systems. Although diffeomorphism-based stability methods are powerful, their theoretical foundation may not be as established and general as Lyapunov stability.

Stability methods (Lyapunov, \ac{ct}, and diffeomorphism) rely on accurate models of \ac{dsil}, posing challenges in deriving precise stability conditions, especially for complex models such as deep \acp{nn}.

\subsubsection{Safety of policy improvement}

The incorporation of exploration techniques enhances the efficiency of \ac{dsil} in exploring complex and uncharted tasks. This is particularly valuable, in scenarios, when the initial dataset lacks coverage of all possible situations. Combining \ac{dsil} with exploration strategies, such as those from \ac{rl}, deep \ac{rl}, or \ac{es}, enhances the robot's ability to generalize learned policies. This adaptability is crucial for navigating diverse and dynamic environments. By incorporating exploration techniques, \ac{dsil} gains flexibility beyond the limitations of demonstrations, proving beneficial in real-world scenarios when encountering novel situations absent from the training data.

However, ensuring the safety of robotic systems is important to prevent damage. Operating within appropriate workspaces and adhering to joint physical limitations becomes challenging, especially in policy exploration scenarios using \ac{rl} within \ac{dsil}. Safety requirements impose constraints on the robot's exploration space, potentially impeding the discovery of optimal policies and leading to suboptimal outcomes.

\change{
Therefore, the policy parameters getting stuck in locally optimal solutions while considering safety constraints in safe exploration presents a significant challenge. The candidate solutions are listed as follows:
\begin{itemize}
\item Diverse exploration: Implement strategies that encourage RL agents to explore various actions and states to avoid getting trapped in local optima.
\item Soft constraint relaxation: Temporarily relax safety constraints during training to allow for more exploration without compromising safety.
\item Reward shaping: Adjust reward functions to discourage unsafe actions, guiding RL agents towards safer policies.
\item Prioritized experience replay: Emphasize experiences with safety violations in the replay buffer, helping RL agents learn from past mistakes.
\end{itemize}}

Among exploration methods, \ac{pi2} operates as a closed-loop method, lacking explicit gradient information for guiding policy improvement. Alternately, \ac{es} methods function as open-loop ones, capable of combining with policy gradients but utilizing gradient information solely as sorting indices for the next generation. Addressing limitations inherent in each approach involves a promising solution: combining policy gradients with \ac{pi2}, as demonstrated by studies like \cite{hu2023pi} and \cite{varnai2020path}. This hybrid approach leverages the strengths of both methods, using gradient information within the closed-loop framework of \ac{pi2} for more informed policy improvement. This integration facilitates a more robust and effective exploration strategy while maintaining a focus on safety considerations.

\subsubsection{Build deep \ac{dsil} model}

Harnessing the capabilities of \acp{dnn} offers a promising avenue for enabling robots to learn directly from high-dimensional data, \eg images and videos. This approach facilitates the direct acquisition of stability skills directly from perceptual inputs. The \ac{ds} model is constructed using \acp{dnn}, providing increased flexibility and trainability compared to traditional machine learning models, which is particularly valuable for when tackling complex tasks.

The primary challenge in the context of deep \ac{dsil} lies in ensuring stability. While related works, such as \ac{imednet}, \ac{ndp}  and \ac{stimednet}, leverage \ac{dmp} and inherit global stability, establishing stability conditions for general deep \ac{dsil} model becomes intricate due to the absence of an explicit expression. Similarly, incorporating additional constraints or employing policy exploration methods within the deep \ac{dsil} model poses challenges.

Furthermore, related works, including \ac{imednet}, \ac{ndp} and \ac{stimednet} have been trained with MNIST and synthetic MNIST image-motion pairs dataset. The challenge lies in collecting a diverse motion-image pairs dataset for various tasks.

The advantages and disadvantages of the surveyed approaches are summarized in Table \ref{tab:main_approach}.

\subsection{Challenges and future directions} 

\subsubsection{Dataset}

In \ac{il}, the dataset plays a crucial role by providing the expert policy. It consists of pairs of observations and corresponding actions, helping the algorithm in learning to mimic expert behavior.  

\noindent\textbf{Imperfect or limited  data}: Typically, \ac{il} relies on high-quality demonstration data. However, real-world demonstrations often lack structure, containing noise and outliers. Therefore, it necessitates the development of algorithms for automatically identifying and handling unstructured data, ensuring that the learning algorithm prioritizes meaningful information. Future research avenues include exploring advanced filtering methods \cite{antotsiou2021adversarial} and outlier detection algorithms \cite{li2021automated} to enhance the model's resilience to noisy demonstrations, thereby promoting more accurate learning.

Acquiring a substantial number of expert demonstrations is often impractical or expensive. Future research endeavors should explore strategies for efficient learning from limited demonstrations. Techniques like data augmentation \cite{antotsiou2021adversarial, zolna2021task}, transfer learning \cite{maeda2017probabilistic}, and one-/few-shot learning \cite{mandi2022towards, dance2021conditioned} can be investigated to optimize the utility of scarce data resources.
For instance, the \ac{dagger} \cite{menda2019ensembledagger, kelly2019hg} continually engages with the environment to generate new data and seeks guidance from the expert policy on the newly generated data. The algorithm iteratively enhances the dataset by integrating both cloned and expert-guided samples. Similar strategies can be employed to enhance the efficiency of sample utilization in \ac{dsil}.

\noindent\textbf{High dimension data}: Section \ref{sec.deepIL} partially addresses aspects of handling high-dimensional data in \ac{dsil} through models like \ac{imednet}, \ac{ndp}, \ac{stimednet}, and \ac{vimednet}. However, the reliance on motion-image pairs data presents a challenge for effectively utilizing pure image or video data in \ac{dsil}.

The primary challenge lies in developing robust feature extraction methods for high-dimensional data. The objective is to derive temporal logic trajectories through \ac{il} from images, a task that differs from conventional image processing tasks in deep learning.  
Secondly, as \ac{dsil} is a set of differential equations, the main challenge is to represent the \ac{dsil} model as a deep \ac{ds}. This representation should handle the intricacies of high-dimensional input-output mappings, addressing stability, robustness, safety constraints, and more. Additionally, real-time performance becomes a challenge due to the potential increase in computational demands with high-dimensional input.

Future directions may concentrate on constructing a deep \ac{dsil} model adept at extracting robust features from images to facilitate motion imitation. Additionally, emphasis can be placed on integrating considerations of stability, robustness, and safety constraints into the model for comprehensive performance. These advancements aim to enhance the adaptability and efficiency of \ac{dsil} models when dealing with high-dimensional data.

\noindent\textbf{Geometry-constrained data}:
Traditional \ac{dsil} encounters challenges in directly encoding skills with geometry constraints, particularly concerning orientation data \cite{michel2023orientation}. Integrating unit quaternions, for instance, may not maintain quaternion norm unity, requiring additional constraints during integration. While \ac{dsil} has partially addressed geometry constraints with specific data types, such as orientation profiles \cite{abudakka2021Periodic}, stiffness/damping \cite{chen2021closed}, and manipulability profiles (manifold of \ac{spd} data) \cite{abudakka2020Geometry}, extending similar approaches to other Riemannian manifolds like the Grassmannian or Hyperbolic manifolds remains non-trivial and has not been fully addressed \cite{saveriano2023dynamic}. 

Future work can extend the \ac{dsil} method to other Riemannian manifolds and represent the geometry constraints with specific data types. This extension aims to enhance the applicability and versatility of \ac{dsil} in handling a broader range of geometry-constrained data, providing more comprehensive solutions for encoding skills in various robotic applications.

\subsubsection{Skills generalization}

\noindent\textbf{Accuracy and stability}:
Achieving robustness to task variability while maintaining accuracy presents a pivotal challenge in skills generalization. 
The conflict between accuracy and stability arises in modeling non-monotonic dynamics, particularly scenarios where the system temporarily moves away from the attractor \cite{fichera2022linearization}.
Despite existing efforts to partially resolve the conflict between stability and accuracy with less conservative stability conditions, such as \ac{clfdm} \cite{khansari2014learning}, and \ac{nsqlf} \cite{jin2023learningss, sindhwani2018learning} (as discussed in Section \ref{sec.stable}), there remains a gap in simultaneously improving robustness and accuracy. The current state of research highlights the need for advancements to harmoniously blend stability, accuracy, and robustness in \ac{dsil} models.

The concentration on single-attractor \ac{ds}, unless stability constraints are integrated, demands prior knowledge of the attractor's location. This assumption significantly limits the applicability of stability methods to learning uni-modal dynamics. The incorporation of multiple dynamics into a single control law based on \ac{ds} adds complexity to effectively model a broader range of dynamical behaviors \cite{fichera2022linearization}.

Future research avenues should be directed towards enhancing \ac{dsil} model performance to achieve a harmonious balance between stability, accuracy, and robustness. Advancements in this direction will contribute to the broader adoption of \ac{dsil} models in real-world applications with diverse and dynamic task environments.

\noindent\textbf{Unseen configurations}:
A significant challenge in \ac{dsil} arises from the limitations of covering all situations within the demonstrations. \ac{il} models often encounter difficulties when confronted with scenarios featuring configurations not present in the training demonstrations. To overcome this challenge, research should focus on enabling models to extrapolate knowledge from existing demonstrations, allowing them to handle novel and unseen situations. This entails exploring techniques that capture underlying dynamics capable of generalizing well across diverse environments, and recognizing the inadequacy of imitating a static set of demonstrations for dynamic and evolving environments.

Addressing the domain adaptation problem of unseen configurations, transfer learning is a candidate solution \cite{desai2020imitation}. This method facilitates the transfer of acquired skills across diverse domains. Domain adaptation is particularly evident in scenarios involving \ac{il} with policy exploration using \ac{rl}, such as simulation-to-real, simulation-to-simulation, and real-to-real transitions \cite{kim2020domain}. To effectively tackle this challenge, future investigations should concentrate on exploring domain adaptation methods \cite{nguyen2021tidot}. These endeavors aim to enhance the transferability of acquired skills, ensuring their applicability and effectiveness across varied and dynamic environments.

Another solution involves exploring continual learning methods \cite{gao2021cril} tailored for handling unseen configurations. This approach allows the model to adapt incrementally to new situations over time, accumulating knowledge from ongoing interactions. Potential research directions include advancements in transfer learning, where efforts could be directed toward developing transferable knowledge representations that empower the model to adapt efficiently to new and unseen situations.

\change {\noindent\textbf{Advanced learning methods}: 
Although  Bahl\etal proposed the structure of a deep \ac{dsil}--\ac{ndp} in \cite{bahl2020neural}, a persistent challenge in the continuous evolution of deep \ac{il} is the limited capacity of existing models to adequately represent the diversity of movements. Consequently, the construction of a larger \ac{dsil} model remains an ongoing challenge.
The progress in \ac{llm} offers a promising solution for \ac{dsil} \cite{ha2023scaling}. \acp{llm} possess the capability to directly learn from diverse data modalities, including language, images, and trajectories, leading to improved skill generalization. This opens up new possibilities for enhancing the versatility and adaptability of robot \ac{il} systems.
By harnessing the expressive power of \acp{llm}, robots can better comprehend the contextual nuances of instructions, interpret visual cues from images, and assimilate the temporal dynamics of trajectories. This holistic learning approach enables the robot to generalize acquired skills across a broader spectrum of tasks and scenarios.}

\change {In addition to \acp{llm}, drawing inspiration from the work of Neural \ac{ode} by Chen\etal \cite{chen2018neural}, which utilizes \acp{ode} to model the dynamics of the network's hidden states in \acp{dnn}. Instead of using discrete layers to model transformations, Neural \acp{ode} use continuous transformations defined by differential equations. Since \ac{dsil} is a set of \acp{ode}, future research could explore integrating Neural \acp{ode} into \ac{dsil} \cite{nawaz2023learning}.}

\change {\noindent\textbf{Advanced control methods}:
Presently, most \textit{fusion} of \ac{dsil} with control theory are offline methods, \eg stability method \ac{lf}. Despite the existence of online \ac{apf}-based obstacle avoidance methods, as discussed in Section \ref{sec.ndsil}, such as point-static \cite{warren1989global} and volume-dynamic \cite{ginesi2021dynamic}, the challenge lies in effectively applying control methods online to enhance generalization, especially concerning stability and safety.}
\change{
As reported by Nawaz\etal \cite{nawaz2023learning}, a nominal \ac{dsil} was trained offline, selecting a target point at each time step using Neural \acp{ode}. Subsequently, \ac{clf} and \ac{cbf} are employed to ensure stability and safety for online motion correction.}
However, a limitation arises from the potential risk of converging to a local minimum due to conflicting safety and task-completion constraints.
Future research directions may explore advanced online control methods to correct motion in real-time, aiming to achieve globally optimal policies for skills generalization.

\change{
Passivity-based control approach can also be used to improve the stability of first-order \ac{dsil}.
In their works \cite{chen2021closed} and \cite{michel2023passivity}, Lee\etal introduced a passivity-based control approach aimed at creating a \ac{vsds} through the regulation of desired motion within the original dynamical system. This \ac{vsds} is distinguished by its safety, compliant interaction behavior, and ability to converge to a reference motion.
Additionally,  as the \ac{dsil} represents a classical dynamical system, alternative control methods such as sliding mode control \cite{incremona2016sliding} and adaptive control methods \cite{polycarpou1992modelling} may offer viable solutions.
Future research directions may explore different control methods to improve the stability of \ac{dsil}.}

\section{Conclusion}
\label{sec.conclusion}
In this paper, we conduct a comprehensive survey on \ac{dsil}, a theory fusion approach that combines machine learning with \acp{ds}. Our primary objective is to categorize and review the extensive literature on \ac{dsil}, employing a systematic review approach and automated searches for relevant papers.
We first discuss two types of \ac{dsil}: \ac{ndsil}. We then propose a taxonomy that includes three main stability methods: Lyapunov stability, contraction theory, and diffeomorphism mapping.
Our survey includes a thorough examination and comparison of policy exploration models in \ac{dsil}, encompassing traditional \ac{rl}, evolution strategies methods, deep \ac{rl}, and establishing a connection from \ac{pi2} to \ac{es} on \ac{dsil}.
Furthermore, we delve into the realm of deep \ac{il} models within \ac{dsil}. Finally, we highlight several open problems and challenges within \ac{dsil}, identifying avenues for future research.

\bibliographystyle{unsrt}

\bibliography{references}

\end{document}